\documentclass[3p,twocolumn]{elsarticle}

\usepackage{hyperref}

\usepackage{color,soul}
\usepackage{amssymb,adjustbox}
\usepackage{amsmath,graphicx,epstopdf,siunitx,caption}

\makeatletter
\def\ps@pprintTitle{%
	\let\@oddhead\@empty
	\let\@evenhead\@empty
	\def\@oddfoot{\footnotesize\itshape Published in Journal of Visual Communication and Image Representation, 49 (2017) 401-411.\hfill \today}%
	\let\@evenfoot\@oddfoot}
\makeatother

\journal{Journal of Visual Communication and Image Representation}

\begin{document}

\begin{frontmatter}

\title{Next-Active-Object prediction from Egocentric Videos}

\author{Antonino Furnari\corref{mycorrespondingauthor}}
\cortext[mycorrespondingauthor]{Corresponding author}
\ead{furnari@dmi.unict.it}
\author{Sebastiano Battiato\corref{}}
\ead{battiato@dmi.unict.it}
\address{University of Catania - Department of Mathematics and Computer Science}

\author{Kristen Grauman}
\address{The University of Texas at Austin - Computer Science Department}
\ead{grauman@cs.utexas.edu}

\author{Giovanni Maria Farinella\corref{}}
\ead{gfarinella@dmi.unict.it}
\address{University of Catania - Department of Mathematics and Computer Science}

\begin{abstract}

Although First Person Vision systems can sense the environment from the user's perspective, they are generally unable to predict his intentions and goals. Since human activities can be decomposed in terms of atomic actions and interactions with objects, intelligent wearable systems would benefit from the ability to anticipate user-object interactions. Even if this task is not trivial, the First Person Vision paradigm can provide important cues to address this challenge. We propose to \emph{exploit the dynamics of the scene to recognize next-active-objects before an object interaction begins}. We train a classifier to discriminate trajectories leading to an object activation from all others and forecast next-active-objects by analyzing fixed-length trajectory segments within a temporal sliding window. The proposed method compares favorably with respect to several baselines on the Activity of Daily Living (ADL) egocentric dataset comprising 10 hours of videos acquired by 20 subjects while performing unconstrained interactions with several objects.
\end{abstract}


\end{frontmatter}

\section{Introduction and Motivation}
The main advantage of wearable cameras is their ability to sense the world from the user's perspective. This makes them ideal for building egocentric systems able to assist the user and augment his abilities~\cite{Damen2015,kanade2012first,pentland1998visual}. Towards this direction, researchers have investigated methods to understand the user's environment~\cite{pentland1998visual,Furnari_2015_ICCV_Workshops,furnari2016recognizing,
	furnari2016temporal,Templeman2014place,torralba2003context}, model his attention~\cite{li2013learning,su2016detecting}, categorize his motion~\cite{poleg2014temporal,poleg2015compact}, summarize the acquired video~\cite{lee2015predicting,xu2015gaze}, recognize performed activities~\cite{fathi2011understanding,Fathi2012,Ma2016going,Pirsiavash2012}, and provide assistance~\cite{Damen2015,Soran2015}.

Despite the fact First Person Vision (FPV) systems are exposed to a huge amount of user-related information, they generally are not able to predict the user's intent and final goals. This makes user-assistance and human-machine interaction limited. As claimed in previous works~\cite{Koppula2013,lan2014hierarchical,Zhou2015}, the ability to anticipate the future is an essential property that humans exploit on a daily basis in order to communicate and interact with each other.
For instance, predicting object interactions before they actually occur can be useful to provide guidance on object usage~\cite{Damen2015}, issue notifications~\cite{Soran2015} or assist the user~\cite{Koppula2013}. Anticipated object interactions can tell us something more about the user's long term goals, as well as the intended activities. Indeed, as observed in~\cite{Ma2016going,Pirsiavash2012,Zhou2016}, it is advantageous to decompose long term egocentric activities in terms of ``atomic actions'' and interactions with objects to improve the final activity recognition task. Previous works investigated anticipation and early recognition of egocentric activities~\cite{Soran2015,Ryoo2015a}. However, being able to anticipate the future at the finer granularity of object interactions is important especially for wearable intelligent systems, which need to reactively communicate with the user in order to provide him feedback and assistance.

Taking advantage of the First Person Vision paradigm, we introduce the novel task of predicting \emph{which objects the user is going to interact with {next}} from egocentric videos. Following recent literature which explores the importance of ``active objects'' for activity understanding~\cite{Pirsiavash2012,Bertasius2016}, we refer to our task as \emph{``next-active-object prediction''}. {According to}~\cite{Pirsiavash2012}, {active objects are those which are being manipulated by the user at the moment. In contrast with the classic idea of active objects, we aim at detecting active object \emph{before} the manipulation actually begins.}

Forecasting next-active-objects in unconstrained settings is hard since humans interact with objects on the basis of their final goals and the responses they get from the environment. Traditional approaches which detect active objects on the basis of the way their appearance changes during manipulation~\cite{Pirsiavash2012} or the presence of hands~\cite{fathi2011understanding,li2015delving} are not directly exploitable in this context where predictions are to be made \emph{before} the object actually becomes active. Moreover, real systems should be able to deal with an ``open world scenario'' where object categories {that may appear in the field of view} might not be known in advance.

\begin{figure}
	\includegraphics[width=0.24\linewidth]{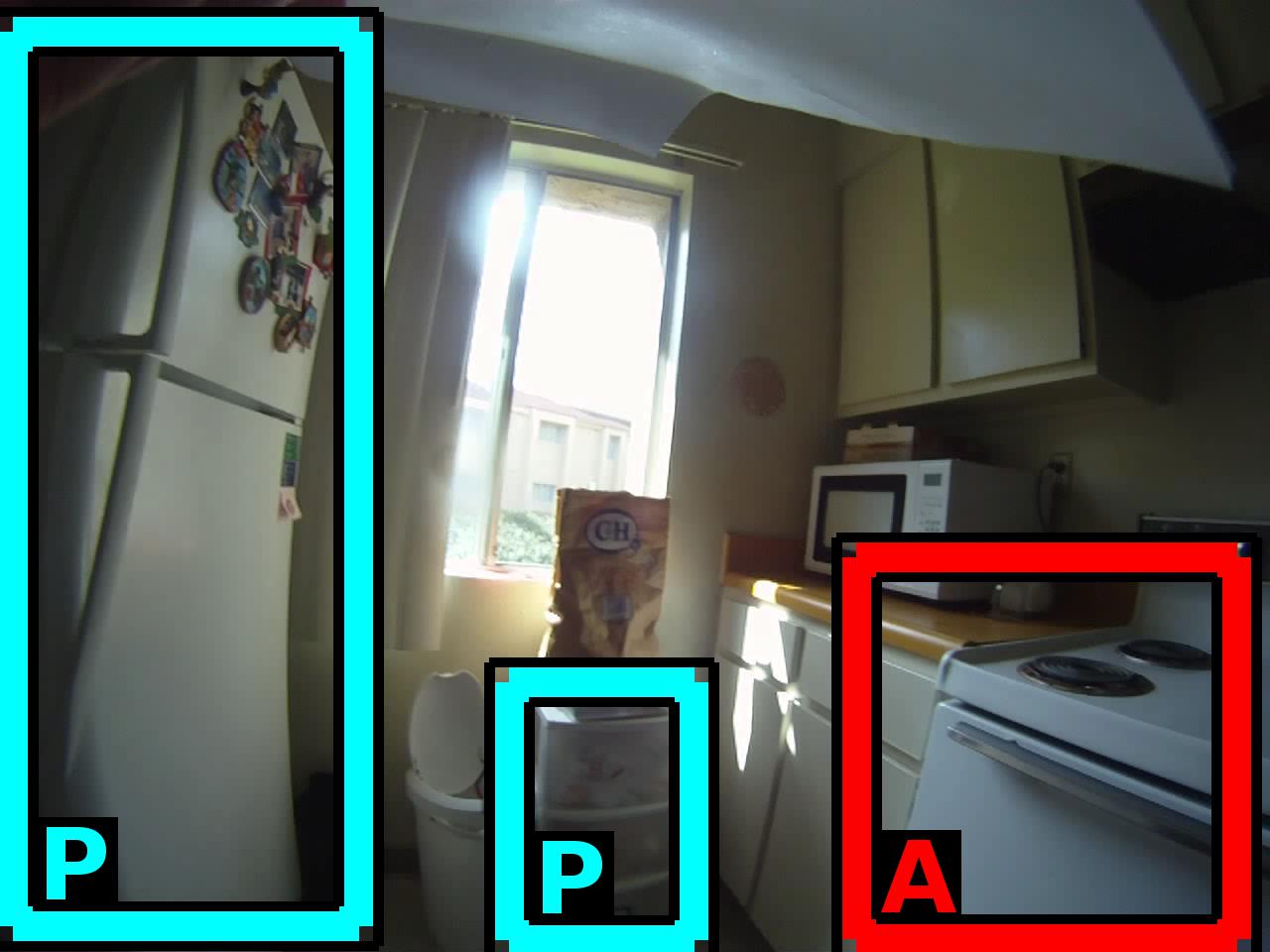} \hfill
	\includegraphics[width=0.24\linewidth]{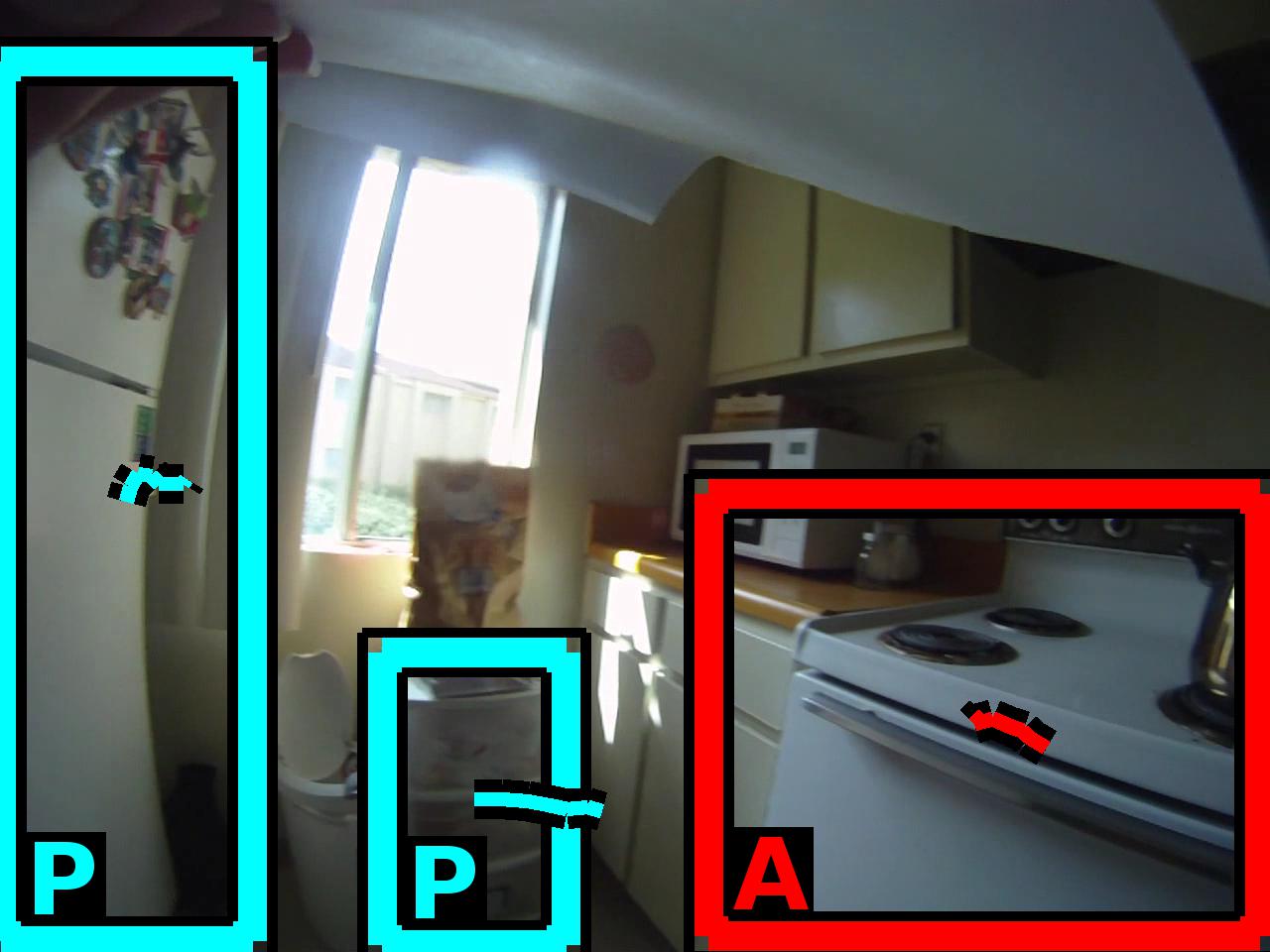} \hfill
	\includegraphics[width=0.24\linewidth]{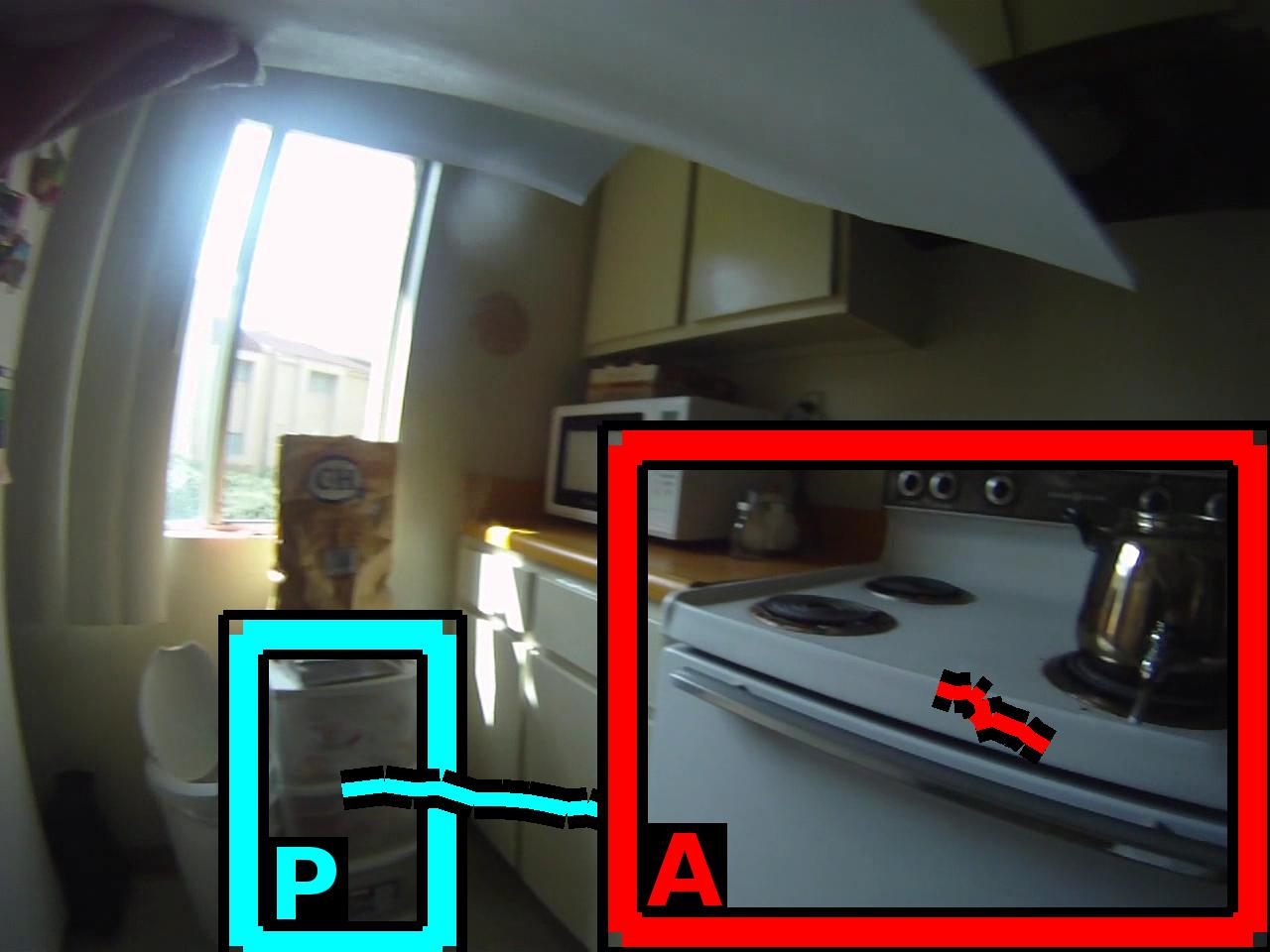} \hfill
	\includegraphics[width=0.24\linewidth]{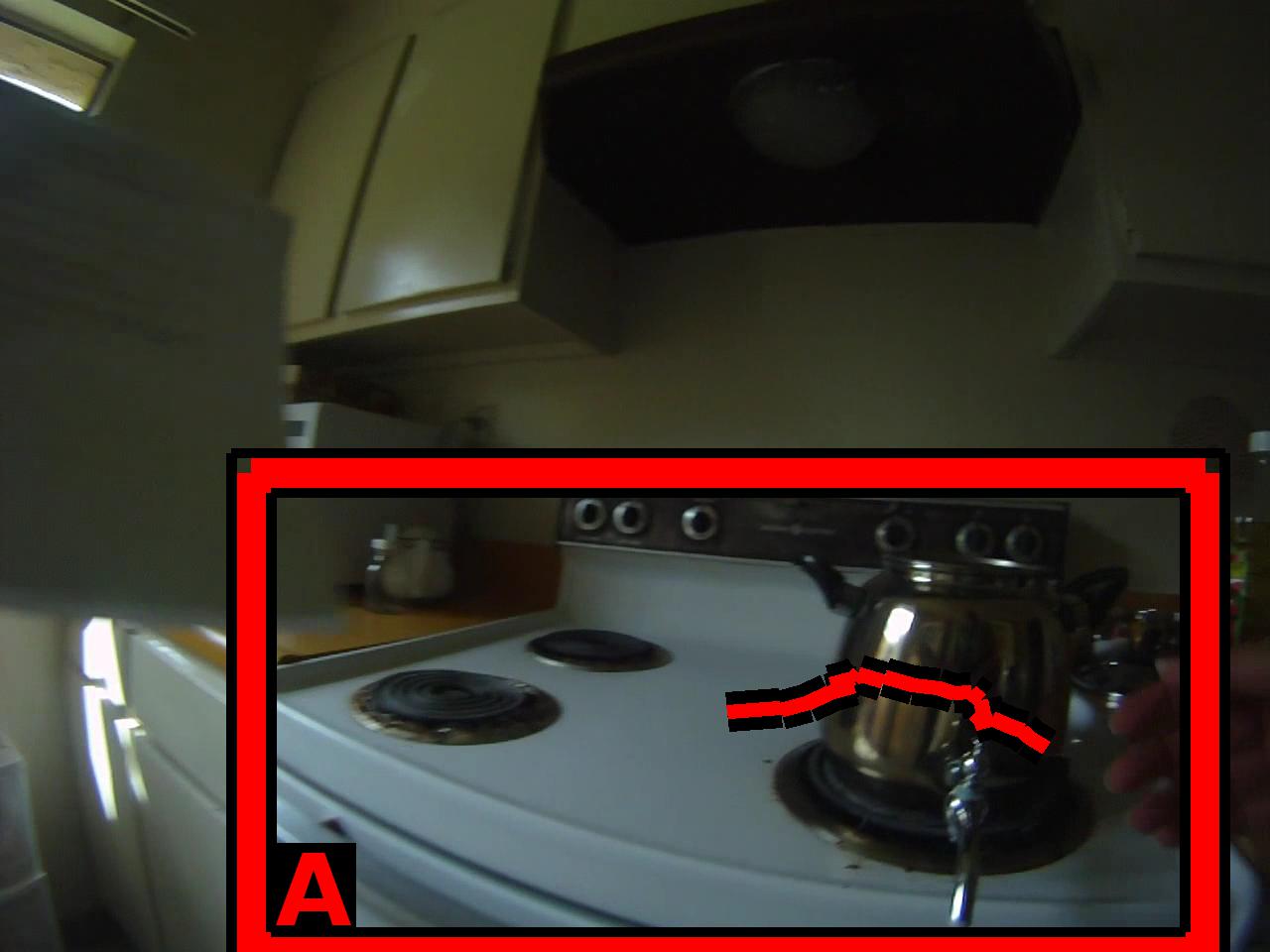}
	
	\vspace{0.5mm}
	\includegraphics[width=0.24\linewidth]{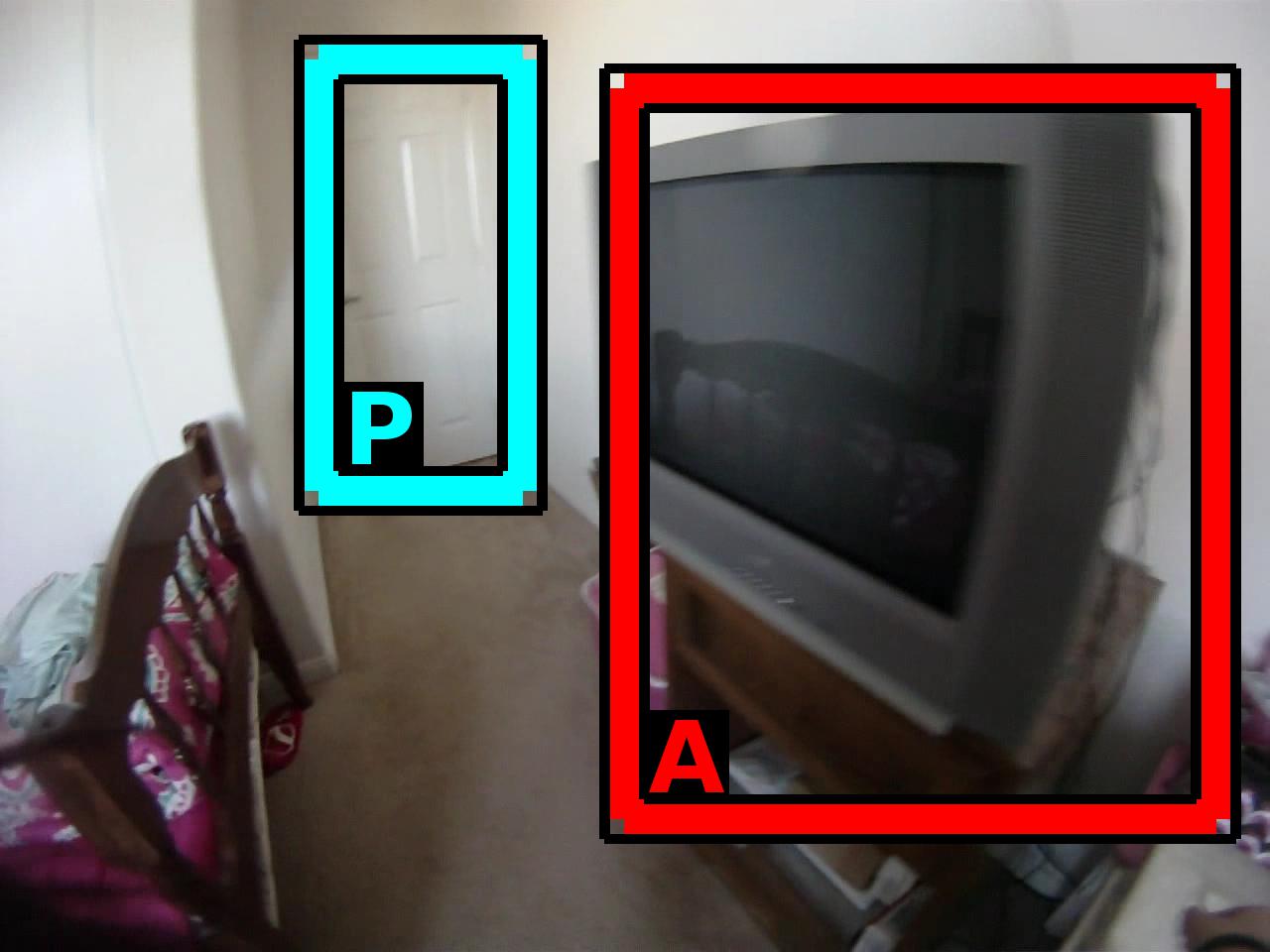} \hfill
	\includegraphics[width=0.24\linewidth]{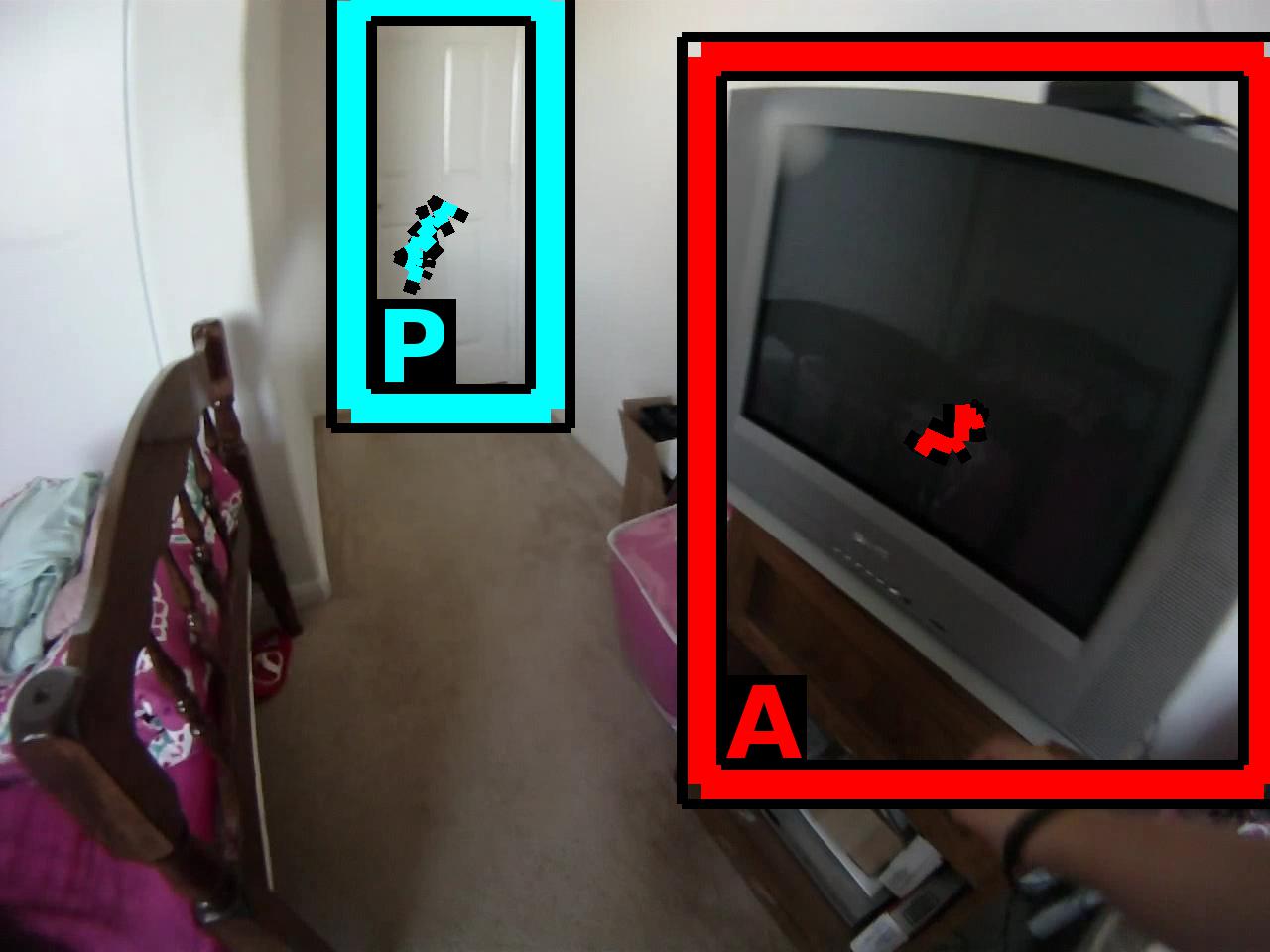} \hfill
	\includegraphics[width=0.24\linewidth]{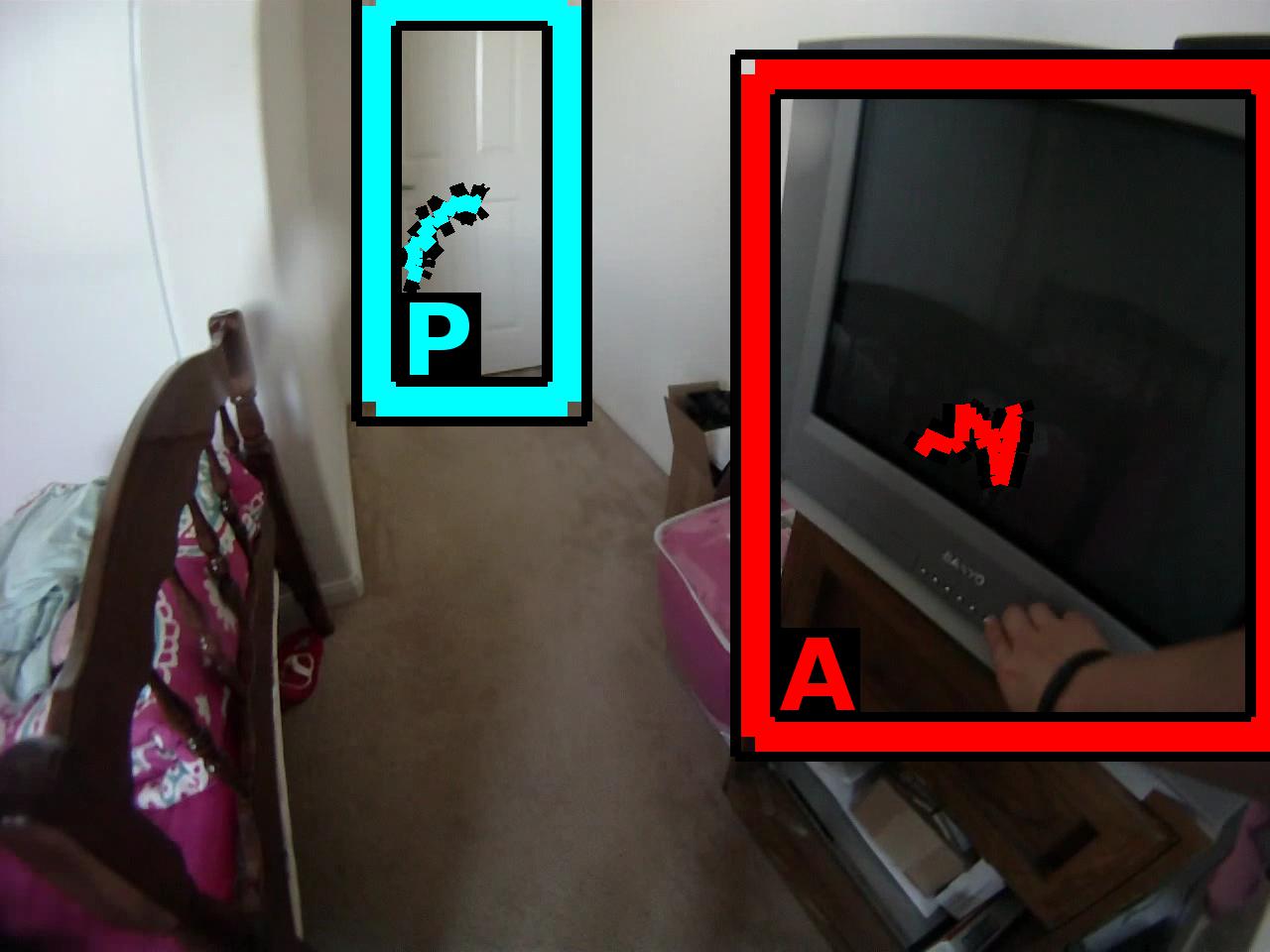} \hfill
	\includegraphics[width=0.24\linewidth]{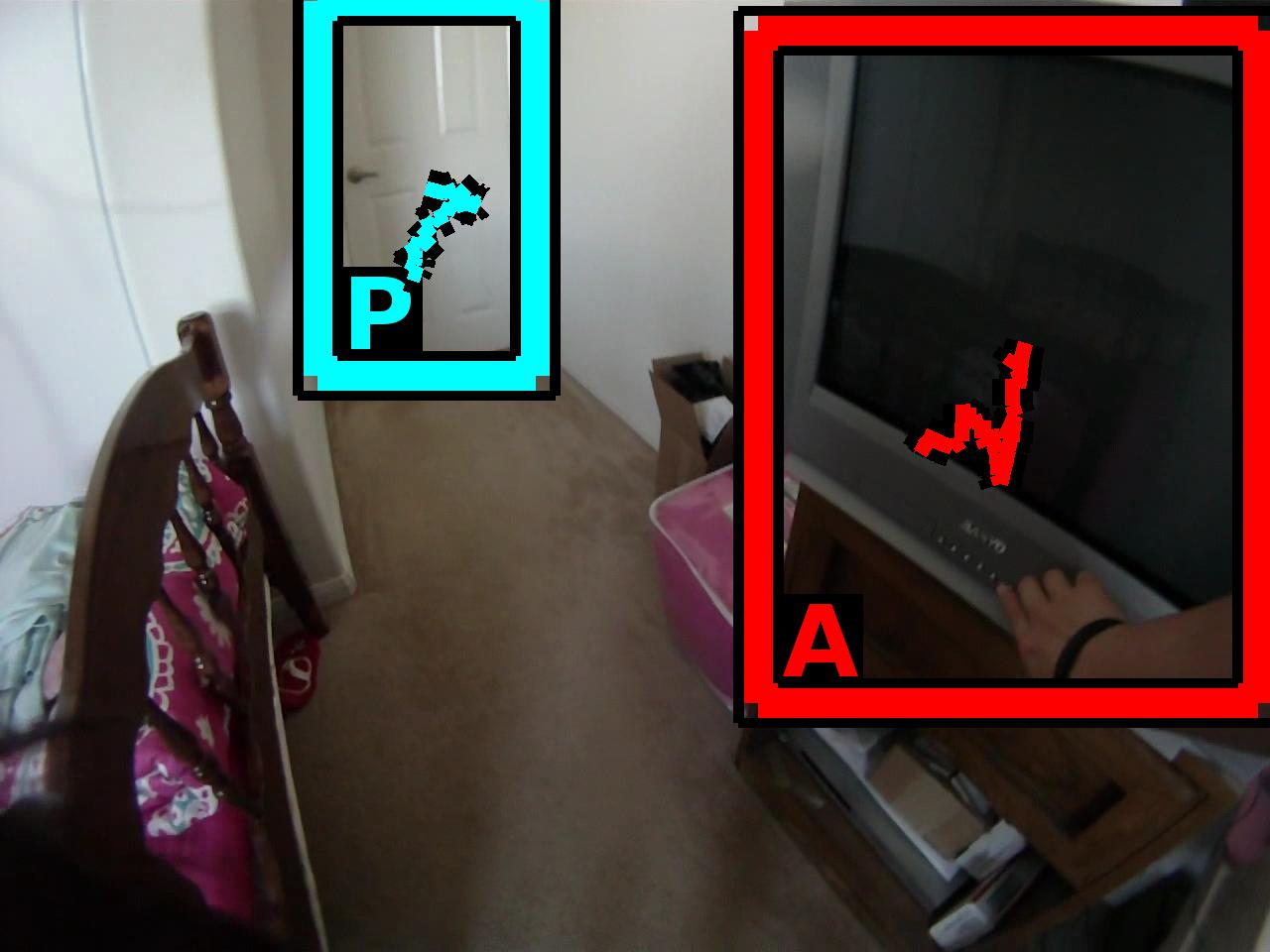}
	
	\vspace{0.5mm}
	\includegraphics[width=0.24\linewidth]{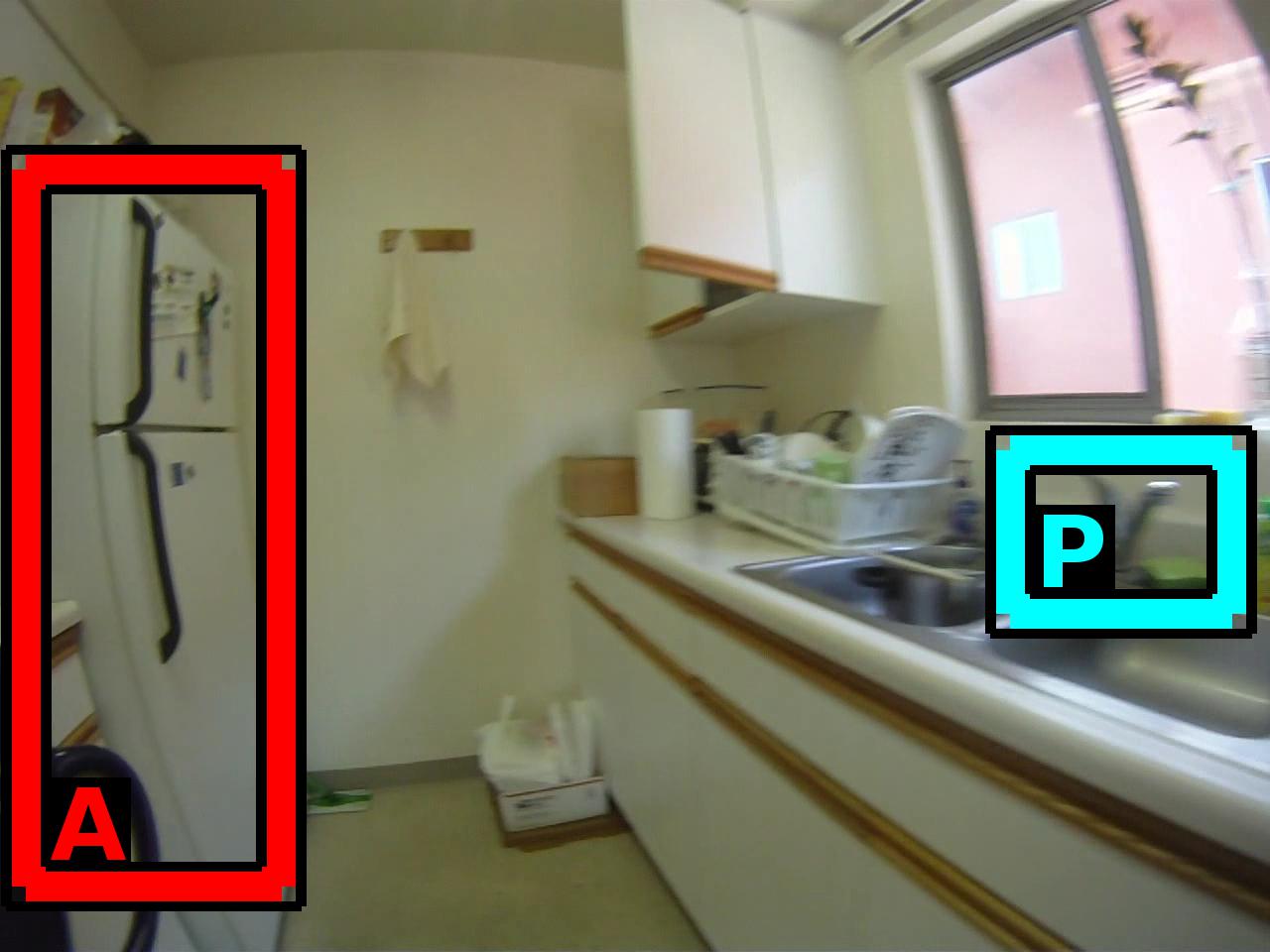} \hfill
	\includegraphics[width=0.24\linewidth]{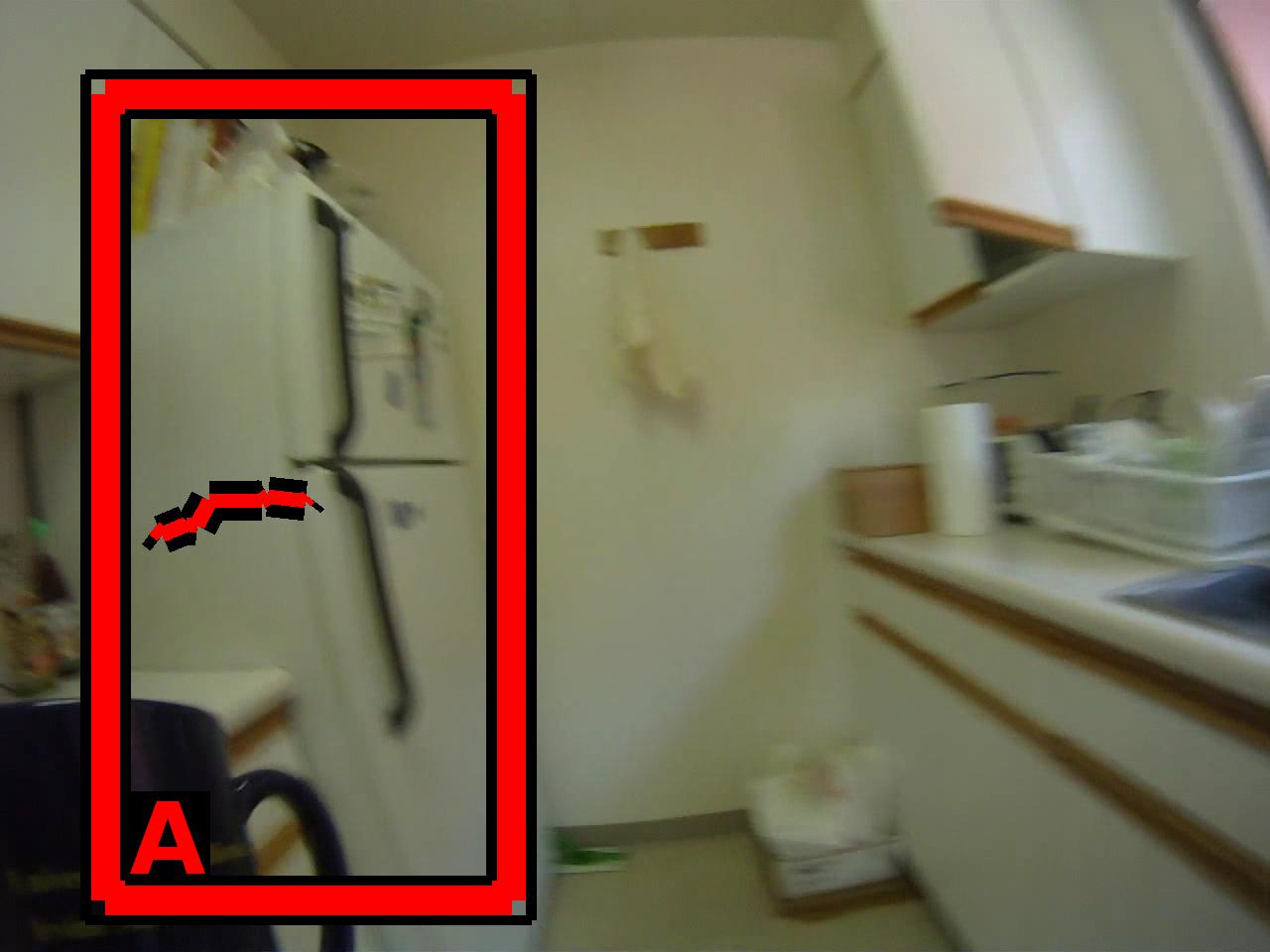} \hfill
	\includegraphics[width=0.24\linewidth]{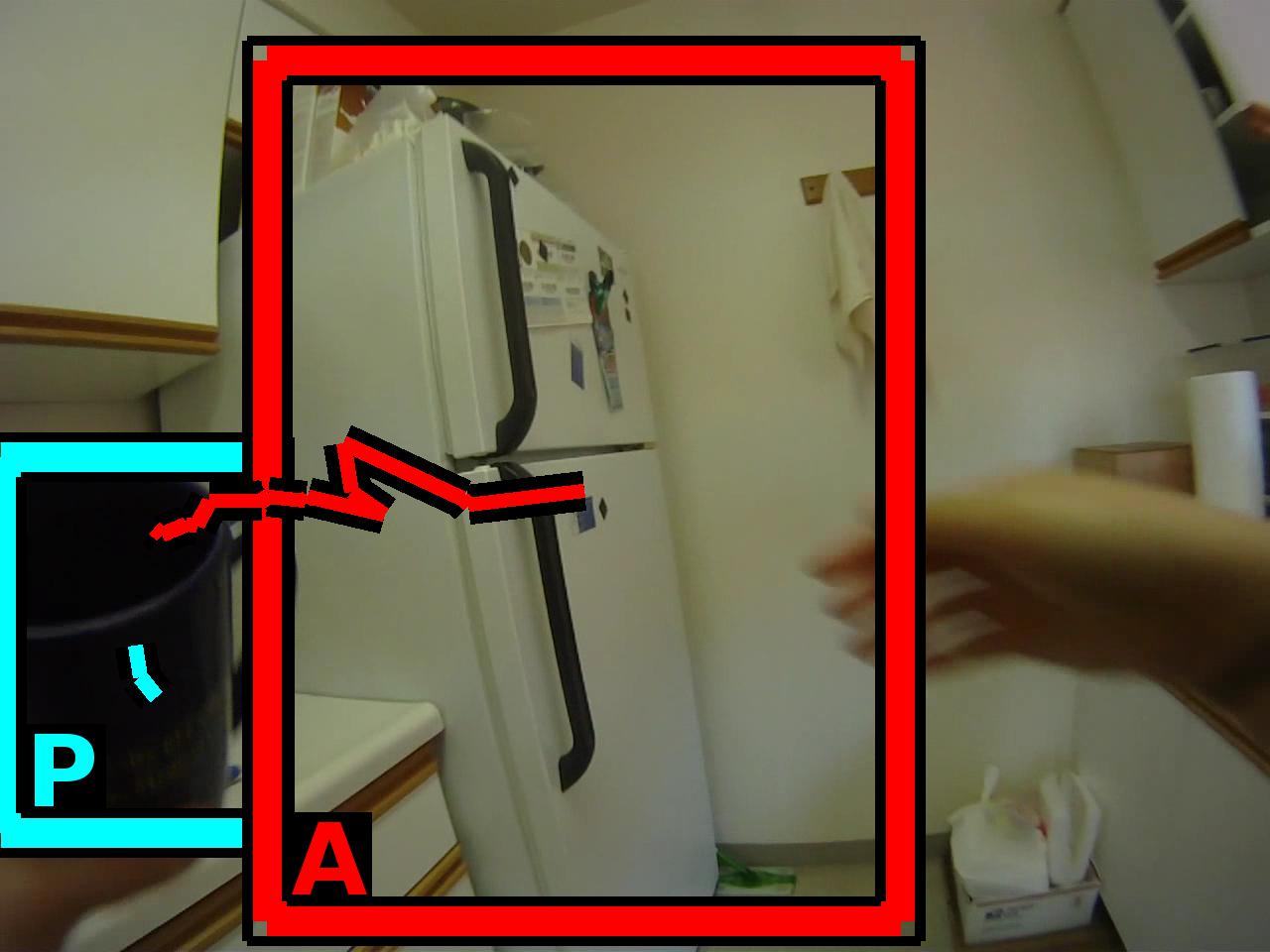}
	\includegraphics[width=0.24\linewidth]{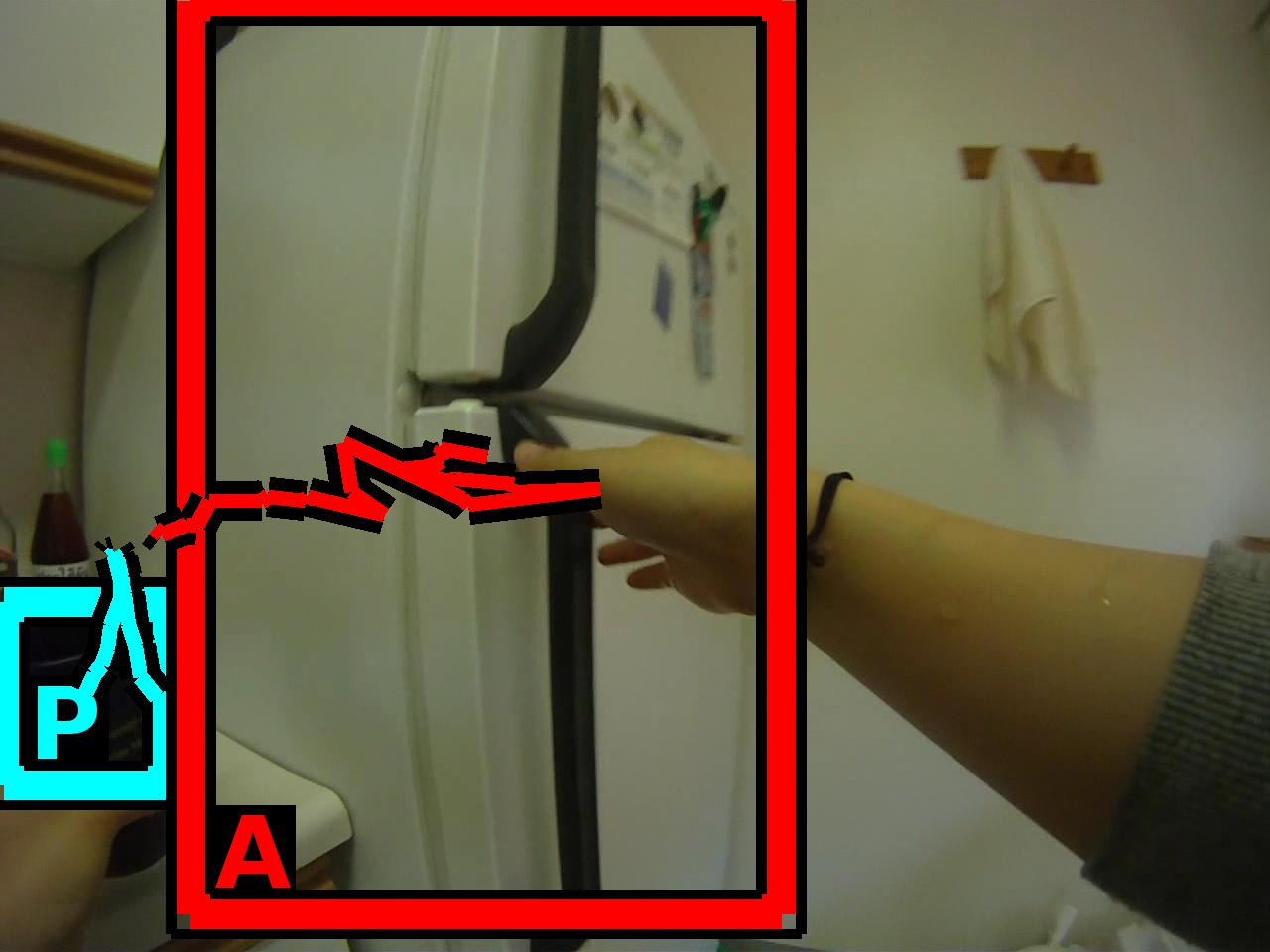}
	
	\caption{Three sequences illustrating next-active-objects (in red, indicated by ``A'') and passive ones (in cyan, indicated by ``P'') along with their trajectories. In each sequence, the dynamics of the scene suggest which objects are going to become active.}
	\label{fig:next_active_object_prediction}
\end{figure}

We argue that the FPV paradigm can provide important cues related to the dynamics of the user with respect to the objects present in the scene. 
Our main hypothesis is that, when a user is performing a specific task, the way he moves and interacts with the environment is influenced by his goals and intended interactions with objects. According to this assumption, in an egocentric scenario, the relative motion of an object in the frame will vary depending on whether the user is planning to interact with that object or not. For instance, the user is expected to move towards an object before interacting with it.
\figurename{~\ref{fig:next_active_object_prediction}} shows three sequences illustrating next-active-objects (in red) and passive ones (in cyan) along with their egocentric object trajectories.\footnote{The reader is also referred to the demo videos available at our web page for some examples of next-active-object prediction: \url{http://iplab.dmi.unict.it/NextActiveObjectprediction/}.} Our hypothesis is that the shape of trajectories, as well as the positions in which they occur in the frame can help to predict next-active-objects discriminating them from those that will remain passive.

We investigate the relevance of egocentric object trajectories in the task of next-active-object prediction. Provided that an object detector/tracker is available, we propose to analyze object trajectories observed in a small temporal window to detect next-active-objects \emph{before} the object-interaction is actually started.
We investigate what properties of object motion are most discriminative and the temporal support with respect to which such motion should be analyzed. The proposed method compares favorably with respect to different baselines exploiting other cues such as the distance of objects from the center of the frame~\cite{Pirsiavash2012}, the presence of hands~\cite{fathi2011understanding,Fathi2012,Ma2016going,li2015delving}, changes in the object appearance~\cite{Pirsiavash2012} and the predictability of the user's visual attention~\cite{Damen2015}.

{In short, our work is the first to investigate the topic of next-active-object prediction from First Person videos. We analyze the role of egocentric object motion in anticipating object interactions and propose a suitable evaluation protocol.}

{The remainder of the paper is organized as follows. Section}~\ref{sec:prediction_related} {reviews the related work. Section}~\ref{sec:prediction_method} {describes the proposed method. Section}~\ref{sec:predicting_experimental} {presents the experimental settings, whereas Section}~\ref{sec:prediction_results} {discusses the results. Finally, Section}~\ref{sec:prediction_conclusion} {concludes the paper.}

\section{Related Work}
\label{sec:prediction_related}
{Our work is related to previous investigations covering different topics. In the following, we review four main research lines: Activity Recognition in First Person Vision (Section}~\ref{sec:activity_fpv}{), Future prediction in Third Person Vision~(Section~}\ref{sec:future_tpv}{), Future prediction in First Person Vision (Section~}\ref{sec:future_fpv}{) and Active Objects~(Section~}\ref{sec:active_objects}{).}

\subsection{Activity Recognition in First Person Vision}
\label{sec:activity_fpv}
Activity recognition from egocentric videos is an active area of research. Through the years, many approaches have been proposed to leverage specific egocentric cues. Spriggs et al.~\cite{Spriggs2009} proposed to use Inertial Measurement Units~(IMU) and a wearable camera to perform activity classification and to segment the video into specific actions. Kitani et al.~\cite{Kitani2011} addressed the problem of discovering egocentric action categories from first person sports videos in an unsupervised scenario. Fathi et al.~\cite{fathi2011understanding} proposed to analyze egocentric activities to jointly infer activities, hands and objects. Fathi et al.~\cite{Fathi2012} concentrated on activities requiring eye-hand coordination and proposed to predict graze sequences and action labels jointly. Pirsiavash and Ramanan~\cite{Pirsiavash2012} investigated an object-centric representation for recognizing daily activities from first person camera views. McCandless and Grauman~\cite{McCandless2013} proposed to learn the spatio-temporal partitions which were most discriminative for a set of egocentric activities. Ryoo and Matthies~\cite{Ryoo2013} considered videos acquired from a robot-centric perspective and proposed to recognize egocentric activities performed by other subjects while interacting with the robot. Li et al.~\cite{li2015delving} proposed a benchmark of different egocentric cues for action recognition. 
The authors of~\cite{Ma2016going,Zhou2016} proposed to integrate different egocentric cues to recognize activities using deep learning. The aforementioned works assume that the activities can be fully observed before performing the recognition process and do not concentrate on future prediction from the observed data.

\subsection{Future prediction in Third Person Vision}
\label{sec:future_tpv}
Previous works have investigated the problem of early action recognition and future action prediction from a standard third person perspective. The considered application scenarios range from video surveillance to human-robot interaction. Ryoo~\cite{Ryoo2011} proposed a method to recognize ongoing activities from streaming videos. Huang et al.~\cite{huang2014sequential} introduced a system which copes with the ambiguity of partial observations by sequentially discarding classes until only one class is identified as the detected one. Hoai and De La Torre~\cite{Hoai2014} exploited Structured Output SVM to recognize partial events and enable early recognition. 
Kong and Fu~\cite{Kong2016max} designed compositional kernels to hierarchically capture the relationship between partial observations. Ma et al.~\cite{Ma2016learning} investigated a method to improve training of temporal deep models to learn activity progression for activity detection and {``early''} recognition tasks.

Beyond early action recognition, other methods have concentrated on the forecasting of future actions before they actually occur. In particular, Kitani~\cite{Kitani2012} modeled the effect of the physical environment on the choice of human actions in the scenario of trajectory-based activity analysis from visual input. Koppula et al.~\cite{Koppula2013} studied how to enable robots to anticipate human-object interactions from visual input in order to provide adequate assistance to the user. Lan et al.ì\cite{lan2014hierarchical} exploited a hierarchical representation of human movements to infer future actions from a still image or a short video clip. Vondrick et al.~\cite{Vondrick2016} proposed to predict future image representations in order to forecast human actions from video.

{Unlike our approach}, such works do not consider egocentric scenarios{. However,} the main motivation behind them is related to ours: \emph{building systems which are able to recognize ongoing events from partial observations and react in a timely way.}

\subsection{Future prediction in First Person Vision}
\label{sec:future_fpv}
Future prediction has been investigated also in the first person vision domain. The main application scenario related to such works concerns user assistance and aiding human-machine interaction. Zhou et al.~\cite{Zhou2015} concentrated on the task of inferring temporal ordering from egocentric videos. Singh et al.~\cite{singh2016krishna} and Park et al.~\cite{SooPark2016} presented methods to predict future human trajectories from egocentric images. Soran et al.~\cite{Soran2015} proposed a system which analyzes complex activities and notifies the user when he forgets to perform an important action. Su and Grauman~\cite{su2016leaving} proposed to predict the next object detector to run on streaming videos to perform activity recognition. Ryoo et al.~\cite{Ryoo2015a} proposed a method for early detection of actions performed by humans on a robot from a first person, robot-centric perspective. Vondrick et al.~\cite{Vondrick2016} proposed to forecast the presence of objects in egocentric videos from anticipated visual representations. 

Our investigation is related to this line of works but, rather than considering prediction at the activity level, we focus on the granularity of user-object interaction and exploit the information provided by object motion dynamics in egocentric videos. {Object-level forecasting is important to develop systems able to timely respond to the user behavior and assist him properly.}

\subsection{Active Objects}
\label{sec:active_objects}
Our interest in next-active-object prediction has also been fostered by the importance of active objects in tasks such as egocentric activity recognition.
In particular, Pirsiavash and Ramanan~\cite{Pirsiavash2012} proposed to distinguish active objects from passive ones. Active objects are objects being manipulated by the user and provide important information about the action being performed (e.g., using the kettle to boil water). Passive objects are non-manipulated objects and provide context information (e.g., a room with a fridge and a stove is probably a kitchen). The primary assumption made by Pirsiavash and Ramanan~\cite{Pirsiavash2012} is that active and passive objects can be discriminated by their appearance (e.g., an active fridge is probably open and looks different from a passive one) and the position in which they appear in the frame (i.e., active objects tend to appear near the center). Active objects have also been been considered in recent research on egocentric activity recognition. Fathi et al.~\cite{fathi2011understanding} suggested to pay special attention to objects manipulated by hands for egocentric activity recognition. Li et al.~\cite{li2015delving} used Improved Dense Trajectories to extract features from the objects the user is interacting with. Ma et al.~\cite{Ma2016going} designed a deep learning framework which integrates different egocentric cues including optical flow, hand segmentation and objects of interest for egocentric activity recognition. Zhou et al.~\cite{Zhou2016} presented a cascade neural network to collaboratively infer the hand segmentation maps and manipulated foreground objects. 

The general idea that some objects are more important than others has been investigated also in other scenarios related to First Person Vision. Lee and Grauman~\cite{lee2015predicting} designed methods to summarize egocentric video by predicting important objects the user interacts with during the day. Bertasius et al.~\cite{Bertasius2016} designed a method for detecting action-objects (i.e., objects associated with seeing and touching actions). Damen et al.~\cite{Damen2015} proposed an unsupervised approach to detect task-relevant objects and provide gaze-triggered video guidance when the user intends to interact with the object. 

Differently than the aforementioned works, we investigate how next-active-objects can be correctly recognized from egocentric video. The prediction requirement, i.e., perform recognition of active objects \emph{before} the interaction begins, makes less effective the exploitation of some cues such as object appearance and the presence of hands, which have been generally used to address active object recognition in the past.

\section{Method}
\label{sec:prediction_method}

We propose to predict next-active-objects from egocentric videos by analyzing egocentric object trajectories. We assume that an object detector trained on a set of $N$ object categories is available. A tracker is used to associate detections related to the same object instance in order to generate object tracks. At each time step, the system analyzes the trajectories observed in recent frames in order to recognize next-active-objects before an interaction actually takes place.

\subsection{Object Tracks}
\label{sec:object_tracks}

We consider an object track as a sequence of bounding boxes across subsequent frames of a video. All bounding boxes are related to the same object instance. 
We follow~\cite{Pirsiavash2012}, where \emph{active objects} are defined as \emph{objects undergoing hand manipulations}.
Therefore, each \emph{bounding box} is labeled as ``active'' if the user manipulates it at that moment or ``passive'' otherwise. 
Bounding boxes $b \in \Re^4$ are represented by the four coordinates of the top-left and bottom-right corners. To generalize over different image sizes and aspect ratios, all coordinates are divided by the frame dimensions in order to be normalized in the interval $[0,1]$. Coordinates are then centered around the normalized center point $(0.5,0.5)$.
\begin{figure}[t]\centering
	\includegraphics[width=\linewidth]{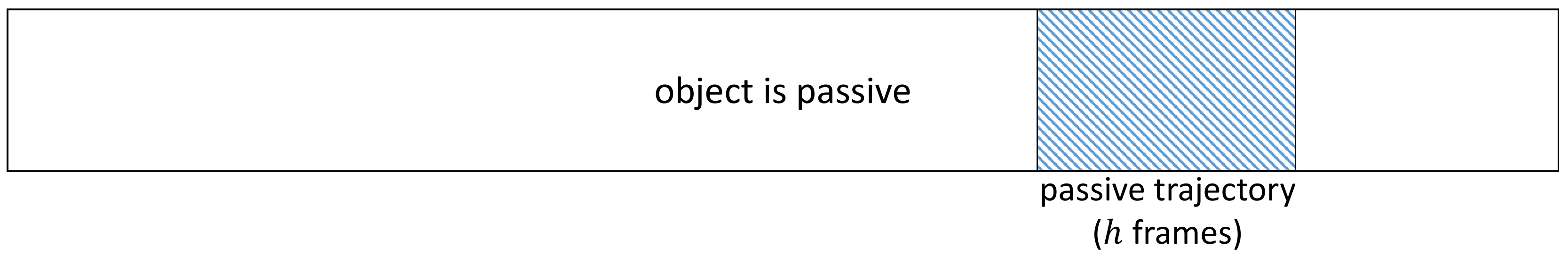}
	\centerline{(a) passive track / passive trajectory}
	
	\includegraphics[width=\linewidth]{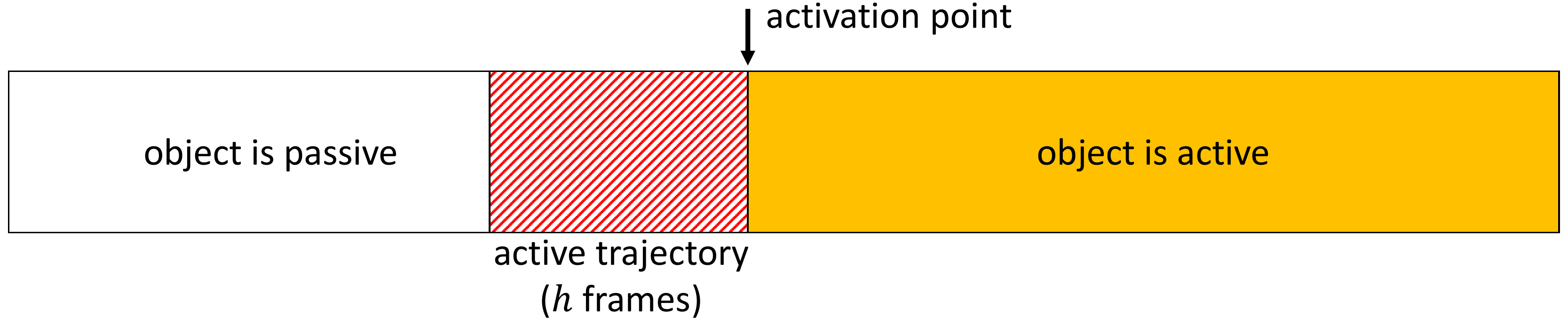}
	\centerline{(b) mixed track / active trajectory}

	\vspace{1mm}
	\caption{An example of \emph{mixed} track (a) and \emph{passive} track (b). The figure also illustrates how active and passive trajectories are extracted from mixed tracks for training purposes (Section~\ref{sec:prediction_object_trajectories}).}
	\label{fig:trajectory_extraction}
\end{figure}

We divide \emph{object tracks} into two categories: \emph{passive} and \emph{mixed}. 
Tracks composed only by passive bounding boxes (i.e., passive objects) are denoted as passive tracks. Tracks containing both passive and active bounding boxes are denoted as \emph{mixed} tracks. In this case, we refer to the point in which an object changes its status from passive to active as {the} ``activation point''. \figurename~\ref{fig:trajectory_extraction} illustrates examples of passive and mixed tracks. Since we are interested in predicting next-active-objects, i.e., objects which are going to change their status from passive to active, we discard all tracks containing only active bounding boxes.

\subsection{Object Trajectories}
\label{sec:prediction_object_trajectories}
At test time, the system should be able to recognize next-active-objects \emph{before} they become active. Hence it can only rely on egocentric object trajectories preceding the activation point. We extract \emph{object trajectories} from the considered \emph{object tracks} and propose to train an active {versus} passive trajectory classifier.

We define an \emph{object trajectory} as a sequence of bounding boxes $T_i=\{b_1,b_2,\ldots,b_h\}$ and consider two classes of \emph{trajectories}: active and passive. Active trajectories are those leading to a change of status from passive to active. Passive trajectories are related to passive objects that will not become active and hence they do not lead to any status change. 

While in principle we would like to predict next-active-objects arbitrarily in advance, we expect that the most discriminative part of active trajectories is the one immediately preceding the status change.
Therefore, in order to train an active vs passive trajectory classifier, we consider fixed length trajectories of $h$-frames. Parameter $h$ has to be chosen to include enough discriminative information while avoiding the noise due to long trajectories including data far away from the activation point. {We discuss specific settings in experimental details below.}

To compose a suitable training set, we extract passive and active trajectories from the object tracks obtained as described in Section~\ref{sec:object_tracks}.
Passive trajectories are randomly sampled from all passive tracks (we extract one trajectory per track).

Active trajectories are sampled from mixed tracks by considering the last $h$ frames preceding each activation point.

\begin{figure}[t]
	\includegraphics[width=0.24\linewidth]{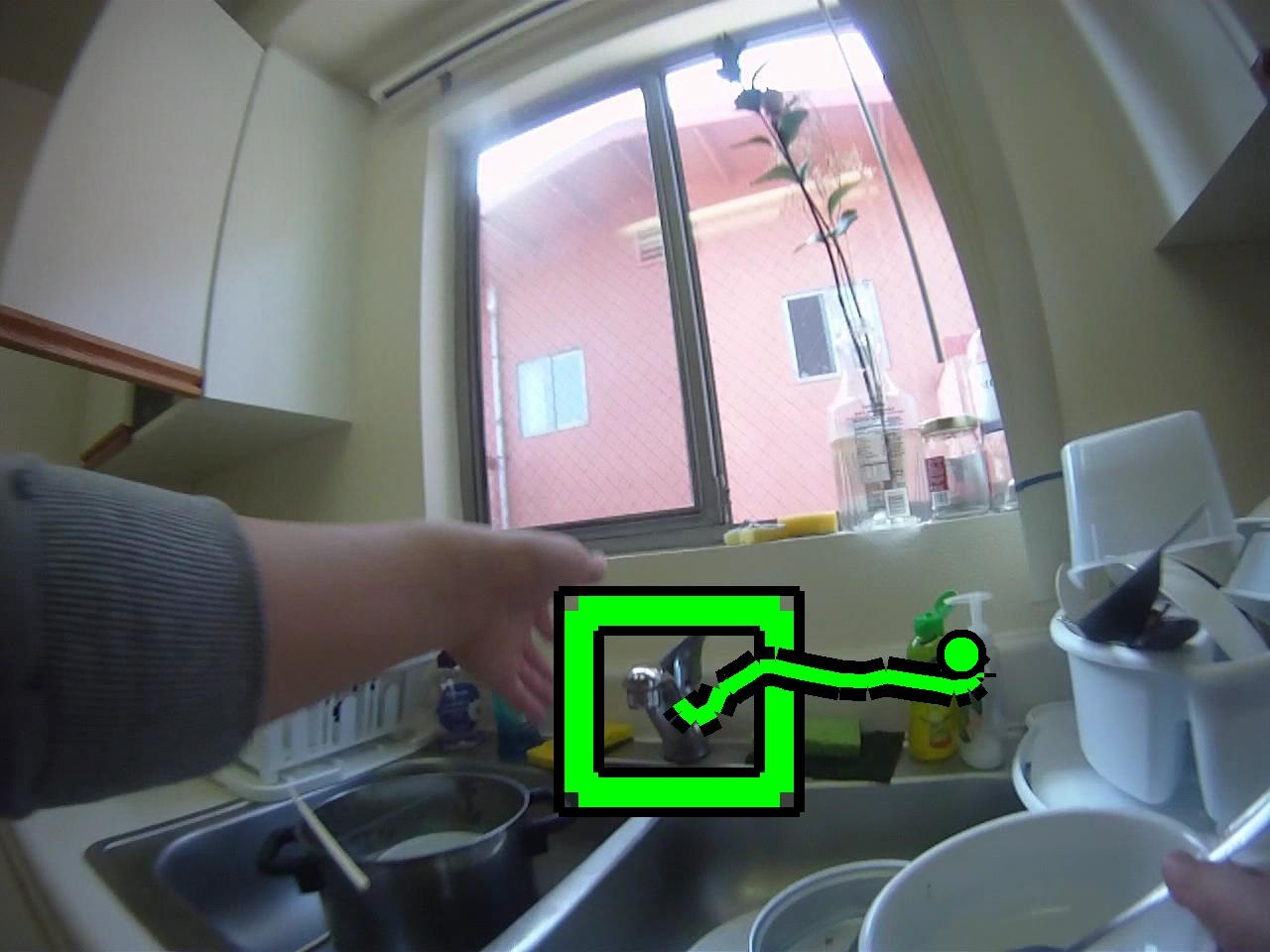}\hfill
	\includegraphics[width=0.24\linewidth]{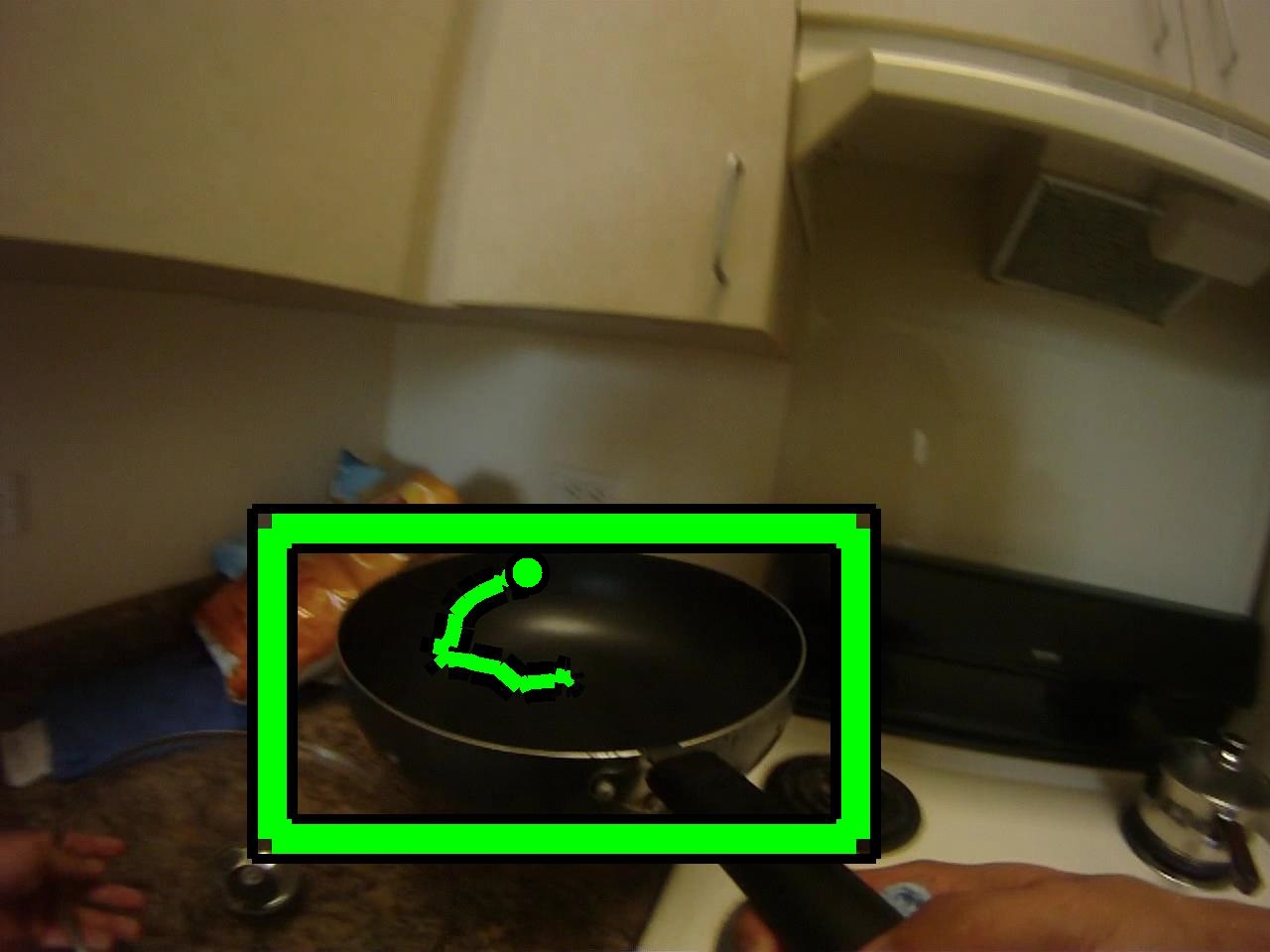}\hfill
	\includegraphics[width=0.24\linewidth]{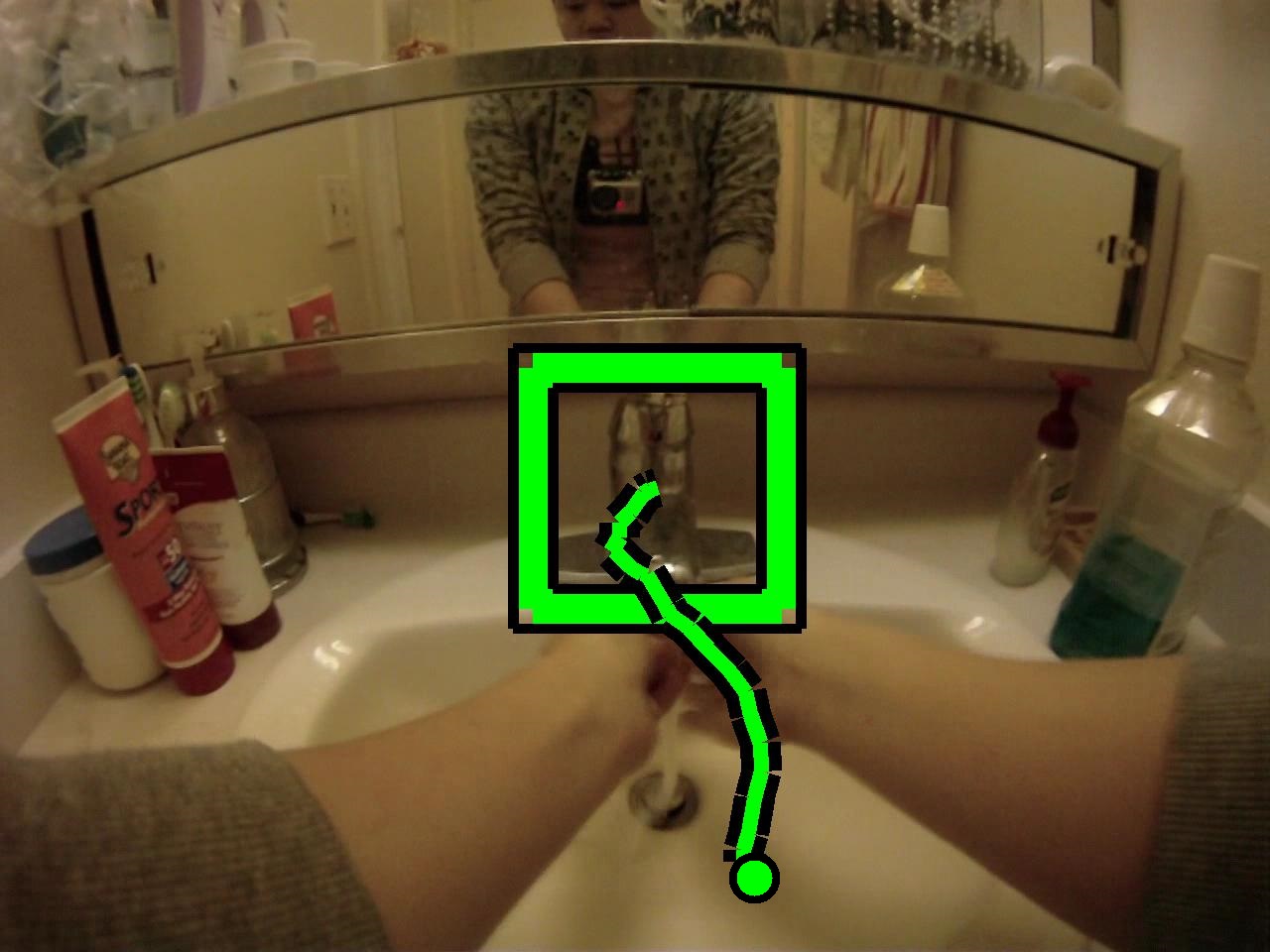}\hfill
	\includegraphics[width=0.24\linewidth]{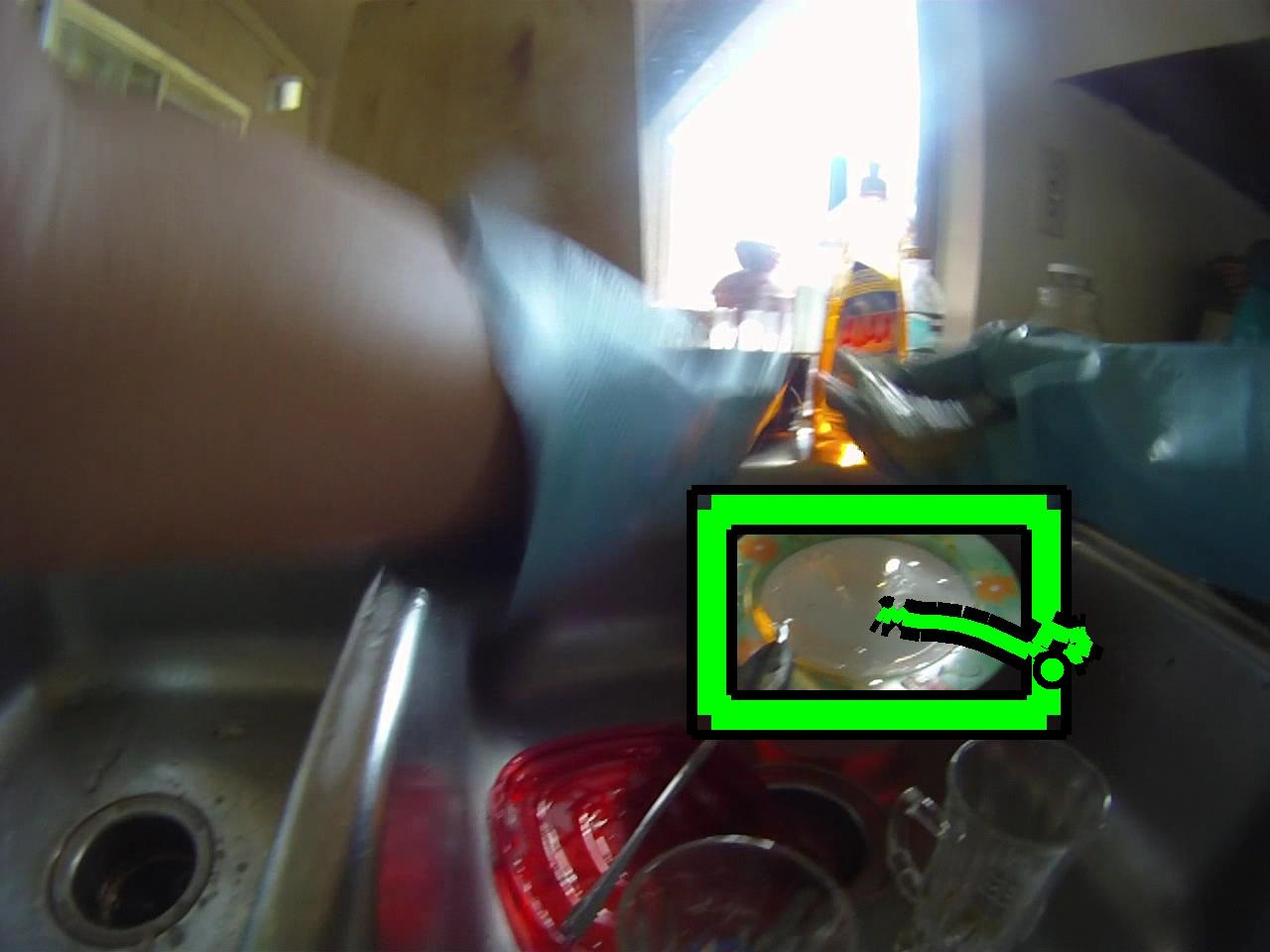}\hfill
	
	\vspace{0.5mm}
	\includegraphics[width=0.24\linewidth]{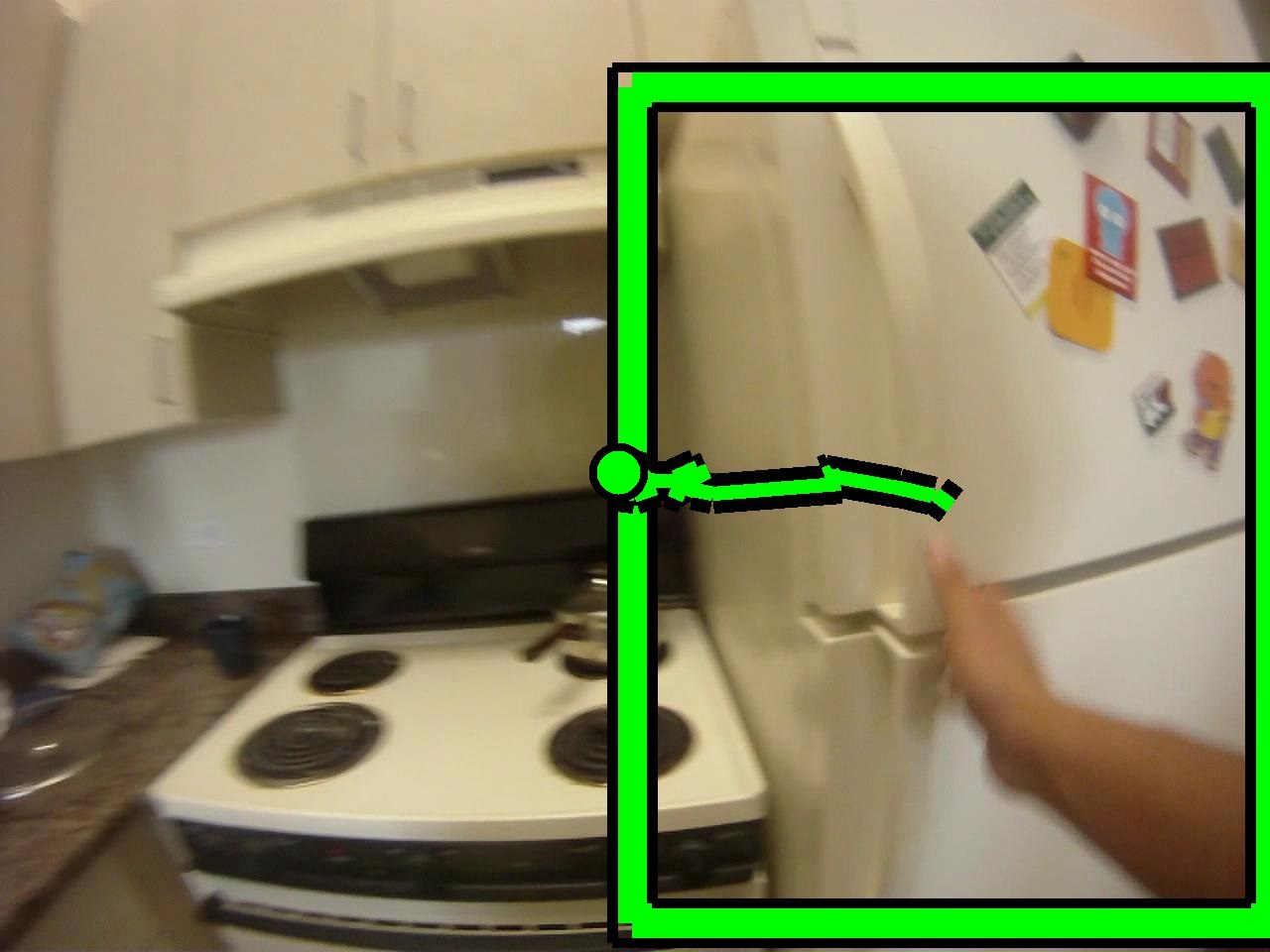}\hfill
	\includegraphics[width=0.24\linewidth]{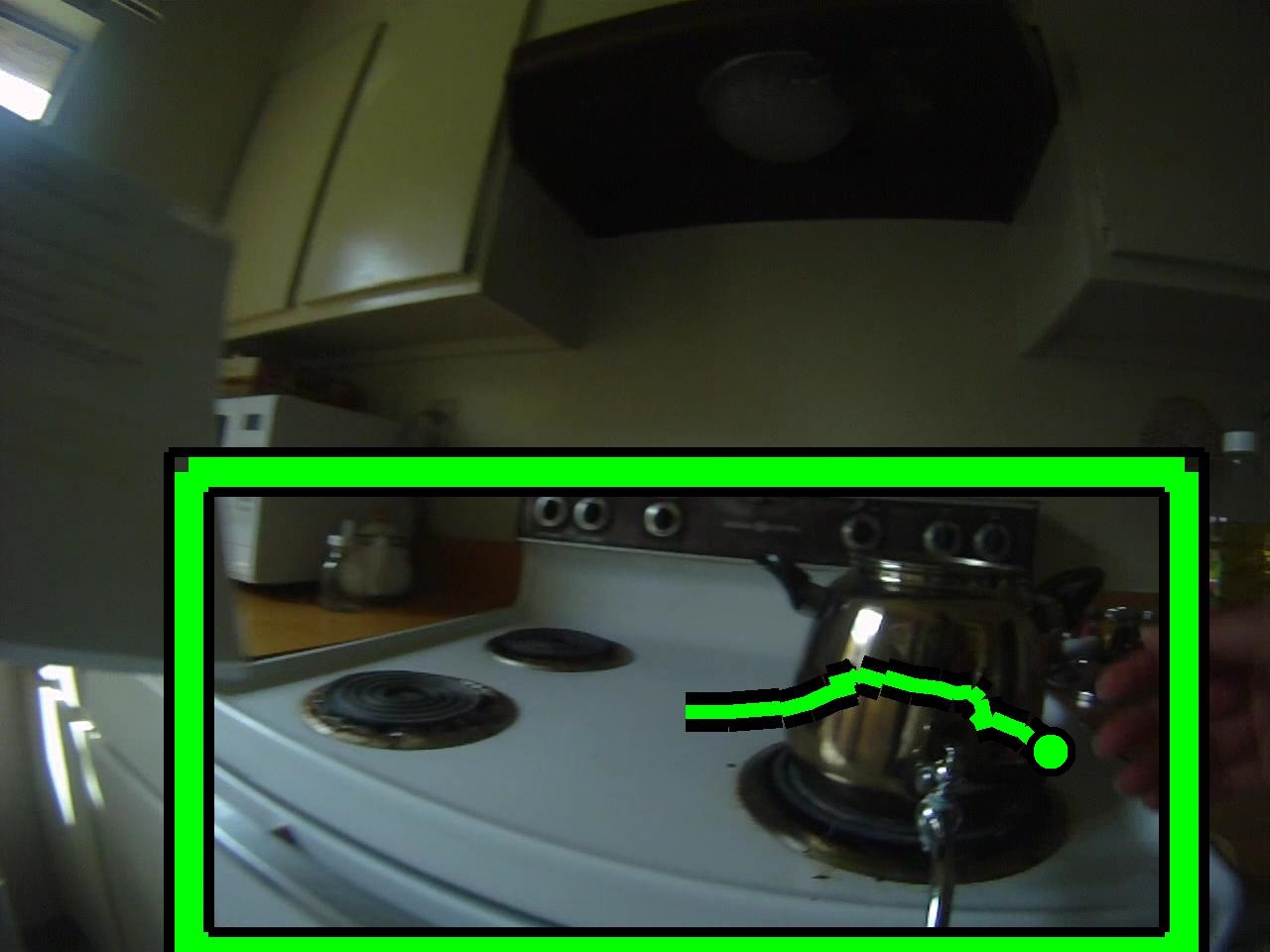}\hfill
	\includegraphics[width=0.24\linewidth]{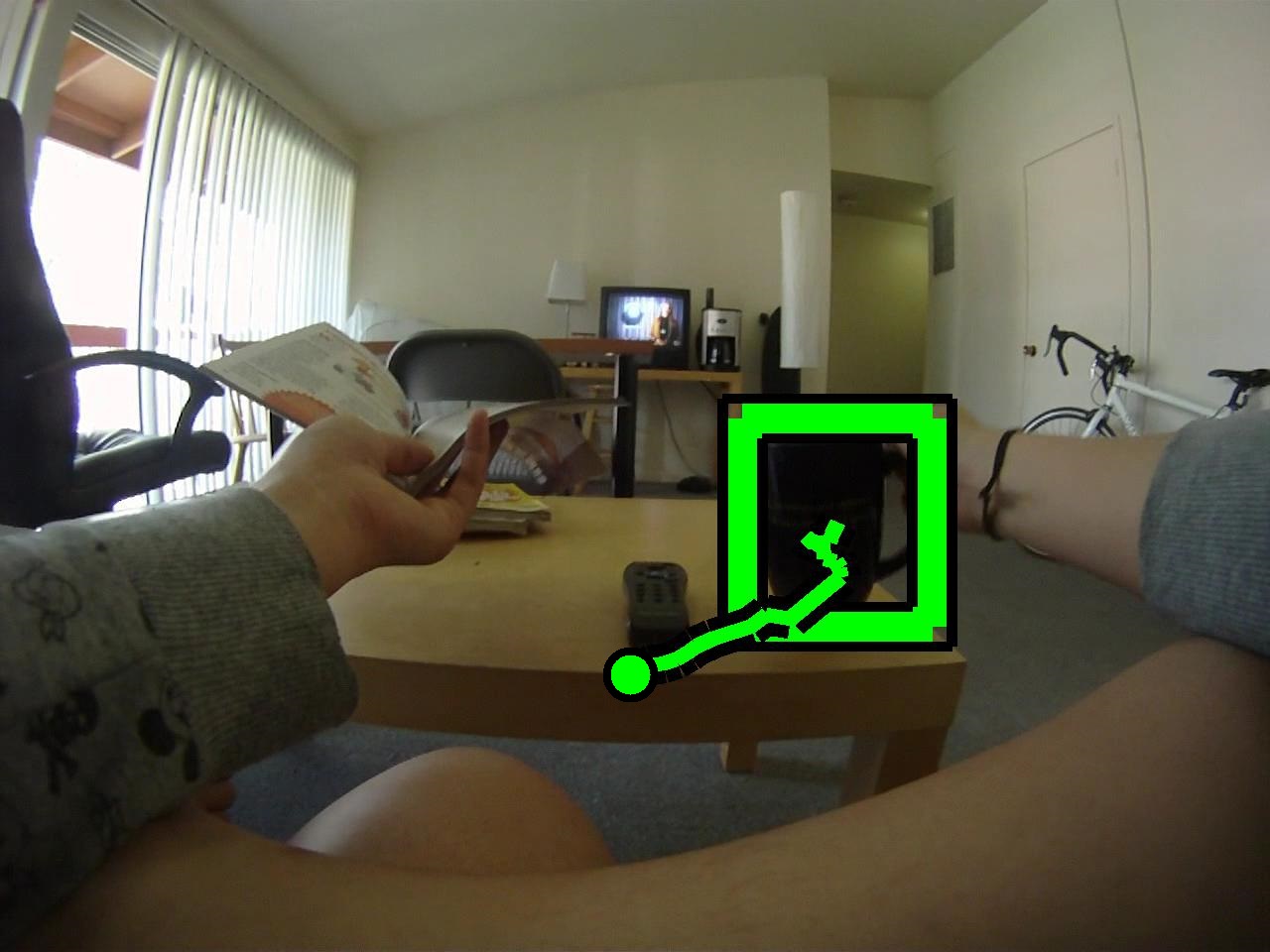}\hfill
	\includegraphics[width=0.24\linewidth]{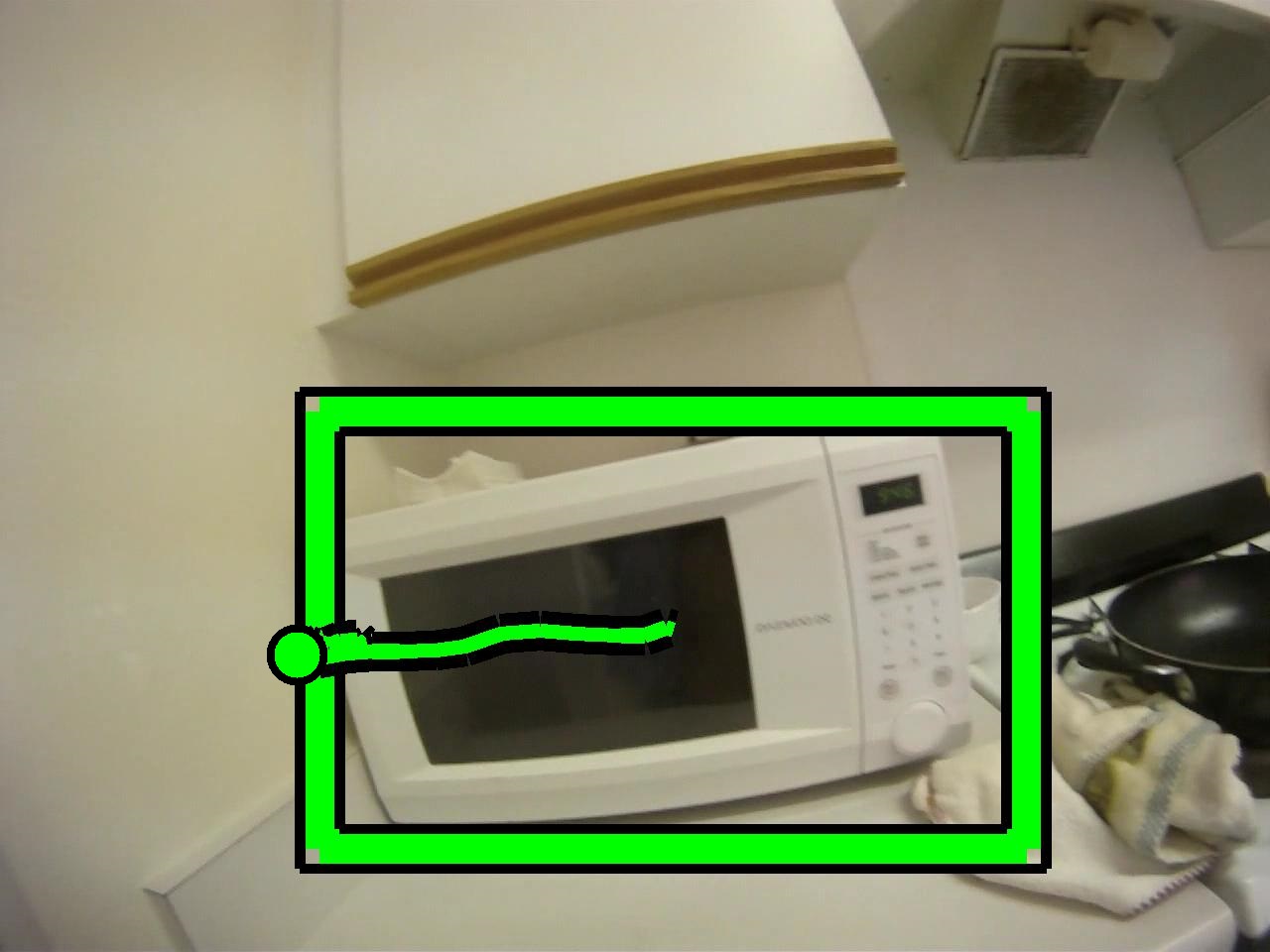}\hfill
	
	\centerline{(a) active trajectories}
	
	\vspace{0.5mm}
	\includegraphics[width=0.24\linewidth]{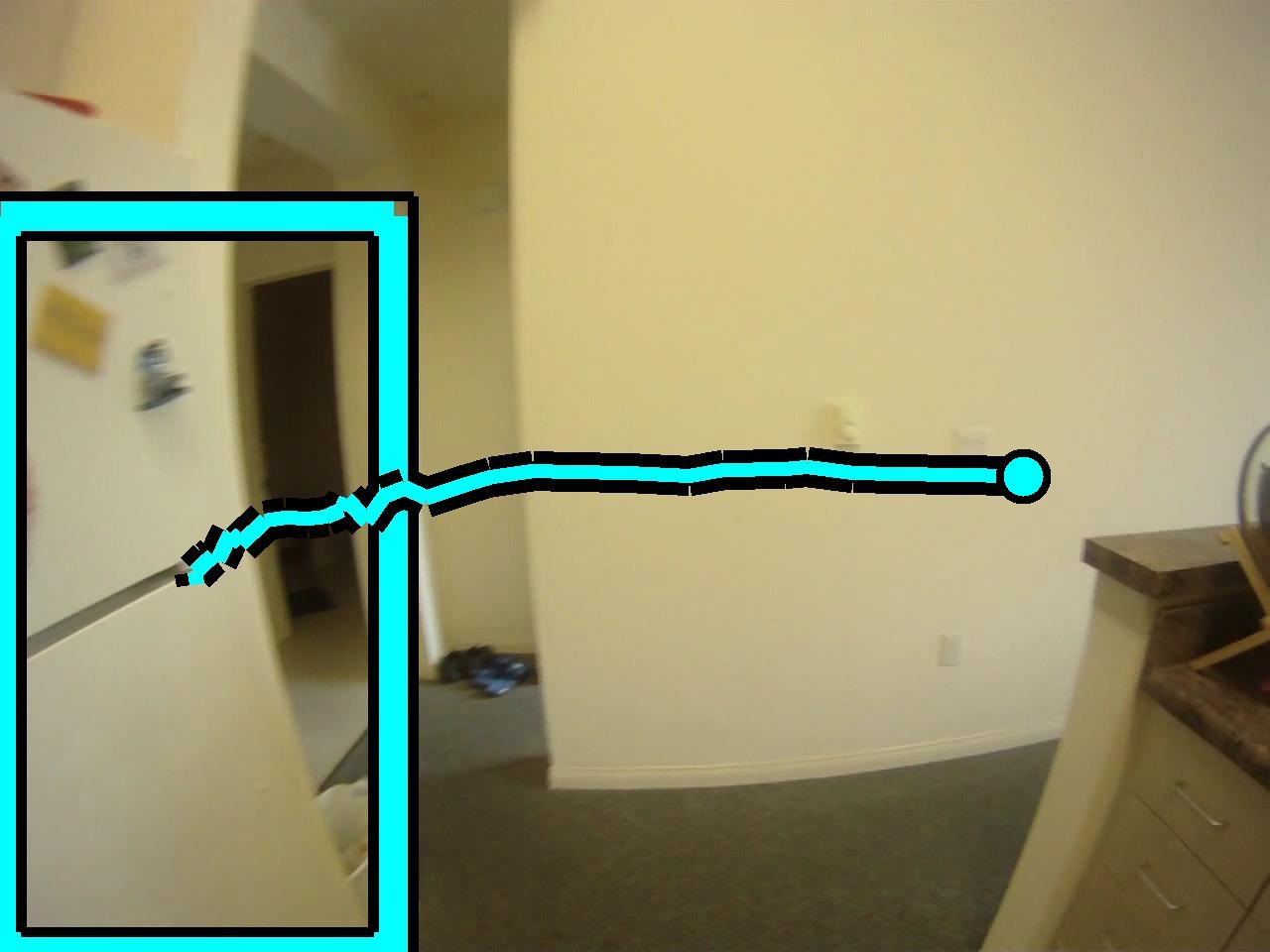}\hfill
	\includegraphics[width=0.24\linewidth]{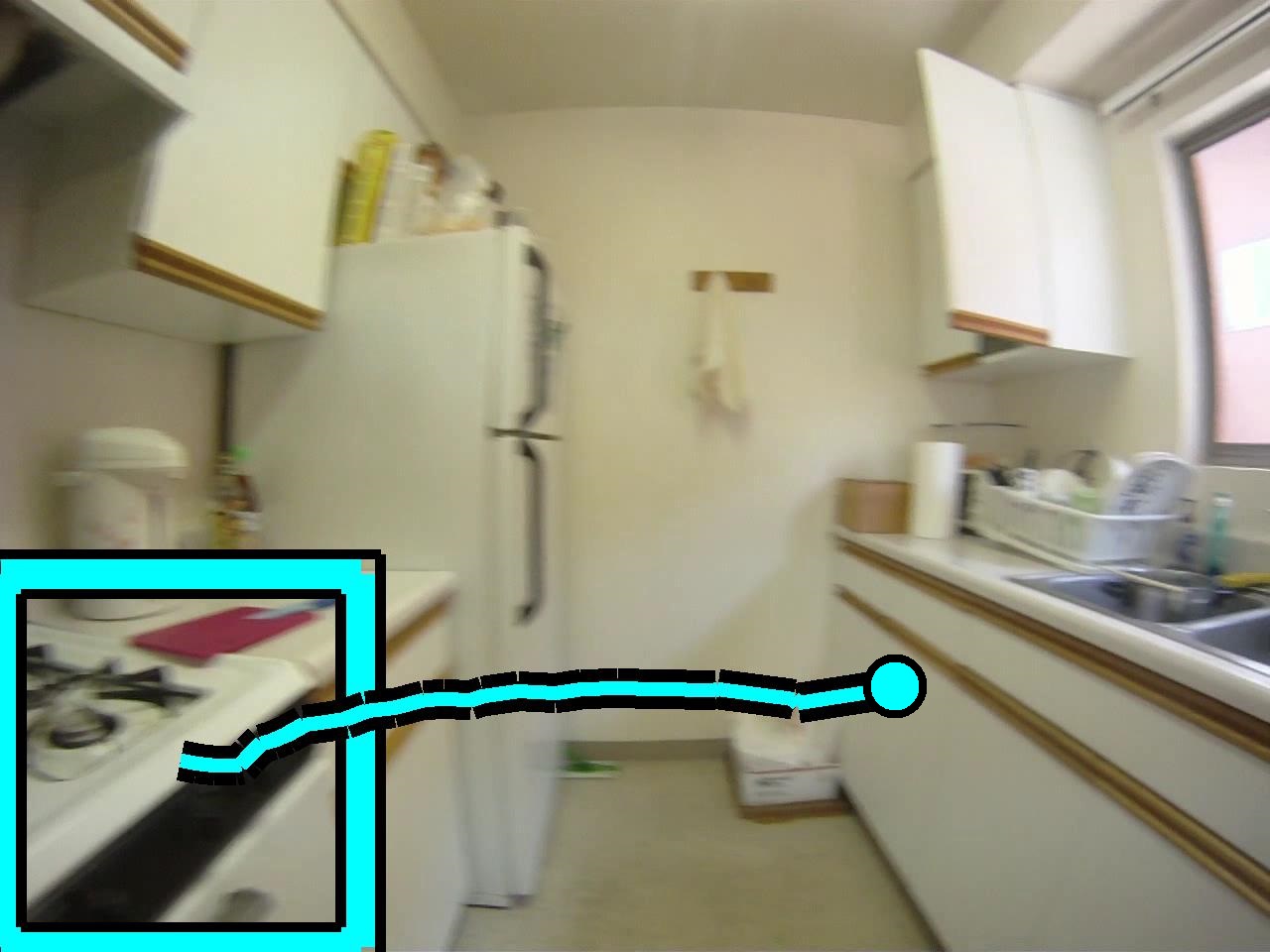}\hfill
	\includegraphics[width=0.24\linewidth]{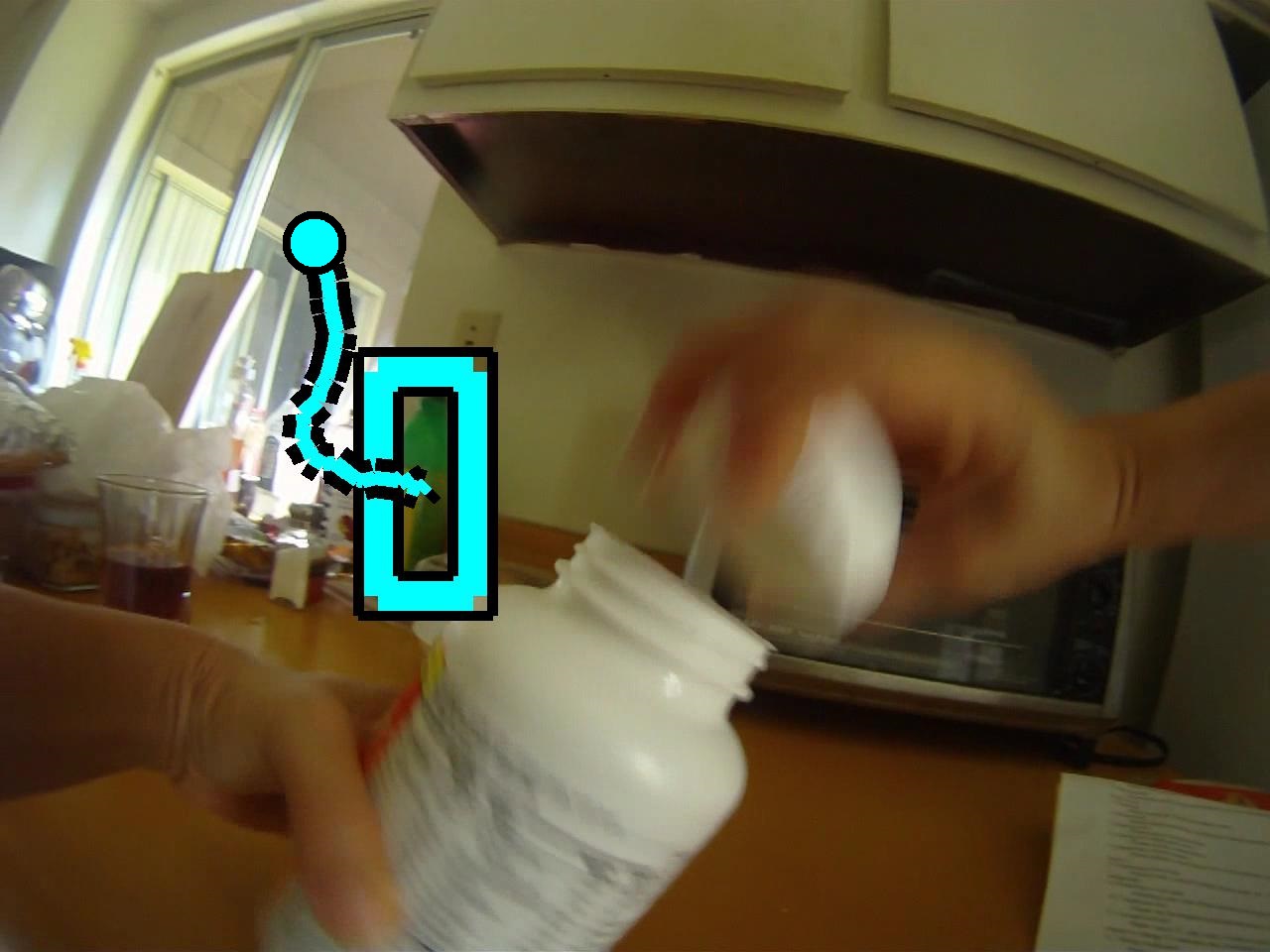}\hfill
	\includegraphics[width=0.24\linewidth]{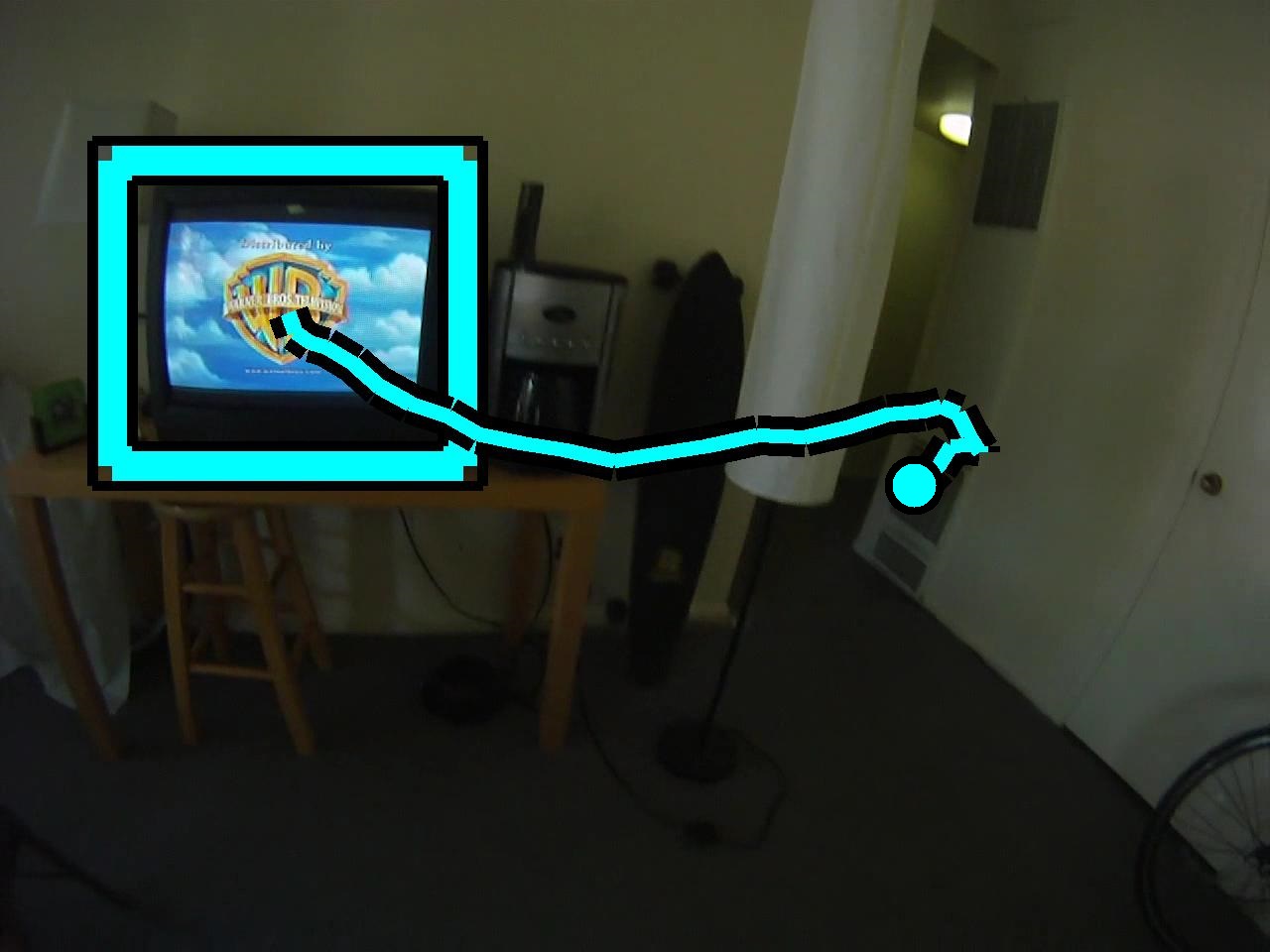}\hfill

	\vspace{0.5mm}
	\includegraphics[width=0.24\linewidth]{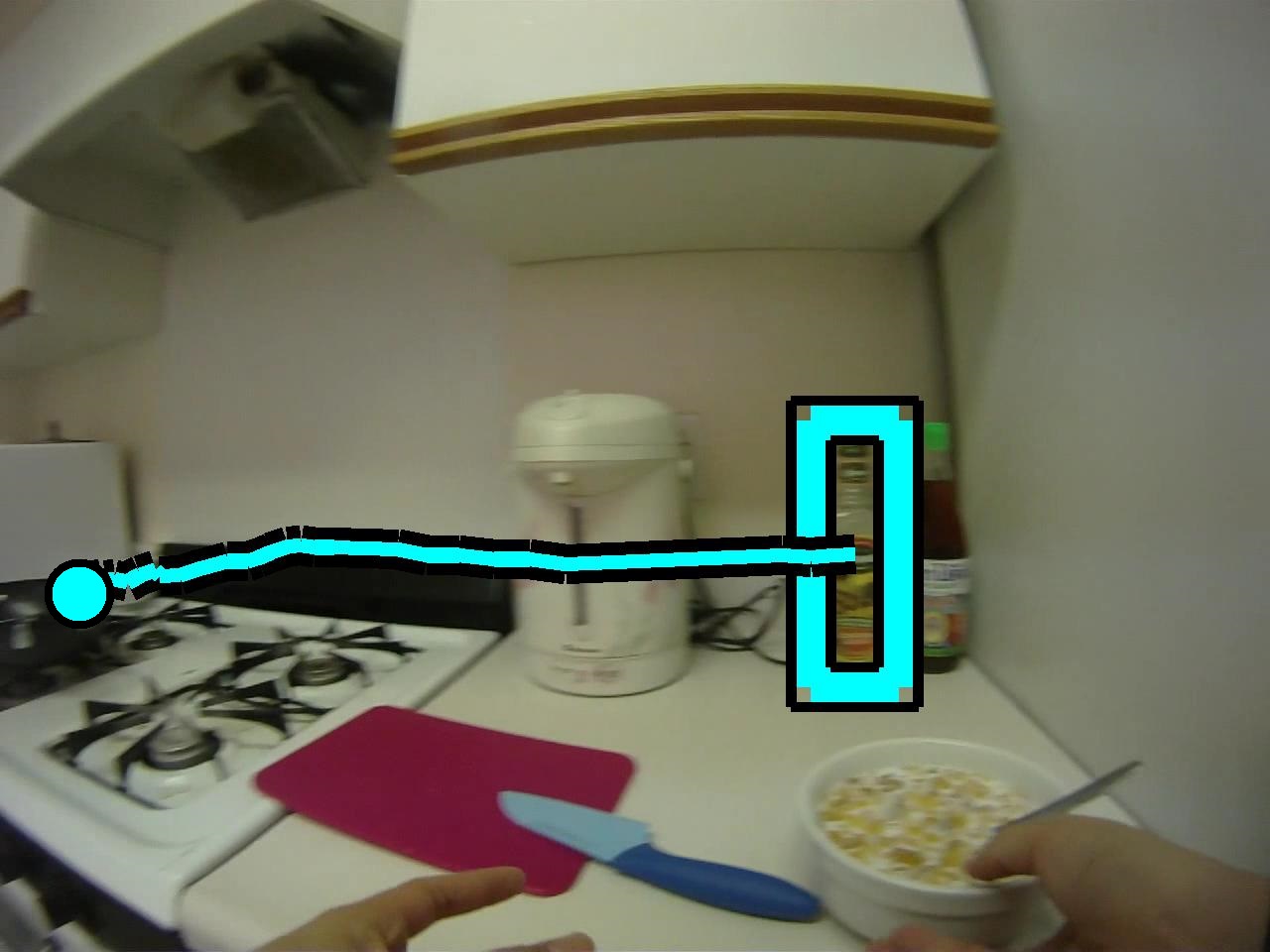}\hfill
	\includegraphics[width=0.24\linewidth]{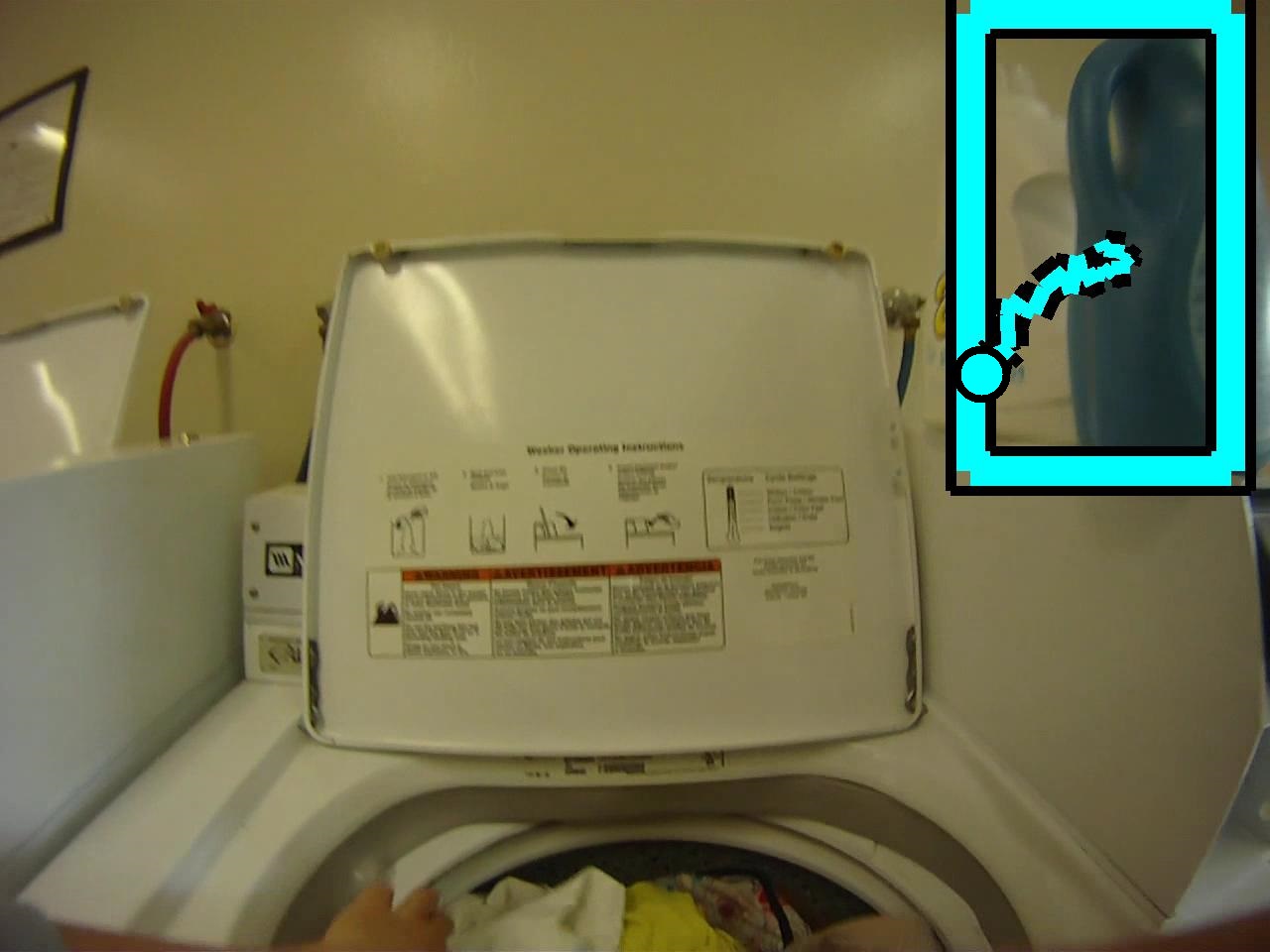}\hfill
	\includegraphics[width=0.24\linewidth]{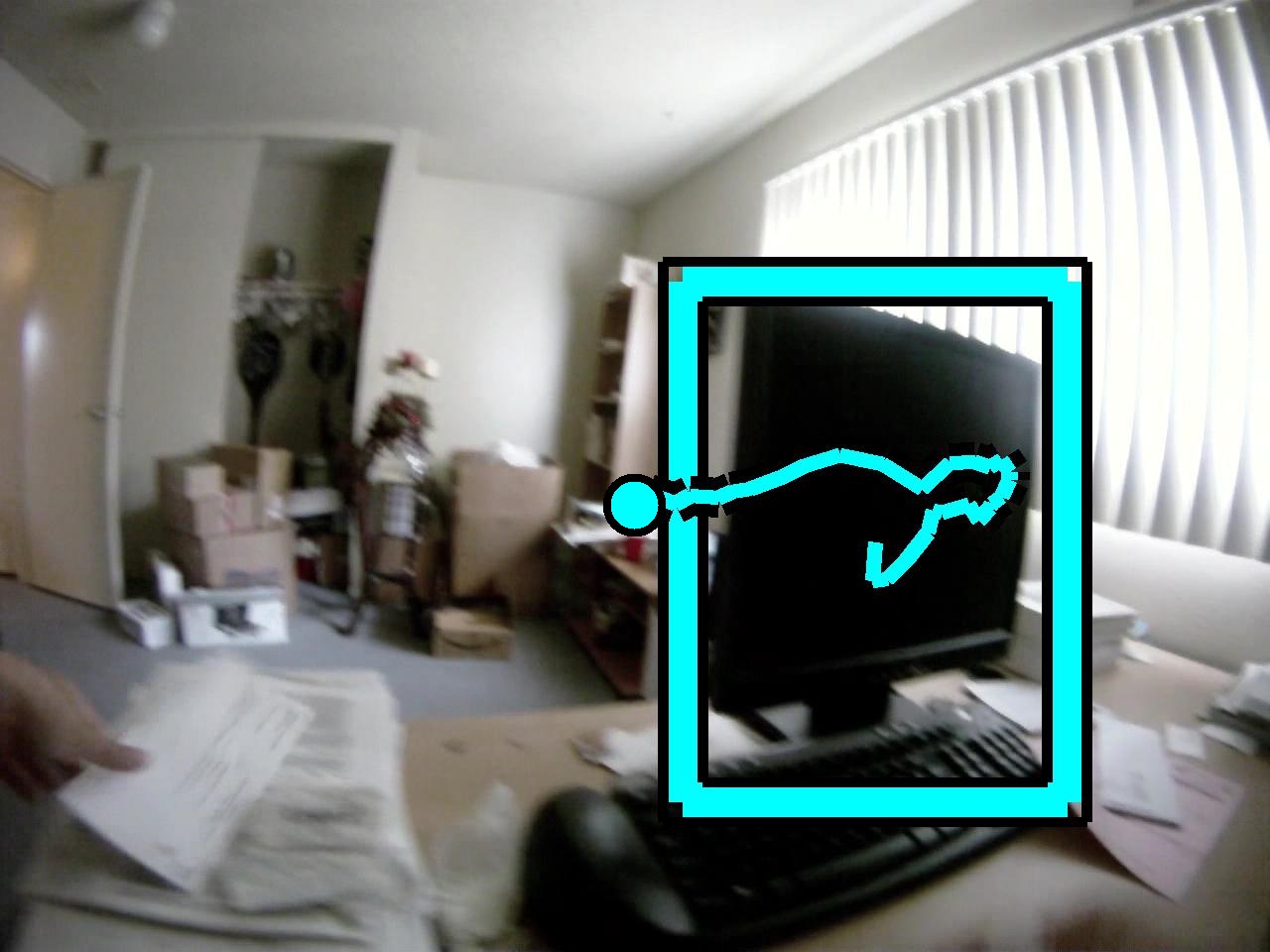}\hfill
	\includegraphics[width=0.24\linewidth]{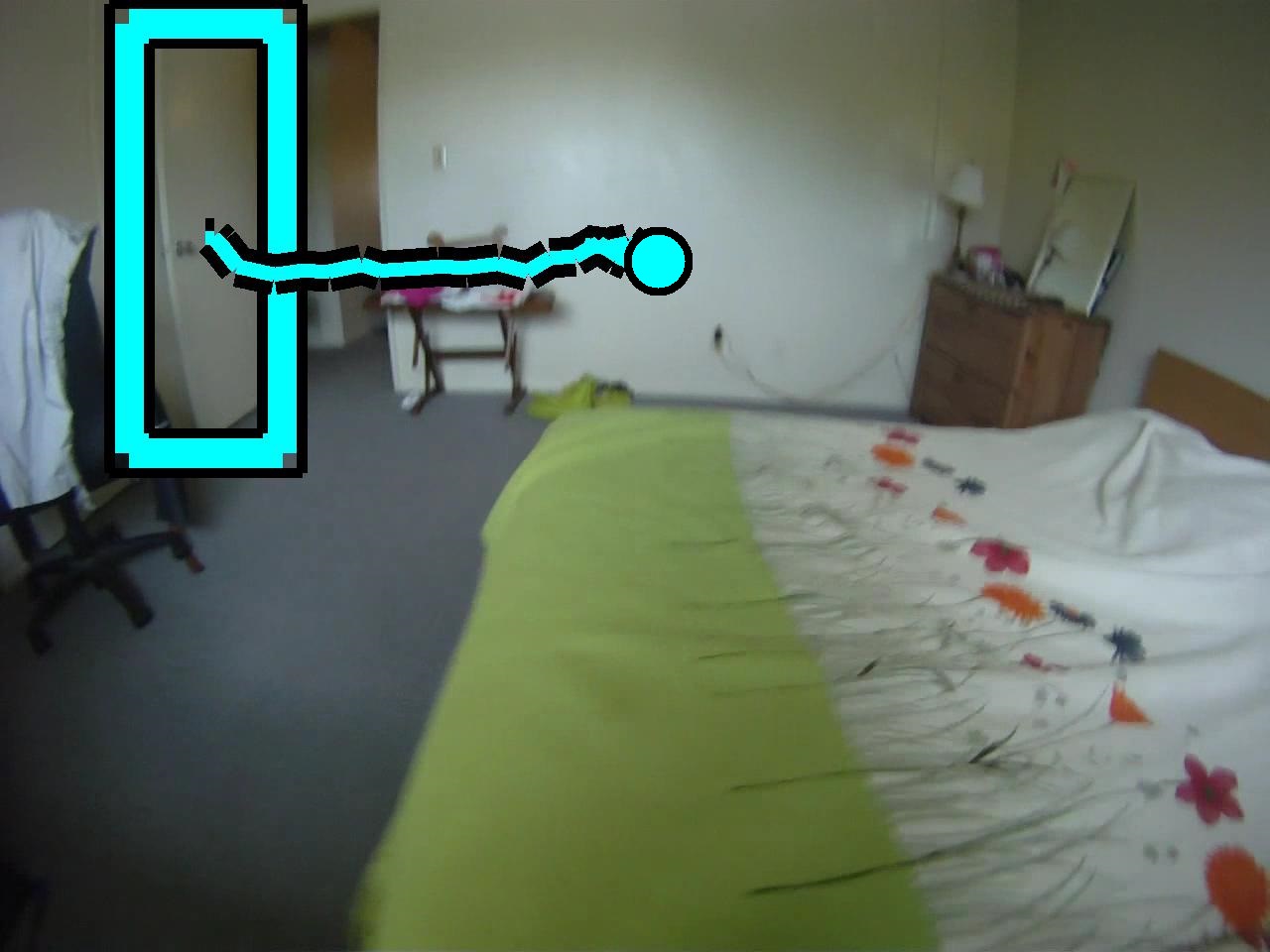}\hfill

	\centerline{(b) passive trajectories}
	\caption{Examples of (a) active and (b) passive trajectories. Starting points of trajectories are indicated by a circle. {Note that, while it is not easy to detect next-active-objects using only appearance, object motion can provide important cues.}}
	\label{fig:trajectory_examples}
\end{figure}

\figurename{~\ref{fig:trajectory_extraction}} illustrates the extraction of active (red) and passive (cyan) trajectories from object tracks, whereas \figurename{~\ref{fig:trajectory_examples}} illustrates some examples of the extracted trajectories. {In particular, as can be noted from} \figurename{~\ref{fig:trajectory_examples}}{, discriminating next-active-objects from passive ones on the basis of their appearance it is not easy. Some objects, indeed, do not change their appearance when they are about to become active (e.g., pan, stove and microwave in subfigure (a)). Others still share similar appearance in both the passive and next-active scenarios (e.g., the fridge at bottom-left of subfigure (a) and top left of subfigure (b)). On the contrary, object motion dynamics (i.e., egocentric object trajectories) can provide meaningful cues for next-active-objects detection.}

\subsection{Active vs Passive Trajectory Classifier}
\label{sec:active_passive_trajectories}
The examples reported in~\figurename{~\ref{fig:trajectory_examples}} show that discriminating active trajectories from passive ones is not trivial. Nonetheless, the egocentric nature of the observations provides some useful cues. Specifically, we expect that
egocentric trajectories of next-active-objects tend to appear at specific scales and absolute positions. For instance, people tend to get closer to next-active-objects and bring them towards the center of their field of view before initiating the interaction. Similarly, people are more likely to pass by other objects avoiding to get too close and pushing them towards the borders of their field of view.

Motivated by these observations, we propose to describe trajectories using 1) the absolute positions in which bounding boxes appear in the frame, 2) differential information about positions, 3) scale and differential information about scale. The main motivation behind point 1) is that absolute position can help discriminate active from passive objects, as observed in~\cite{Pirsiavash2012}. Point 2) is derived from the trajectory shape descriptor used within Dense Trajectories~\cite{Wang2013}. Such descriptor represents the ``shape of the trajectory'' as a sequence of displacement vectors. Point 3) is inspired by~\cite{cipolla1992surface}, where the derivative of the bounding box area is used to estimate ``time to contact''. Each trajectory $T_i$ is hence described as follows:
\begin{eqnarray}
\label{eq:descriptor}
\mathcal{D}(T_i)=(xc_1,yc_1,\ldots,xc_h,yc_h,s_1,\ldots,s_h, \nonumber \\\Delta xc_2, \Delta yc_2, \ldots, \Delta xc_h, \Delta yc_h, \Delta s_2, \ldots, \Delta s_h)
\end{eqnarray}
where $xc_j$ and $yc_j$ are the coordinates of the centers of the bounding box $b_j$, $s_j$ is its area, $\Delta xc_j = (xc_j-xc_{j-1})$, $\Delta yc_j = (yc_j-yc_{j-1})$ and $\Delta s_j = (s_j-s_{j-1})$ encode differential information about position and scale. 
If the length of $T_i$ is $h$, the dimension of the descriptor is $|\mathcal{D}(T_i)|=6h-3$.

\begin{figure}[t]\centering
	\includegraphics[width=\linewidth]{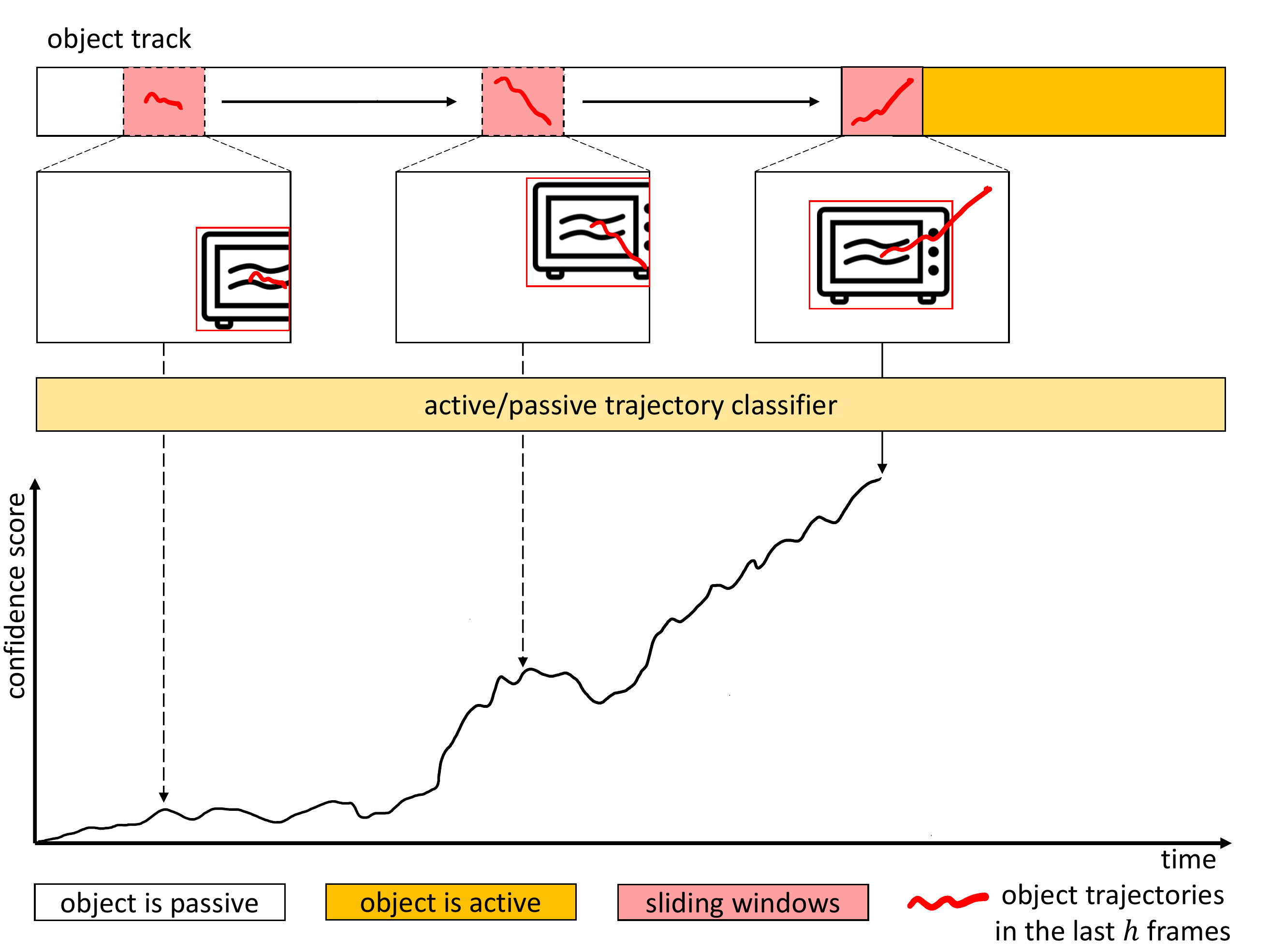}
	\caption{Temporal sliding window processing of object tracks. At each time step, the trained binary classifier is run over the trajectories observed in the last $h$ frames and a confidence score is computed. {Our aim is to predict if an object is going to become active before it actually does, i.e., to fire towards the left of the time plot above.}
	}

	\label{fig:sliding_window}
\end{figure}

\subsection{Sliding Window prediction}
In order to predict which objects are going to become active and which are not over time, we use a temporal sliding window approach. At each time step, the system analyzes the last $h$ frames of the trajectories of each tracked object and predicts them as either active or passive by exploiting the trajectory classifier. If an object has been tracked for less than $h$ frames, it is discarded. For each analyzed object, the system draws a bounding box and assigns a confidence score equal to the probability given by the classifier. This way, likely next-active-objects will get a high confidence score, while passive objects will retain a lower one. \figurename{~\ref{fig:sliding_window}} illustrates the proposed temporal sliding window approach.

\section{Experimental Settings}
\label{sec:predicting_experimental}
{In this Section, we discuss the experimental settings of our experiments. These include the dataset used for the evaluations, how object detection and tracking are carried out and how trajectory classifiers are trained.}

\subsection{Dataset}
\label{sec:prediction_dataset}
For our experiments, we consider the {Activity of Daily Living (ADL)} dataset~\cite{Pirsiavash2012}, which contains egocentric videos acquired using a chest-mounted camera by $20$ subjects performing daily activities. The dataset contains bounding box annotations for $45$ object classes. Annotations related to the same object instance are grouped into tracks and each annotation is labeled as passive or active. We carry out our evaluations on the ADL dataset since it is the only publicly available dataset featuring untrimmed egocentric videos of object interactions ``in the wild'' {(e.g., subjects move through different environments and interact with many different objects)}, including annotations for both active and passive objects.

Since objects are annotated every $30$ frames, reasoning about object trajectories is difficult. To overcome this limitation, we temporally augment the original annotations by tracking objects in those frames which are not annotated. To this aim, we use the short term tracker CMT {(Consensus-based Matching and Tracking)} proposed by Nebehay and Plugfelder in~\cite{Nebehay2015}.
The tracker is always initialized using original ground truth object annotations and tracking is carried on until the next annotation is reached. Active/passive flags are interpolated accordingly.

\subsection{Object Detection and Tracking}
\label{sec:detection_tracking}
At test time, our system analyzes input video to detect objects, group detections into tracks and classify object trajectories to predict next-active-objects. To perform object detection, we consider the state of the art Faster R-CNN method~\cite{Ren2015} based on the VGG-16 network~\cite{Simonyan2015}. We start from the original set of $26$ objects proposed in~\cite{Pirsiavash2012}. Since in our work we propose to detect next-active-objects on the basis of their trajectories and not their appearance, we {do not} train our object detector to distinguish between active and passive objects as done in~\cite{Pirsiavash2012}. Considering only passive objects and removing classes represented by few training samples (i.e., less than $1000$), we obtain a set of $19$ object classes. As in~\cite{Pirsiavash2012}, we train the object detector on images extracted from the first $6$ videos, while the remaining $14$ videos are used to train/test the proposed next-active-object prediction method. 
Note that, in order to train the object detector, we consider only object annotations originally contained in the dataset, while tracked bounding boxes are discarded at this stage. The Faster R-CNN model is trained using the ``end2end'' procedure proposed in~\cite{Ren2015}. {The trained detector achieves a mean Average Precision (mAP) of $0.2772$ on the test set of $14$ videos, which compares favorably with respect to the $0.1515$ mAP scored by the deformable part models employed in}~\cite{Pirsiavash2012}{. Please note that, as pointed out in}~\cite{Pirsiavash2012}{, even performing object detection on the ADL dataset is hard due to the presence of small objects and non-iconic views.}

In order to deal with object trajectories, bounding boxes detected across different frames and related to the same object instance need to be correctly associated. This can be done using a tracker based on a data association algorithm such as the real-time (260 Hz) SORT tracker proposed in~\cite{Bewley2016}. Note that, since we compute detections for each frame at test time, we only need a mechanism able to understand when two detections performed in subsequent frames are related to the same object instance. Hence, it is not necessary to employ a visual tracker such as the CMT tracker used at training time to temporally augment annotated object tracks. At test time, we run the SORT tracker on top of the detections obtained using the Fast-RCNN object detector to obtain object tracks. In our experiments, objects detected with a low confidence score (less than $0.8$) are discarded before employing the SORT tracker. Please note that the Fast-RCNN/SORT component is run directly on the input video. Ground truth tracks are used only for training purposes and are not exploited at test time, unless otherwise specified.

\subsection{Trajectory Classification}
\label{sec:trajectory_classification}
We train Random Decision Forests~\cite{ho1998random}\footnote{We set the number of trees to $25$ and do not set any limit for the maximum height of each tree.} to discriminate between passive and active object trajectories. In the considered dataset, the number of negative trajectories is usually far larger than the number of active ones. To mitigate such imbalance, at training time, the number of passive trajectories is randomly subsampled to match the number of active ones, in order to obtain a balanced training set. 
Testing is always performed on the original unbalanced data. 

We assess the performance of the trained classifiers with respect to different factors, including the temporal support with respect to which trajectories are analyzed, the employed trajectory descriptor and the generalization to unseen object classes. All results are reported in terms of Precision-Recall curves and related Average Precision (AP) values.

\section{Results}
\label{sec:prediction_results}

We perform all our experiments in a leave-one-person-out fashion on the set of $14$ videos which have not been used to train object detectors (as done in \cite{Pirsiavash2012}).
At each leave-one-out iteration, trajectory classifiers are learned on videos acquired by $13$ subjects and tested on data acquired by the remaining subject. We make sure that training and testing data are always acquired by different subjects to prevent the system from over-fitting to a single user, i.e., learning the specific way he moves and interacts with objects. All reported results are averaged across the $14$ leave-one-out iterations. 

{In the rest of this section, we first discuss the performance of the trajectory classifier component alone in Section~}\ref{sec:performance_trajectory_classifier}{), then analyze the performance of the overall system and report comparative results with respect to several baselines in Section~}\ref{sec:performance_overall}{.}

\subsection{Performance of the Trajectory Classifier}
\label{sec:performance_trajectory_classifier}
{In this section, we analyze the performance of the trajectory classifier component with respect to different encoding schemes and parameters.}

\subsubsection{Trajectory Length and Encoding}
\label{sec:temporal_extent}
In Section~\ref{sec:active_passive_trajectories}, we assumed that the last part of an active trajectory is the most discriminative for our task. Therefore, we proposed a sliding window approach which analyzes fixed-length trajectories within a temporal window of size~$h$. To support that analyzing trajectories within a fixed-length temporal window is optimal, we compared the proposed method to a different schema which, at each time step, analyzes the whole trajectory observed up to that point.
In this second schema, in order to obtain a fixed-length descriptor, trajectories are represented with a multiscale approach. Using a temporal pyramid with $l$ levels~\cite{Pirsiavash2012}, each trajectory is divided into $2^l-1$ segments. Bounding boxes within the same segment are averaged and the results concatenated. This leads to fixed-length trajectories 
which are hence represented using the descriptor introduced in Eq.~\eqref{eq:descriptor}. Note that the maximum number of splits operated by the temporal pyramid is equal to $2^{(l-1)}$, therefore, trajectories shorter than this number are discarded in our experiments.

\begin{figure}\centering
	\includegraphics[width=\linewidth]{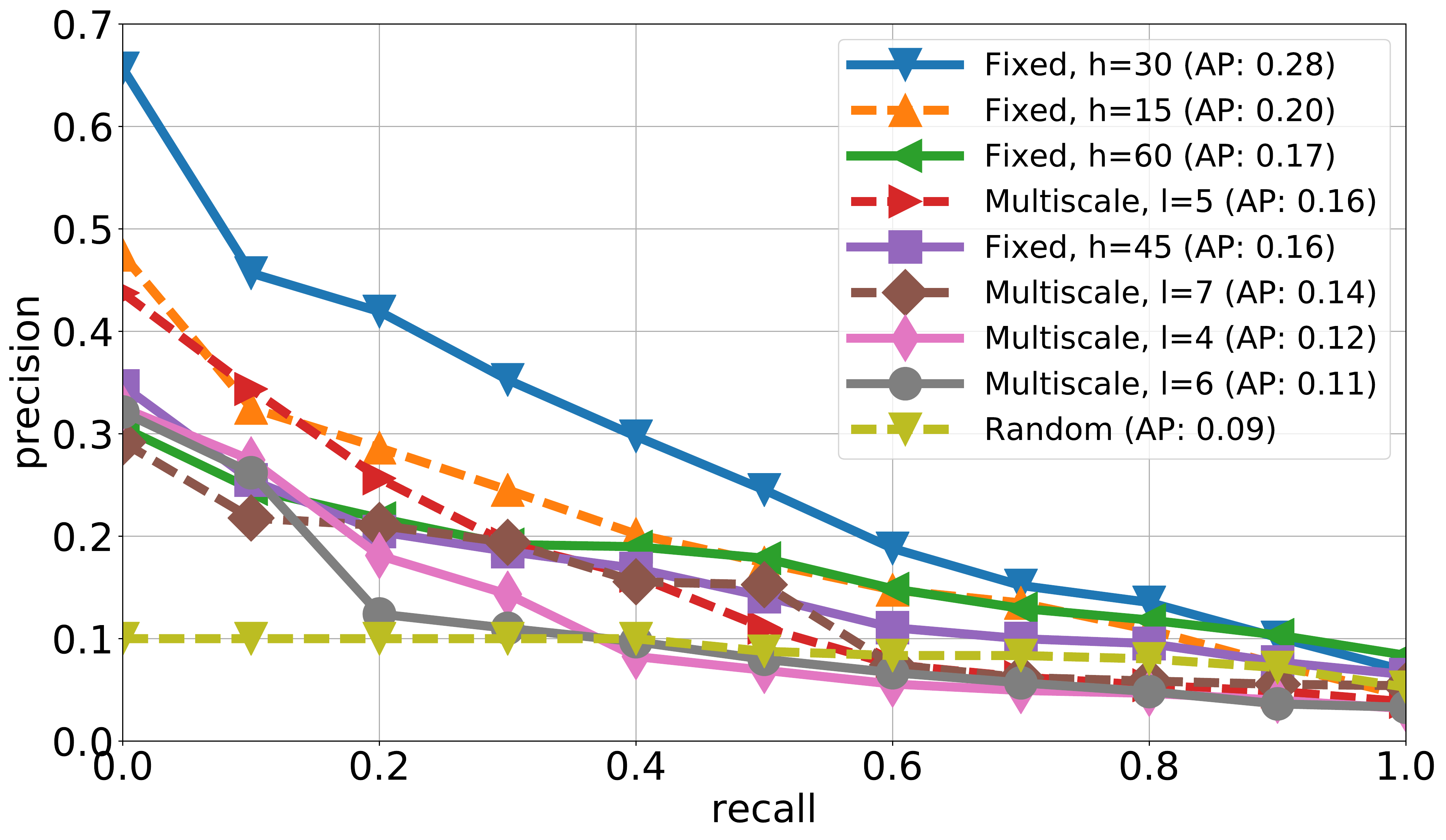}
	\caption{Precision-recall curves related to different trajectory description schemes. 
		Elements in the legend are sorted by average precision in descending order.}

	\label{fig:temporal_extent}
\end{figure}

{As discussed in} Section{~\ref{sec:trajectory_classification}}{, training/test trajectories are extracted from object tracks related videos $7$ to $20$ of the ADL dataset. Since active objects are naturally rarer than passive ones, the resulting dataset is highly unbalanced, which active trajectories amounting to about $6\%$ of the dataset. Given this premise, and since our focus is on detection, rather than classification, we report our results in terms of precision-recall curves and AP scores.}

\figurename{~\ref{fig:temporal_extent}} reports precision-recall curves of the classifiers learned on trajectories extracted according to the two considered schemes. In particular, our method scores an AP of $0.28$ {on more than $2$ hours of test video}, while the chance level is $0.09$. 
The proposed fixed-length trajectory approach has been evaluated considering different lengths $h=\{15,30,45,60\}$. Similarly, the multiscale approach has been evaluated considering different number of levels $l=\{4,5,6,7\}$. Please note that the minimum trajectory lengths associated to the considered numbers of levels are respectively $\{8,16,32,64\}$. The random baseline is obtained performing classification with a random binary decision. {It should also be noted that the choice of the parameters related to the length of trajectories depends on the frame-rate at which videos are acquired. In this paper, we assume a standard framerate of $30fps$.} 

As can be observed in \figurename{~\ref{fig:temporal_extent}}, classifiers based on fixed-length trajectories tend to outperform methods based on multiscale trajectories. This suggests that the last part of active trajectories is the most discriminative and that motion information too far away from the activation point introduces noise in the observations. Among the methods based on fixed-length trajectories, the best performing scheme is the one analyzing trajectories of length $h=30$. This value will be used in all the following experiments.

\subsubsection{Analysis of Trajectory Descriptors}
\label{sec:trajectory_descriptors}
As discussed in Section~\ref{sec:active_passive_trajectories}, the proposed trajectory descriptor introduced in Eq.~\eqref{eq:descriptor} includes information about absolute positions and scales, as well as differential information about position and scale. We analyze the impact of each of these kinds of information comparing the proposed descriptor against the following baselines:

\begin{itemize}
	\item \textbf{Motion Magnitude}: we consider discriminating active trajectories from passive ones on the basis of the amount of motion characterizing the trajectory $T_i$ under analysis. The amount of motion is measured as the sum of the magnitudes of the displacement vectors: $M(T_i)=\sum_{j=2}^{h}\sqrt{\Delta xc_j^2+\Delta yc_j^2}$. Classification is hence performed by thresholding on $M$. The optimal threshold is selected at training time as the one best discriminating active from passive trajectories in the training set;

	\item \textbf{Relative Trajectories}: are the descriptors proposed by Wang et al. in their work on Dense Trajectories~\cite{Wang2013}: $\mathcal{D}(T_i)=\frac{(\Delta xc_2, \Delta yc_2, \ldots, \Delta xc_h, \Delta yc_h)}{\sum_{j=2}^{h}\sqrt{\Delta xc_j^2+\Delta yc_j^2}}$. These descriptors encode only the ``shape'' of the trajectory and do not include any information about absolute positions; 

	\item \textbf{Absolute Trajectories}: described as the concatenation of the centers of all bounding boxes: $\mathcal{D}(T_i)=(xc_1, yc_1, \ldots, xc_h, yc_h)$. Such descriptors include positional information but do not encode scale and differential information; 

	\item \textbf{Absolute Trajectories + Differential Positions}: described as the concatenation of positions and differential information about position: $\mathcal{D}(T_i)=(xc_1, yc_1, \ldots, xc_h, yc_h, \Delta xc_2, \Delta yc_2, \ldots, \Delta xc_h, \Delta yc_h)$. These descriptors encode location and trajectory shape but do not include scale information;

	\item \textbf{Absolute Trajectories + Scale}: described as the concatenation of positions and bounding box scales: $\mathcal{D}(T_i)=(xc_1, yc_1, \ldots, xc_h, yc_h, s_1, \ldots, s_2)$. These descriptors encode location and scale but do not include differential information.
\end{itemize}

\begin{figure}\centering
	\includegraphics[width=\linewidth]{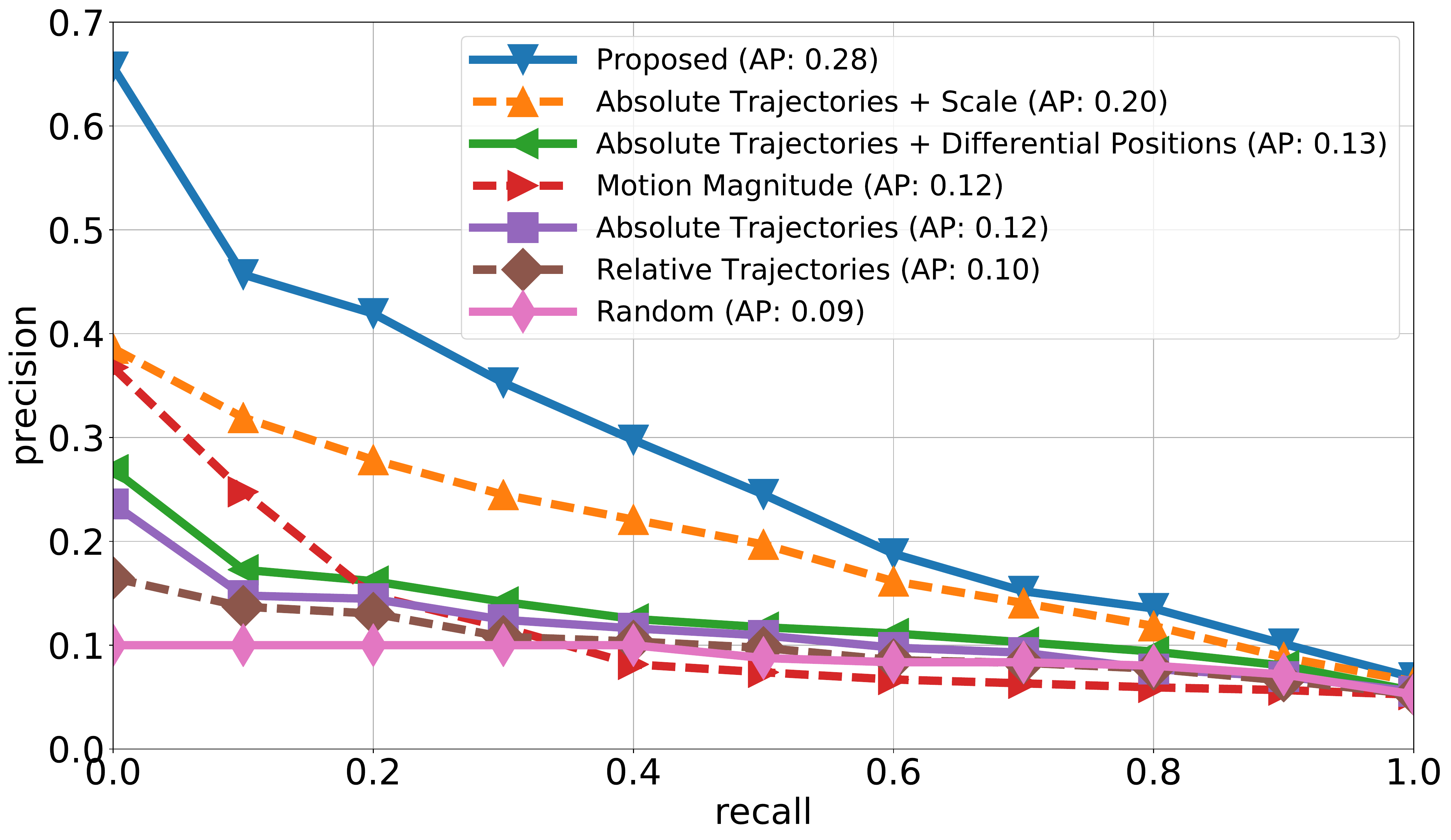}
	\caption{Precision-recall curves related to the proposed method and compared baselines.}
	
	\label{fig:trajectory_descriptors}
\end{figure}

\figurename~\ref{fig:trajectory_descriptors} shows precision-recall curves for the proposed method and the compared baselines. As can be observed, relative trajectories~\cite{Wang2013} (AP: $0.10$) are less discriminative than absolute trajectories (AP: $0.12$). This confirms the observation according to which position can help discriminate active and passive objects~\cite{Pirsiavash2012}. Combining absolute and differential positional information improves performances marginally (AP: $0.13$). Adding scale (AP: $0.20$) and above all, combining with differential information as we propose allows to obtain the best results (AP: $0.28$). Interestingly, the motion magnitude baseline performs better than some competitors (AP: $0.12$). {This reveals that discriminating between moving and static objects is already a reasonable baseline to reject some passive objects. However, it should be noted that reasoning about trajectories, positions and scales is essential to achieve better results.}

\subsubsection{{Analysis with respect to Time of prediction}}
\begin{figure}[t]\centering
	\includegraphics[width=\linewidth]{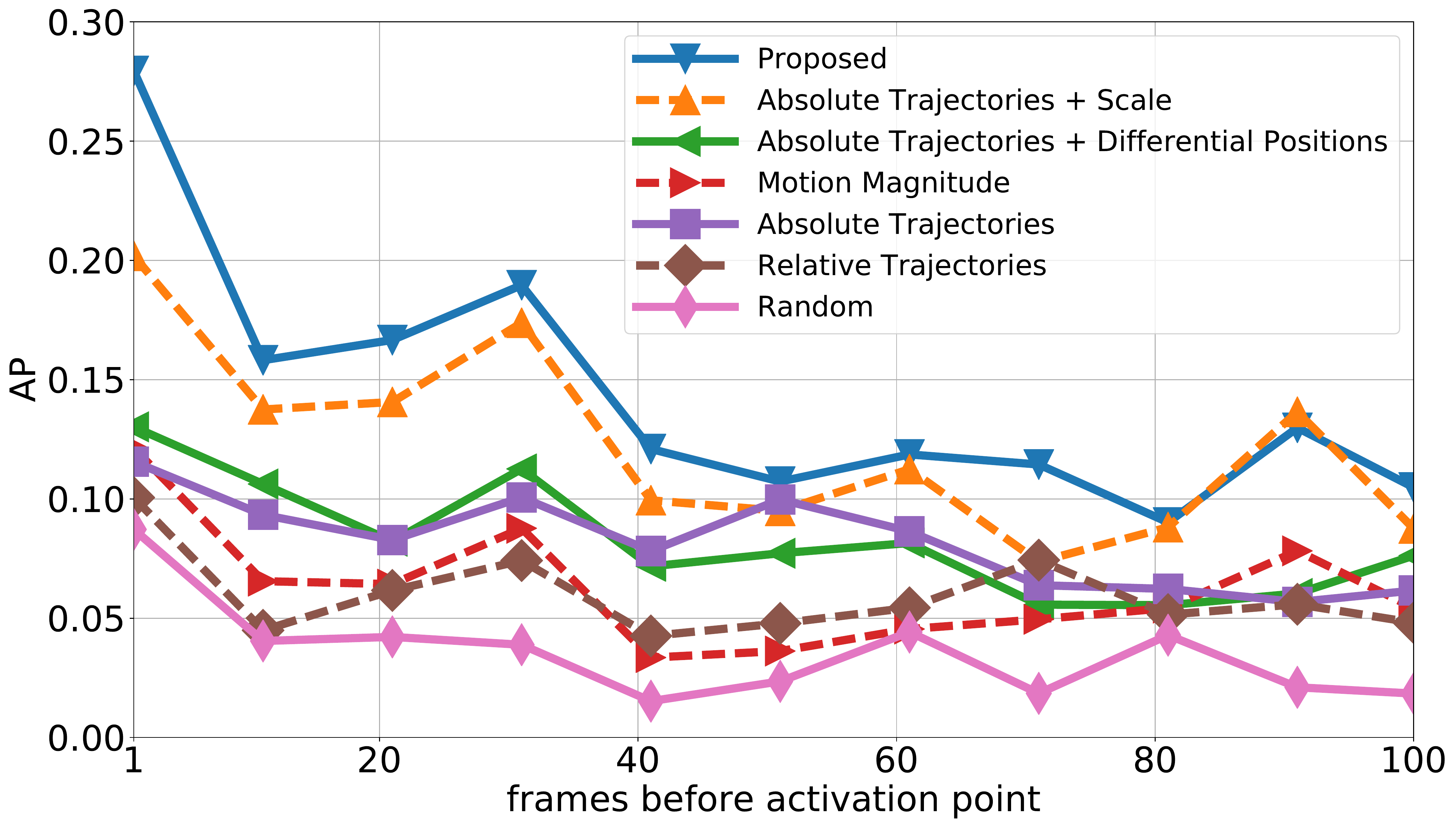}
	\caption{Average Precision of the considered methods when they are evaluated a given number of frames before the activation point. Methods retain some predictive power up to about $30$ frames ($1$ second) before the activation point. The proposed method generally performs better than others and significantly better than the random baseline $100$ frames (about $3$ seconds) before the activation point.}
	\label{fig:trajectory_descriptors_distance}
\end{figure}

{The proposed classifier is trained to discriminate trajectories observed in the last $30$ frames before an object activation from all others. We report experiments to assess up to what extent the learned classifier is still able to detect next-active-objects a number of frames in advance with respect to the activation point.} \figurename~\ref{fig:trajectory_descriptors_distance} {reports AP results when classifiers are evaluated a given number of frames before the activation point. As it can be expected, best results are obtained in proximity to the activation point. However, all classifiers retain a certain amount of predictive power up to $30$ frames (1 second) before the activation point. Moreover, it should be noted that the proposed descriptor generally achieves best results as compared to other descriptors and still performs significantly better than the random baseline $100$ frames (about $3$ seconds) before the activation point.}

\subsubsection{{Generalization to Unseen Object Classes}}
\label{sec:generalization_objects}
We have trained a single active {versus} passive classifier including data from all considered object classes. While training object-specific trajectory classifiers might be advantageous, the limited number of samples related to a single object class could pose a challenge. Moreover, ideally, a real system needs to be able to handle ``open-world'' situations in which objects belonging to previously unseen categories may become active. {For example, a camera wearer ought to be able to enter a new environment where objects unavailable in training are nonetheless important to detect as next-active at test time.}
We find that next-active-object trajectory classification can generalize to previously unseen object classes up to a given extent. To assess this property, we performed a leave-one-object-out experiment. For each object class, we trained trajectory classifiers on data related to all other object classes. Classifiers have been hence tested on data including only the object class which was removed from the training set. 

\tablename~\ref{tab:generalization} reports the results for the considered object classes. Classes missing from \tablename~\ref{tab:generalization} are those which were not represented by any sufficiently long trajectory (at least $h$ frames) in the dataset. Classifiers learned from training sets not containing the target object class (``w/o'' column) are compared to classifiers learned from training sets containing also instances from the target object class (``with'' column). Similar performances are achieved for many object classes (e.g., door, tv, book, mug/cup, laptop), whereas for others learning from instances of the same object class is more beneficial. {This may imply that, for some object classes, it is crucial to learn specific motion/position/scale information. This is probably the case of ``oven/stove'' which, being a fixed object, tends to appear at specific locations and scales, or ``bottle'' and ``kettle'' which have peculiar aspect ratios and tend to appear at smaller scales.}
However, it should be noted that, on average, removing the object class from the training set implies a reasonable performance loss of $0.11$ AP.

\begin{table} \small
	\centering
	\hfill
	\begin{adjustbox}{width=\linewidth}
		\begin{tabular}{lcc}
			& \multicolumn{2}{c}{\textbf{AP}}\\
			\textbf{Object} & \textbf{w/o} & \textbf{with} \\ \hline
			oven/stove &  0.60 & 0.82  \\
			tap & 0.47 & 0.59  \\
			door & 0.19 & 0.18  \\
			tv remote & 0.58 & 0.73  \\
			bottle & 0.50 & 1.00  \\
			pan & 0.56 & 0.83   \\
			tv & 1.00 &1.00 \\
			\hline
		\end{tabular}
	\hfill
		\begin{tabular}{lcc}
			& \multicolumn{2}{c}{\textbf{AP}} \\
			\textbf{Object} & \textbf{w/o} & \textbf{with} \\ 	\hline
			fridge & 0.47 & 0.73  \\
			book & 1.00 & 1.00  \\
			microwave & 0.67 & 0.40   \\
			kettle & 0.33 & 0.75  \\
			mug/cup & 0.52 & 0.62   \\
			dish & 0.51 & 0.32   \\
			laptop & 0.63 & 0.65  \\
			\hline
		\end{tabular}
\end{adjustbox}
	\hfill

	\caption{Average precision results related to the leave-one-object-out experiment.}
	\label{tab:generalization}

\end{table}

\subsection{{Performance of the Overall System}}
\label{sec:performance_overall}
{In this section, we discuss the performance of the overall system, comparing it to several baselines.}

\subsubsection{{Comparative Experiments}}
\label{sec:comparative_experiments}
In order to compare different methods in a common evaluation scheme, we frame next-active-object prediction as an object detection task. We assume that, at each time step, each method produces a series of bounding boxes around the predicted next-active-objects and assigns a confidence score to them.

We define our ground truth starting from the object annotations of the ADL dataset augmented by tracking as described in Section~\ref{sec:prediction_dataset}. Since we wish to predict next-active-objects as soon as possible, all annotations which are on the passive segments of a mixed track (see \figurename~\ref{fig:trajectory_extraction}) are considered as valid detections. All other annotations, namely, the ones which are on passive tracks and the ones which are in the active part of mixed tracks are not considered valid detections. The performance of the investigated methods is measured computing precision-recall curves and Average Precision (AP) values. 
A prediction is considered correct if there is a significant overlap (area of intersection over union (IOU) $\geq 0.5$) with an annotation of the same object class. {Note that, since we are first to tackle the problem, no existing methods are available for direct comparison. Therefore, we adapt known techniques to our problem and propose a series of baselines with respect to which we compare the proposed method. Considered baselines are discussed in the following:}

\begin{itemize}
	\item \textbf{Motion Magnitude}: the same baseline discussed in Section~\ref{sec:trajectory_descriptors} based on thresholding over motion magnitude;

	\item \textbf{Relative Trajectories}: the same baseline discussed in Section~\ref{sec:trajectory_descriptors} based on the trajectory descriptors introduced by Wang et al.~\cite{Wang2013};

	\item \textbf{Center Bias}: this baseline considers the assumption made by Pirsiavash and Ramanan~\cite{Pirsiavash2012}, according to which active objects tend to appear near the center of the frame. The baseline analyzes the object detections produced by the Faster-RCNN detector and takes into account the confidence score assigned to each predicted bounding box $s_o$. For each detected object, we compute a score $s_c$ which is inversely proportional to its distance from the center of the frame. The final confidence score is obtained as $s=s_c \cdot s_o$;

	\item \textbf{Hand Bias}: the presence of hands is a cue often considered for detecting active objects~\cite{fathi2011understanding,Fathi2012,Ma2016going,li2015delving}. To leverage this cue, we detect hands from the input videos by using the models proposed in~\cite{egohands2015iccv}. Similarly to the center bias baseline, for each object detection we compute two scores $s_{lh}$ and $s_{rh}$ which are inversely proportional to the distances of the object from the left and right hand respectively. If one of the two hands is missing, a score equal to zero is assigned. The final confidence score is obtained by $s=s_o \cdot (s_{lh} + s_{rh})$, where $s_o$ is the confidence score assigned to the predicted bounding box;

	\item \textbf{Active/Passive Objects}: a method inspired by the work of \cite{Pirsiavash2012}. predictions are obtained using a Faster R-CNN object detector trained to detect active and passive objects separately. The detector is hence trained on 38 classes (19 active objects and the corresponding 19 passive ones). {It should be noted that, while this baseline does not completely fit our task (the detector is not explicitly trained to detect next-active-object), it is still useful to ensure that the problem cannot be trivially tackled by means of such a well-known technique for active object recognition;}

	\item \textbf{Saliency-Based Models}: this set of baselines is inspired by Damen et al.~\cite{Damen2015}, who propose to detect task relevant objects using a gaze tracker, exploiting the anticipatory nature of eye gaze fixation~\cite{land2006eye}. Since we do not assume the availability of a gaze tracker, we implement such baselines using saliency prediction models. The baseline works as follows. Saliency maps are first extracted from each frame. Starting from the Faster-RCNN detections, each predicted bounding box is assigned a score equal to the mean saliency value within the bounding box.
	Given the different levels at which saliency is defined~\cite{furnari2014experimental}, we consider the model proposed by Vig et al.~\cite{vig2014large} for eye fixation prediction, the model proposed by Seo et al.~\cite{seo2009static} for dynamic saliency from videos, and the model proposed by Zhang et al.~\cite{zhang2015MBD} for salient object segmentation;

	\item \textbf{Random}: starting from the Faster-RCNN detection, each bounding box is assigned a random score in the interval $[0,1]$.
\end{itemize}
\begin{figure}\centering
	\includegraphics[width=\linewidth]{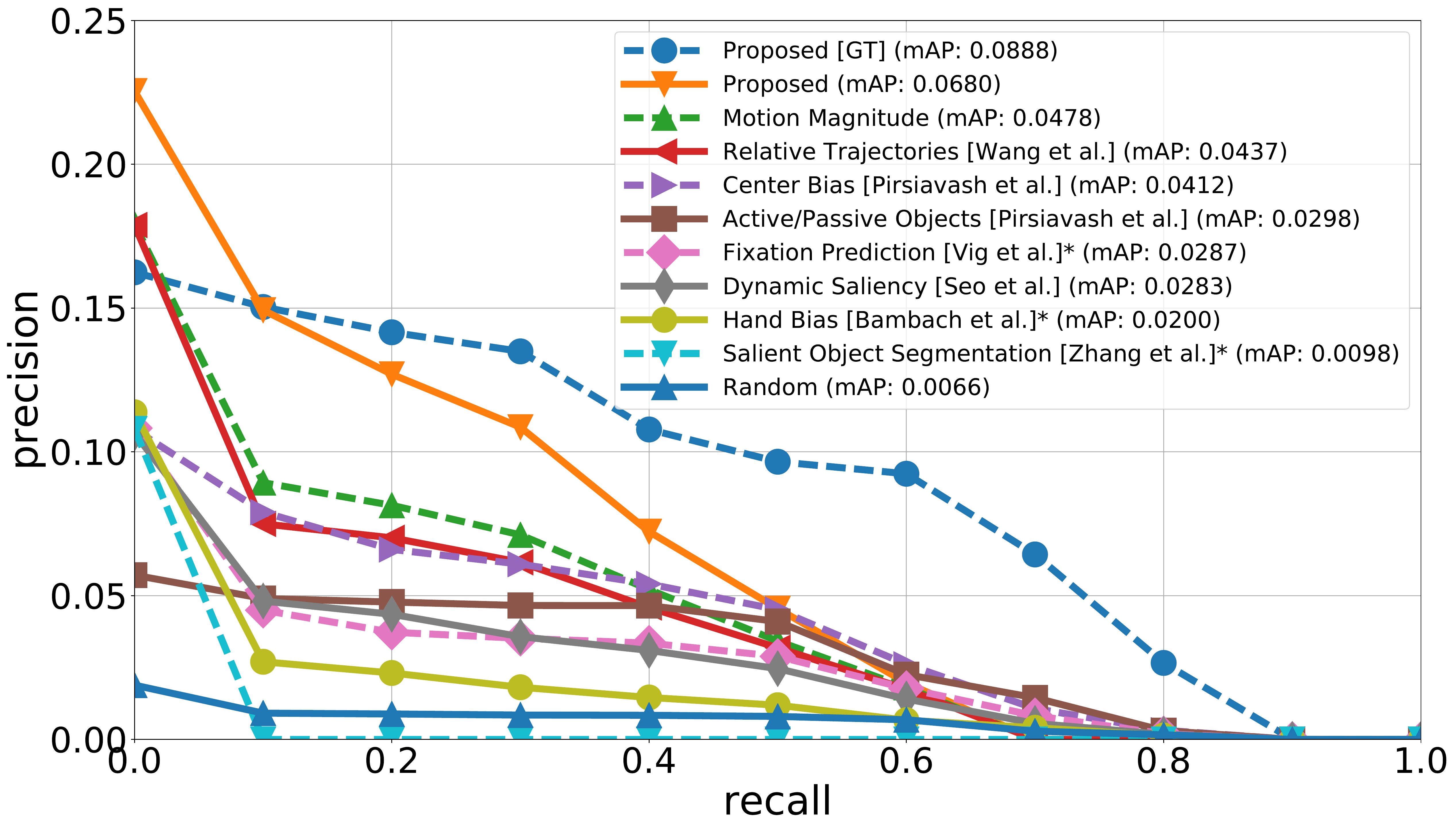}
	\caption{{Precision-recall curves of the compared methods. It should be noted that methods based on egocentric motion analysis perform better than those based on appearance, attention, center or hand bias. The proposed approach is the best performing among the competitors. }}

	\label{fig:comparisons_pr}
\end{figure}

\figurename~\ref{fig:comparisons_pr} reports the precision-recall curves scored by our method and all baselines. To reduce computational burden, the methods indicated by the ``*'' symbol have been evaluated on a subset of the data obtained taking one frame every $30$ frames. The figure also reports results of the proposed approach when run directly on ground truth object tracks (method ``Proposed[GT]''). All other methods are run on object tracks detected/tracked as described in Section~\ref{sec:detection_tracking}.
Among methods run on detected tracks, the proposed one is the best performing one (AP: $0.0680$), followed by the motion magnitude (AP: $0.0478$) and relative trajectory baselines~\cite{Wang2013} (AP: $0.0437$). It is worth noting that the best performing methods are all based on egocentric object motion. The method based on center bias outperforms the appearance-based baseline derived from~\cite{Pirsiavash2012} ($0.0412$ vs $0.0298$ AP values). Our main insight about this behavior is that object appearance is likely to change while the object is being manipulated (i.e., active object already observed) rather than before (i.e., next-active-object prediction). 
The baseline based on hand bias does not achieve good performance (AP: $0.0200$). This is probably due to different factors. First, detecting hands in unconstrained egocentric videos is not trivial~\cite{egohands2015iccv}. Second, hands are not always visible until the object manipulation actually begins. Saliency-based baselines perform worse than others. It should be noted that such methods have been designed to predict current and not future visual attention mechanisms and that they have not been specifically designed for the egocentric scenario. {Moreover, while we perform our evaluations on the ADL dataset, which have been acquired using a chest-worn camera, baselines exploiting attention mechanisms are based on observations generally applicable to the scenario of head-mounted cameras}~\cite{Damen2015}. In particular, attention-based methods might be unable to leverage head-motion cues as expected.

{We would like to note that all presented results are characterized by low Average Precision. This is due to the very low number of frames containing at least one next-active-object, which, from our analysis, amounts to only the $5\%$ of tested frames. Under these circumstances, methods are likely to be affected by the presence of false positives, as it is suggested by the observation of precision recall curves in }\figurename~\ref{fig:comparisons_pr}{, where the maximum precision value achieved at zero recall is equal to $0.23$.} Moreover, results highlight how, due to the ambiguity introduced by human discretion, the prediction of next-active-objects from egocentric video is a hard task. Nevertheless, the proposed analysis points out the importance of egocentric object motion in the considered task and does not exclude that better results could be achieved integrating also other cues such as the way object and scene appearance changes through time and the relationship with the activity performed at the moment of the interaction.

\subsubsection{{Performance Analysis of the Proposed Approach}}
{To assess possible limitations introduced by the object detector/tracker component} \figurename~\ref{fig:comparisons_pr} {also reports results obtained running the proposed method directly on ground truth object trajectories (method ``Proposed [GT]''). As can be expected, the method performs better when run on ground truth trajectories. However, the relatively small increment in AP score ($0.0888$ vs $0.0680$), suggests that overall performance is not substantially limited by the object detector/tracker component.}

\begin{table}
	\centering
	\begin{adjustbox}{width=\linewidth}
	\begin{tabular}{rllllllll}
		\hline
		Conf. Threshold 	& 0.5 & 0.56 & 0.68 & 0.74 & 0.8 & 0.83 & 0.86 & 0.90  \\
		Frac. Act. Pred. 	& 0.92 & 0.89 & 0.79 & 0.67  & 0.48 & 0.32 & 0.24 &  0.11  \\
		\hline
		Precision 			& 0.05 & 0.05 & 0.05 & 0.05 & 0.04 & 0.04 & 0.04 & 0.04 \\
		Recall 				& 0.27 & 0.26 & 0.22 & 0.17 & 0.10 & 0.09 & 0.05 & 0.02 \\
		\hline
	\end{tabular}
\end{adjustbox}
	\caption{Fraction of active objects classified as next active (Frac. Act. Pred.) for given confidence thresholds, along with precision and recall of the overall system. }
	\label{tab:active_fire}
\end{table}

{As already discussed, the main source of error is due to the influence of false positive predictions. Among such cases, the method should not fire in the presence of objects which are already active. To assess performance in this regard, we report in} \tablename~\ref{tab:active_fire} {the fraction of predictions mistakenly performed in the presence of active objects when different confidence thresholds are used to obtain detections from confidence scores. To put such numbers in context, we also report precision and recall of the overall system for the selected confidence threshold. Results are reported for a single iteration of the leave-one-out-procedure, where the test is performed on video $7$ and all other videos are used for training. The proposed method tends to fire in the presence of active objects. The fraction of wrong predictions can be lowered by increasing the confidence threshold, but this also decreases the overall performance of the system. While this remains a limitation of the proposed system, it should be noted that it has not been explicitly trained to classify active objects as not being next-active.}

\begin{figure*}
	\includegraphics[width=0.15\linewidth]{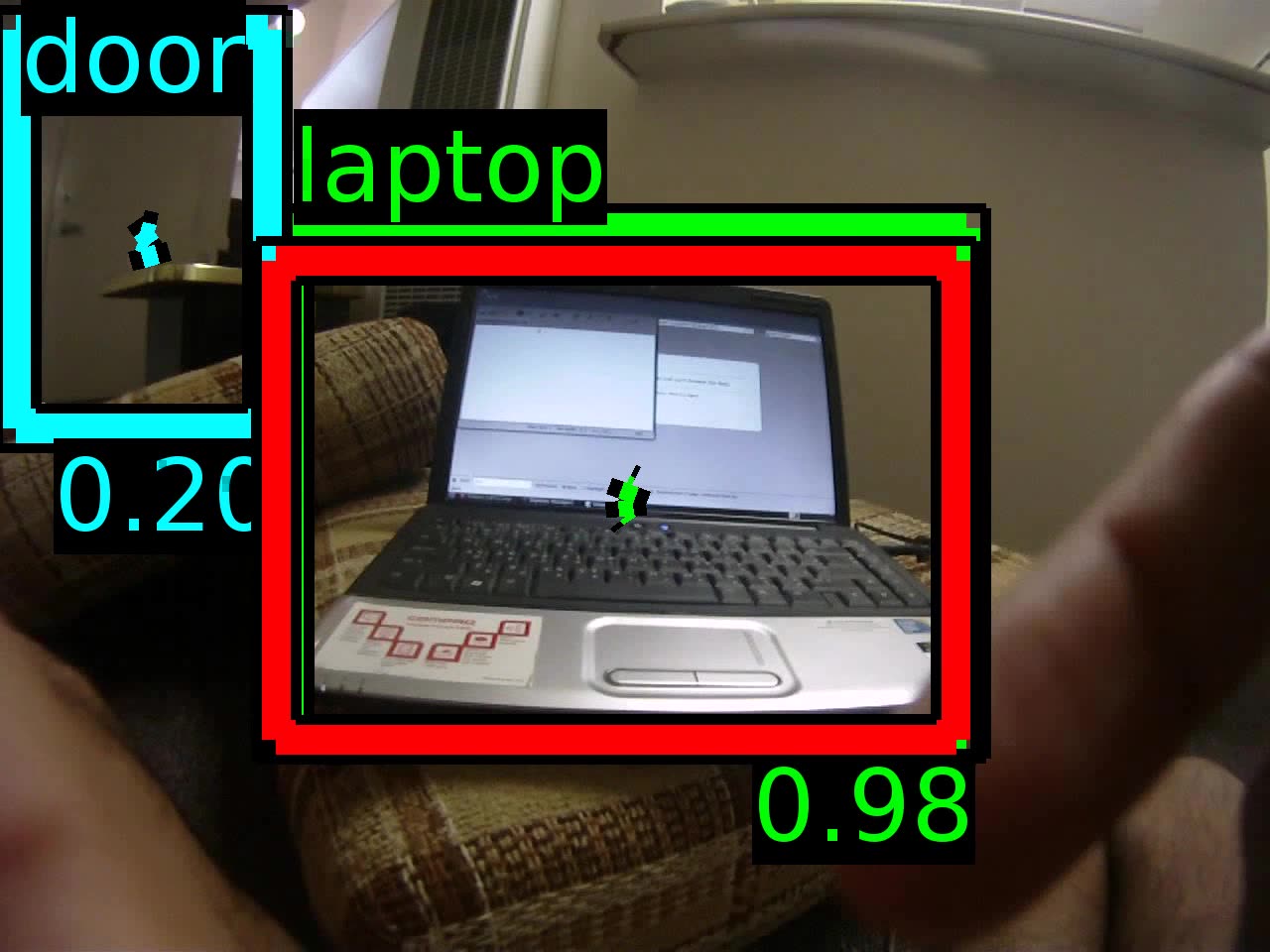} \hfill
	\includegraphics[width=0.15\linewidth]{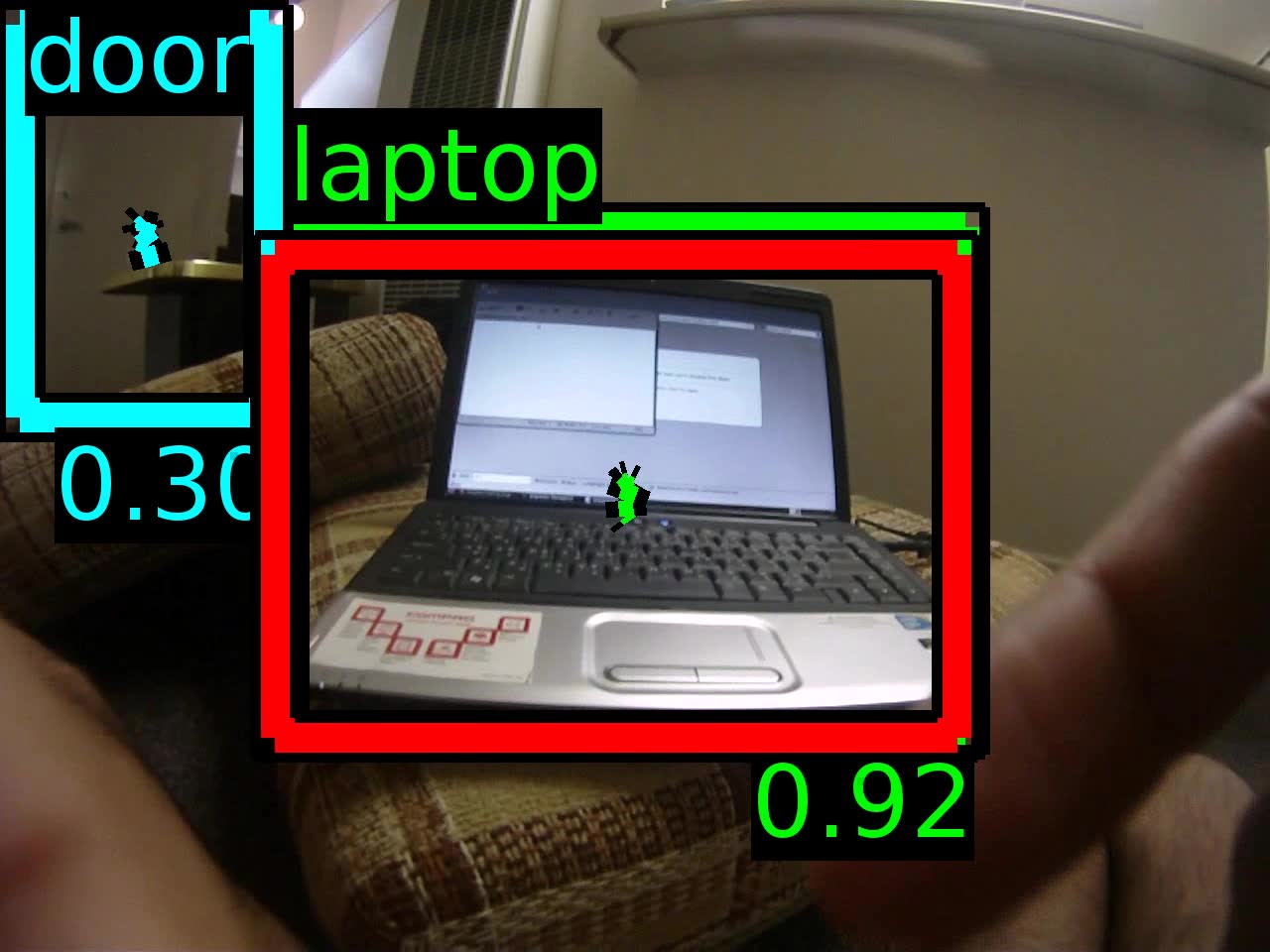} \hfill
	\includegraphics[width=0.15\linewidth]{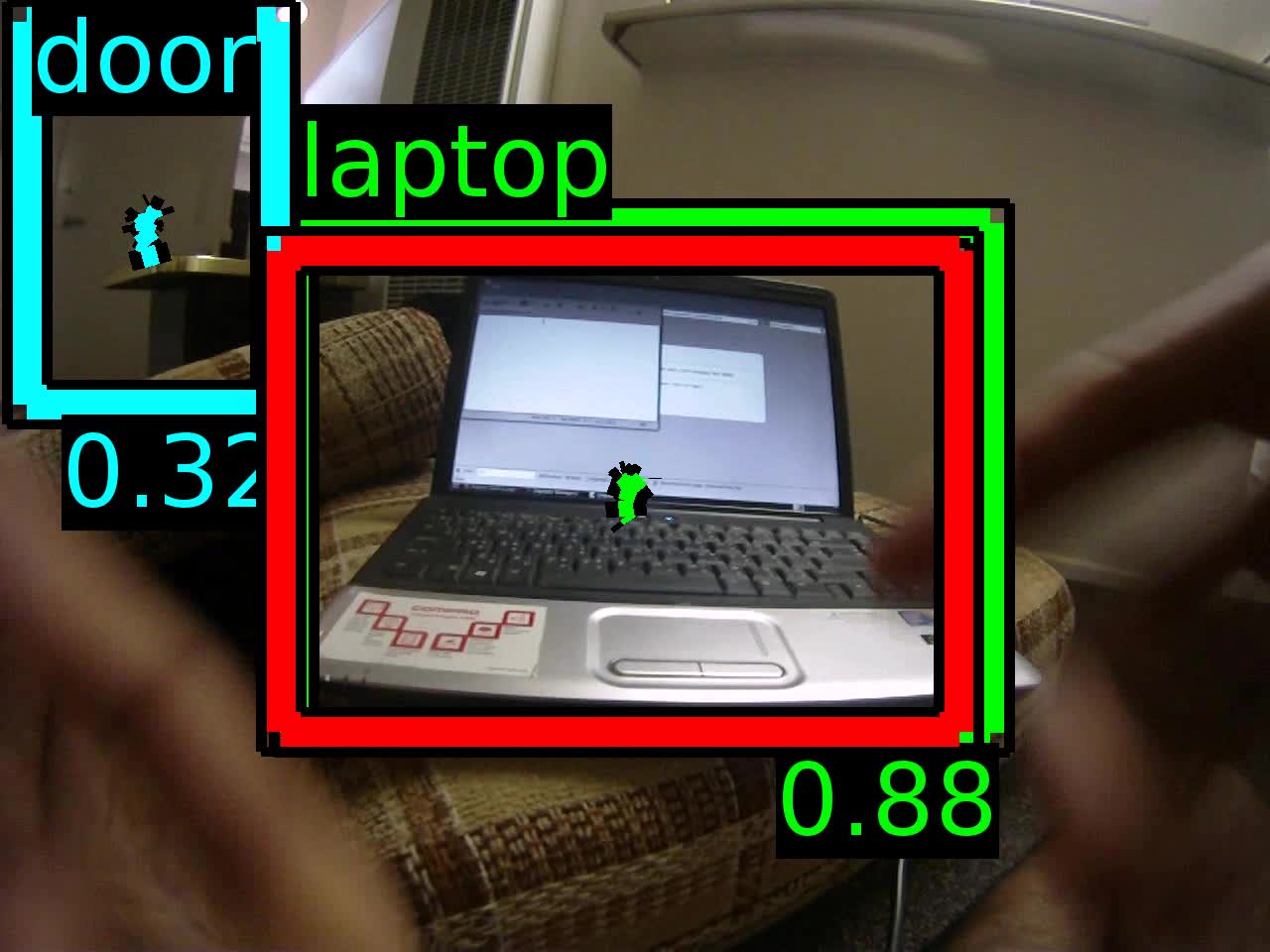} \hfill
	\includegraphics[width=0.15\linewidth]{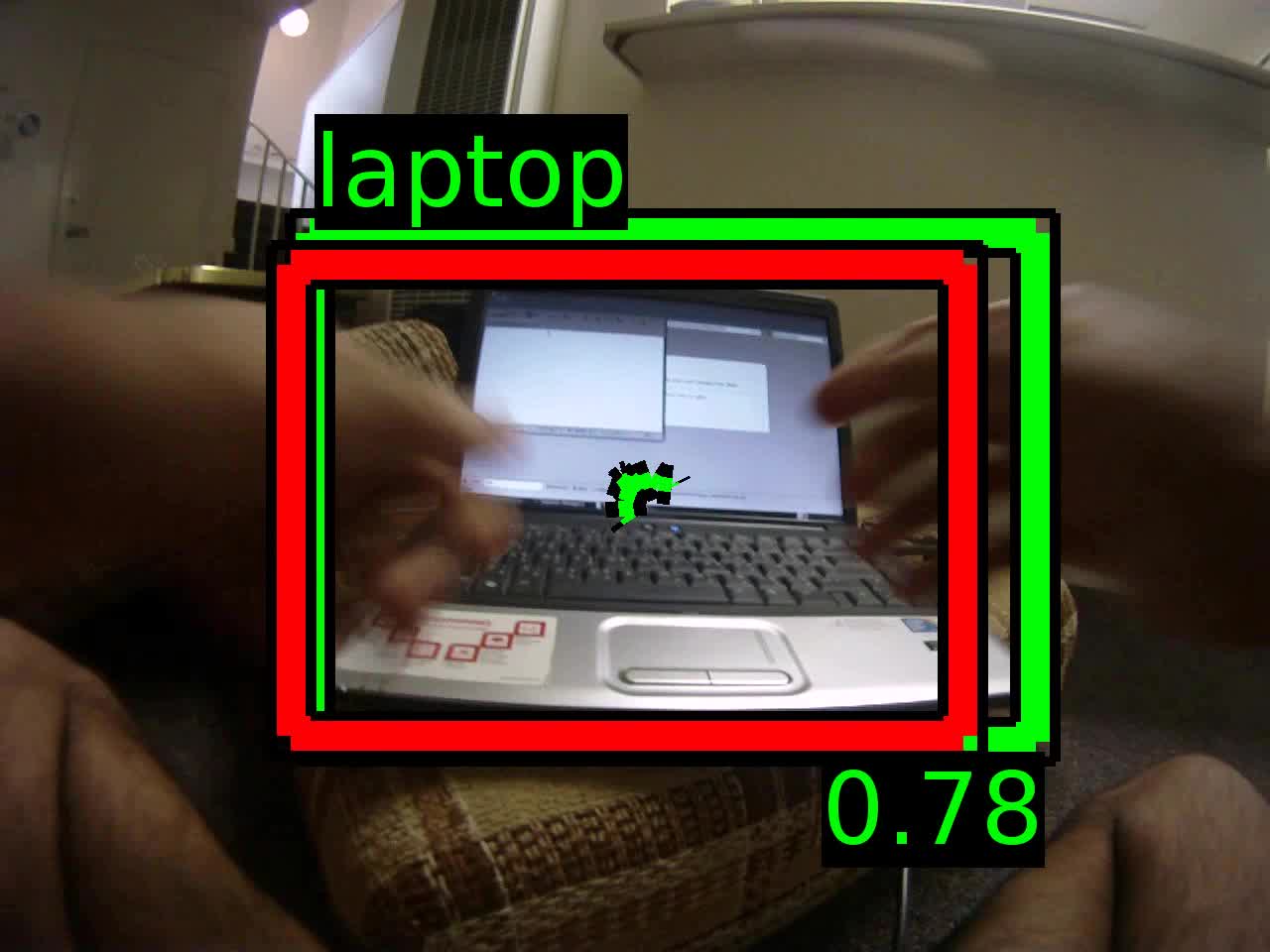} \hfill
	\includegraphics[width=0.15\linewidth]{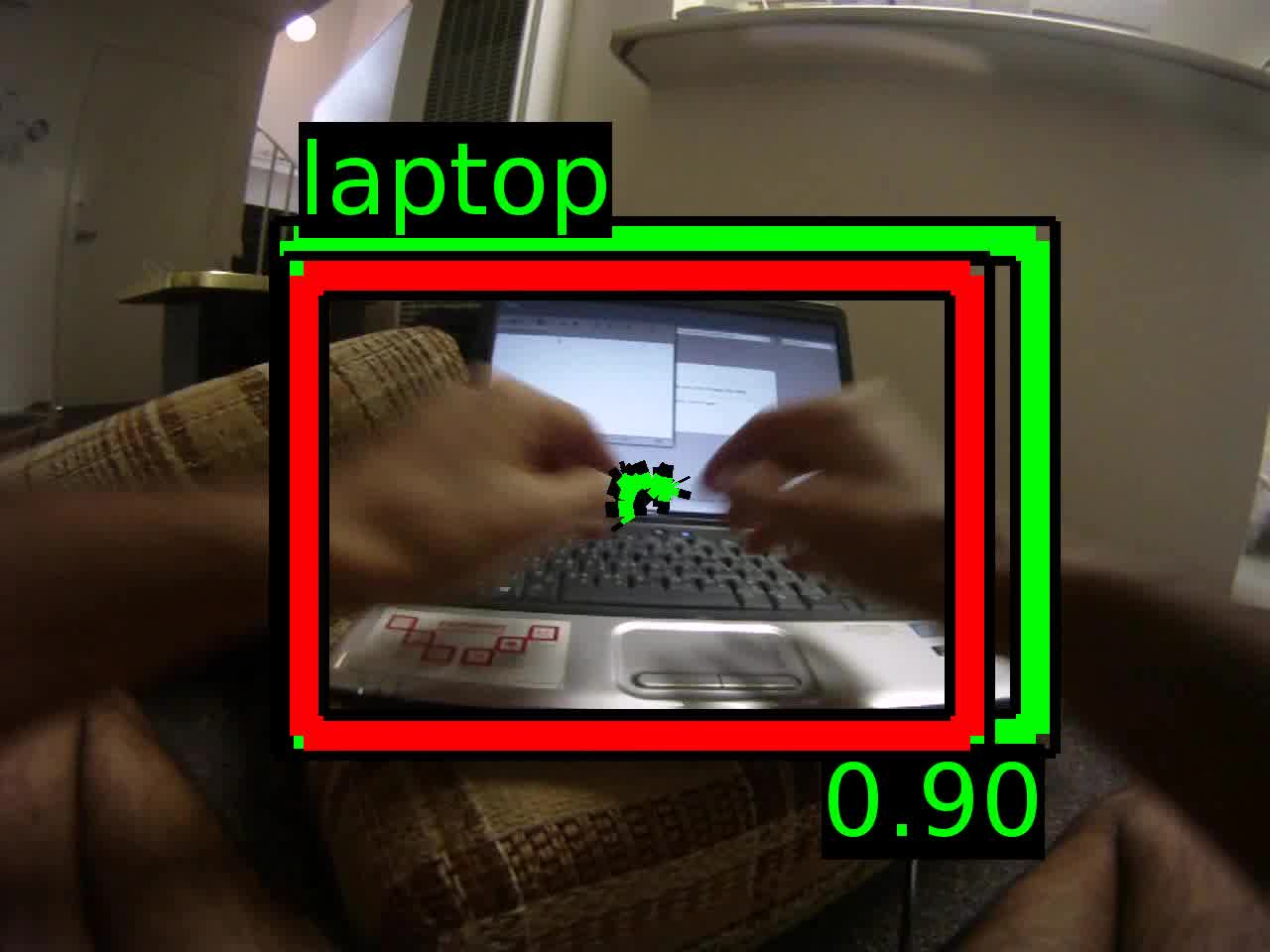} \hfill
	\includegraphics[width=0.15\linewidth]{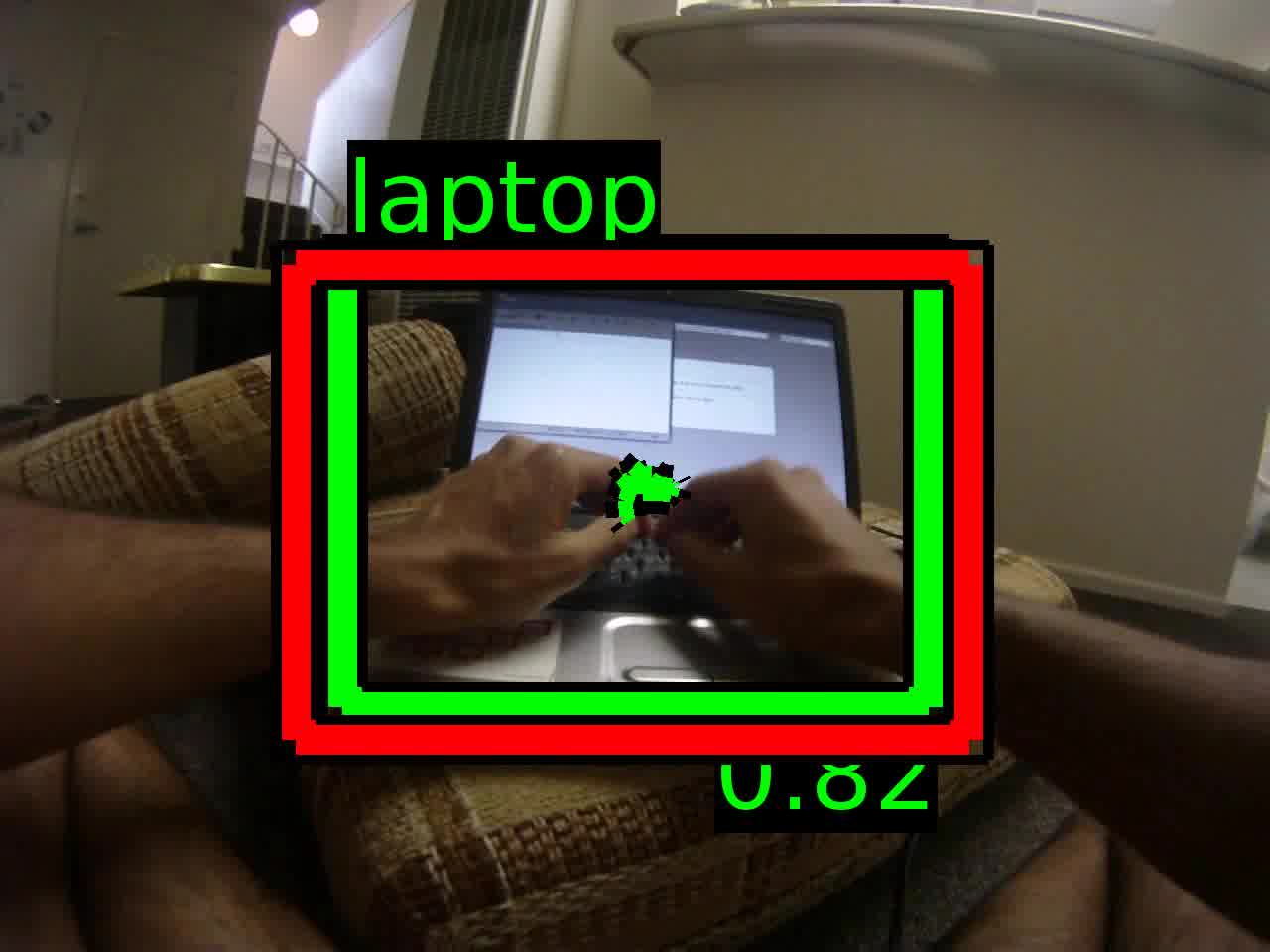}
	\vspace{1mm}
	\includegraphics[width=0.15\linewidth]{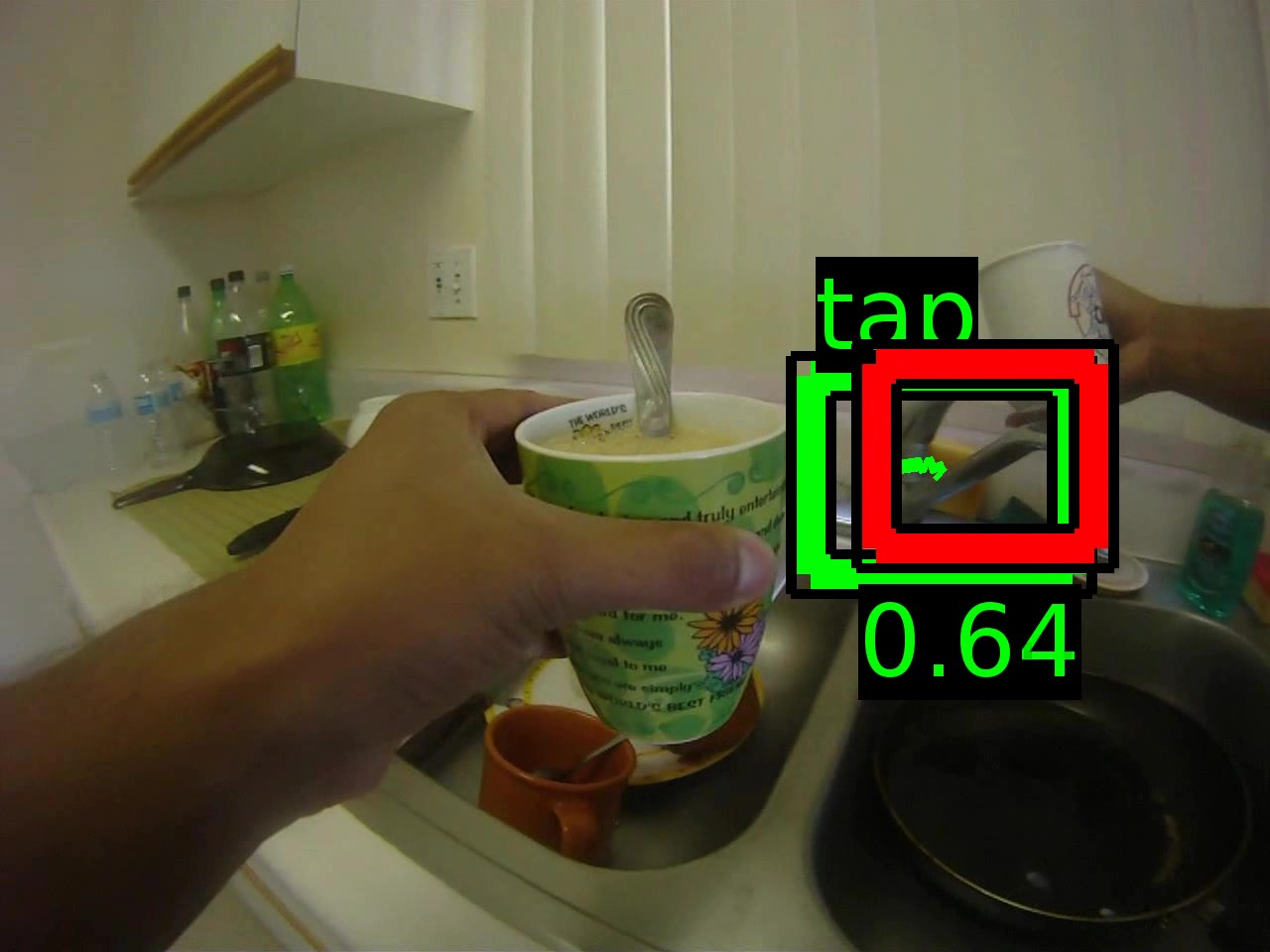} \hfill
	\includegraphics[width=0.15\linewidth]{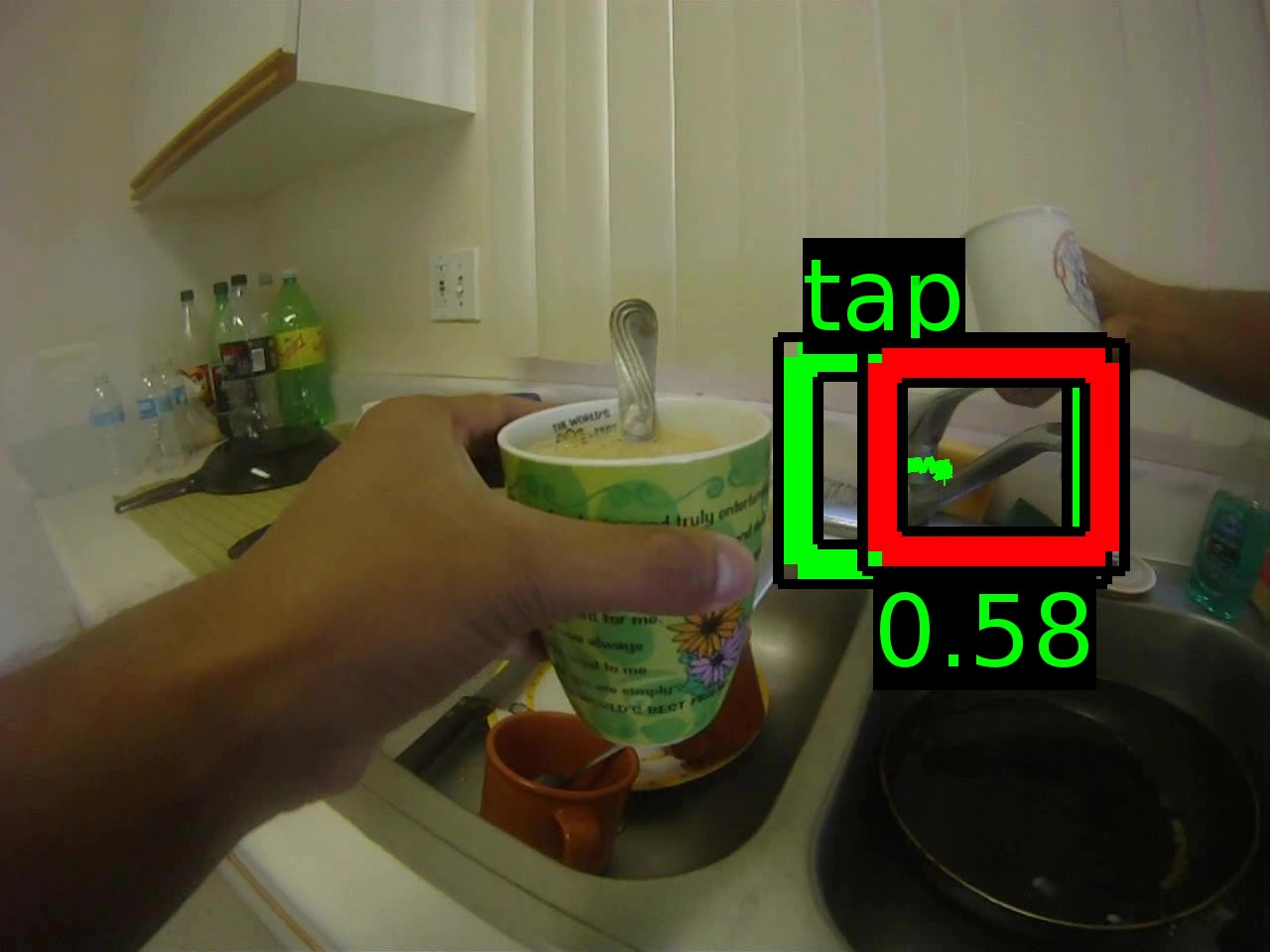} \hfill
	\includegraphics[width=0.15\linewidth]{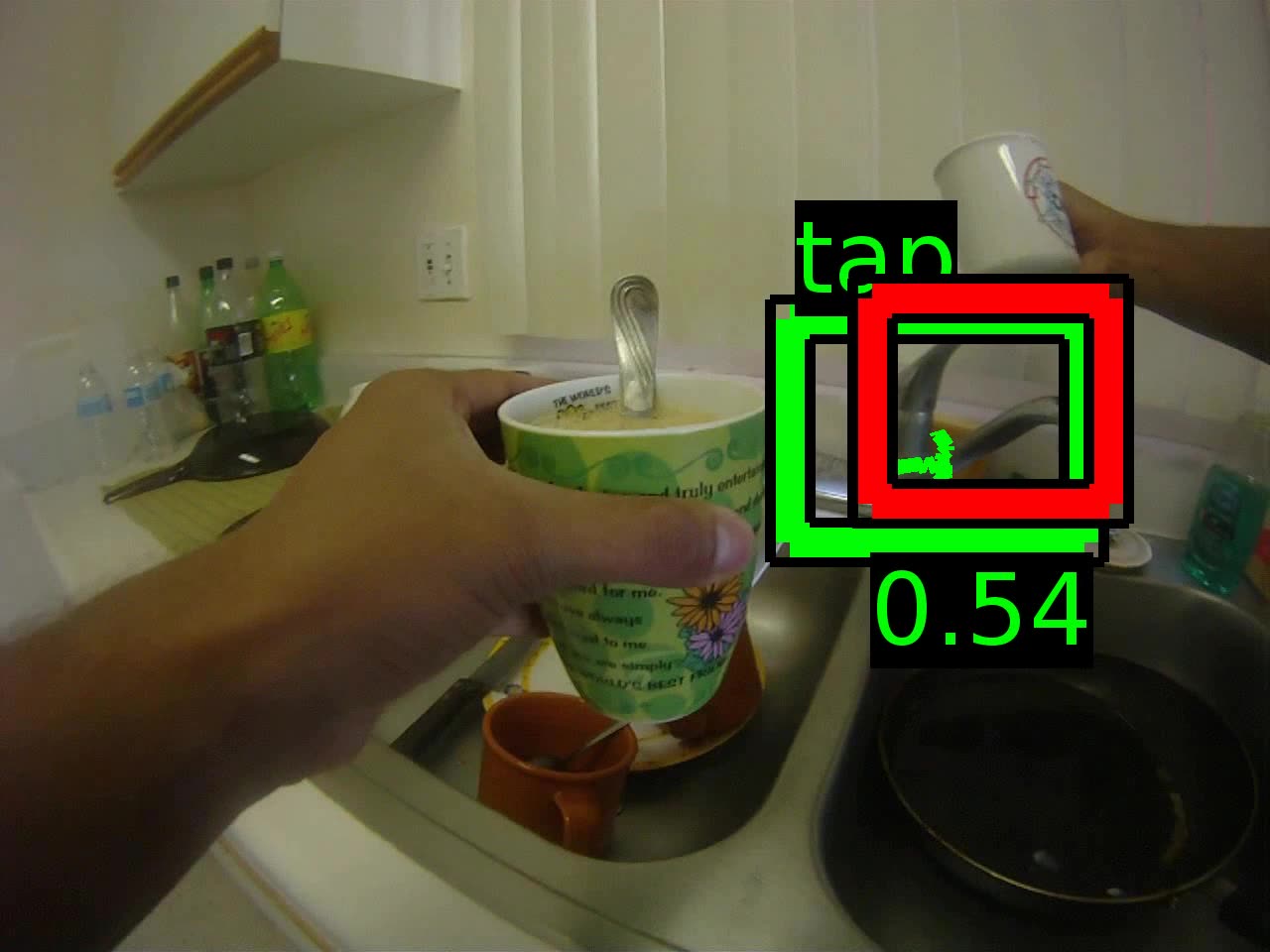} \hfill
	\includegraphics[width=0.15\linewidth]{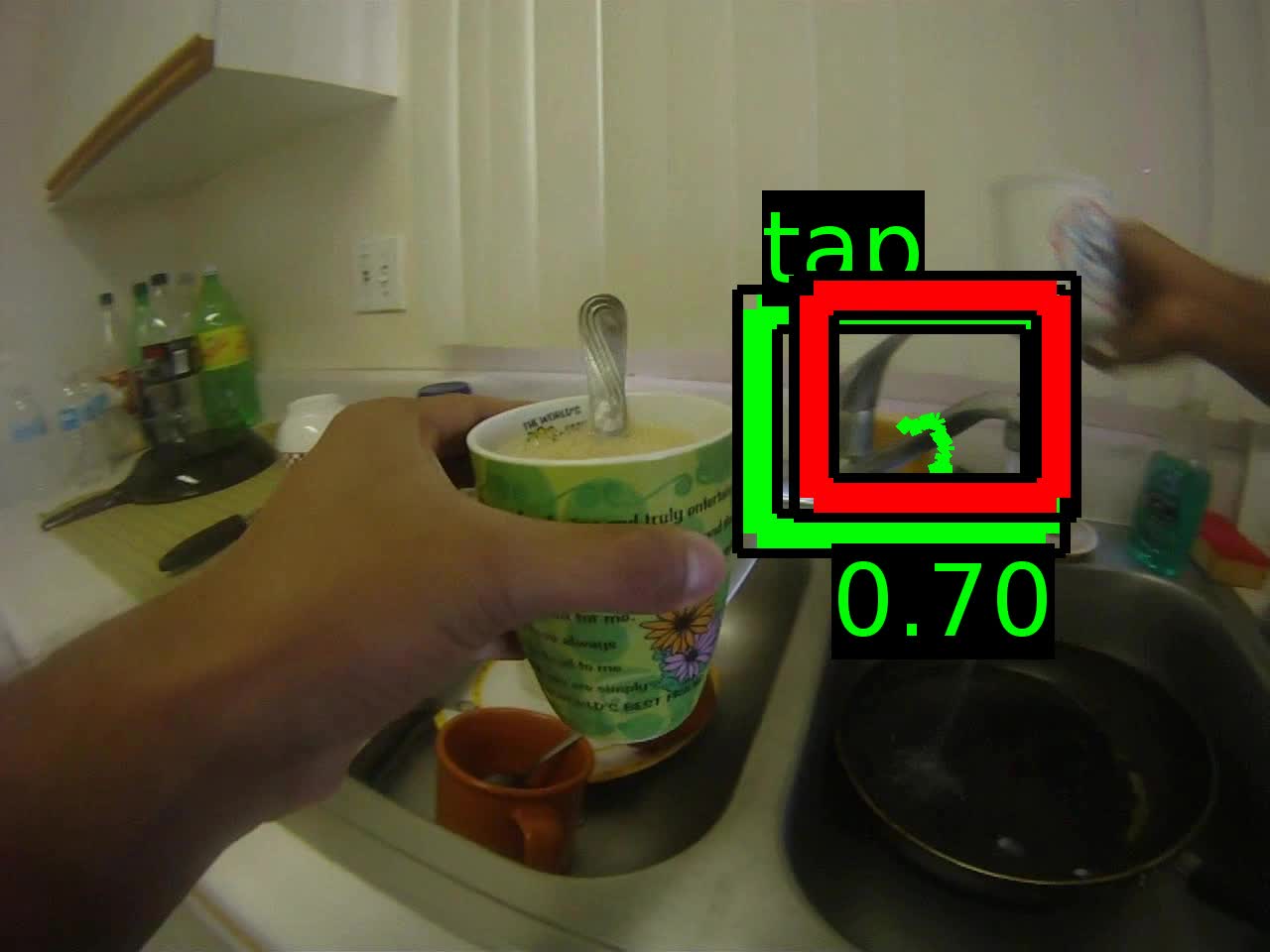} \hfill
	\includegraphics[width=0.15\linewidth]{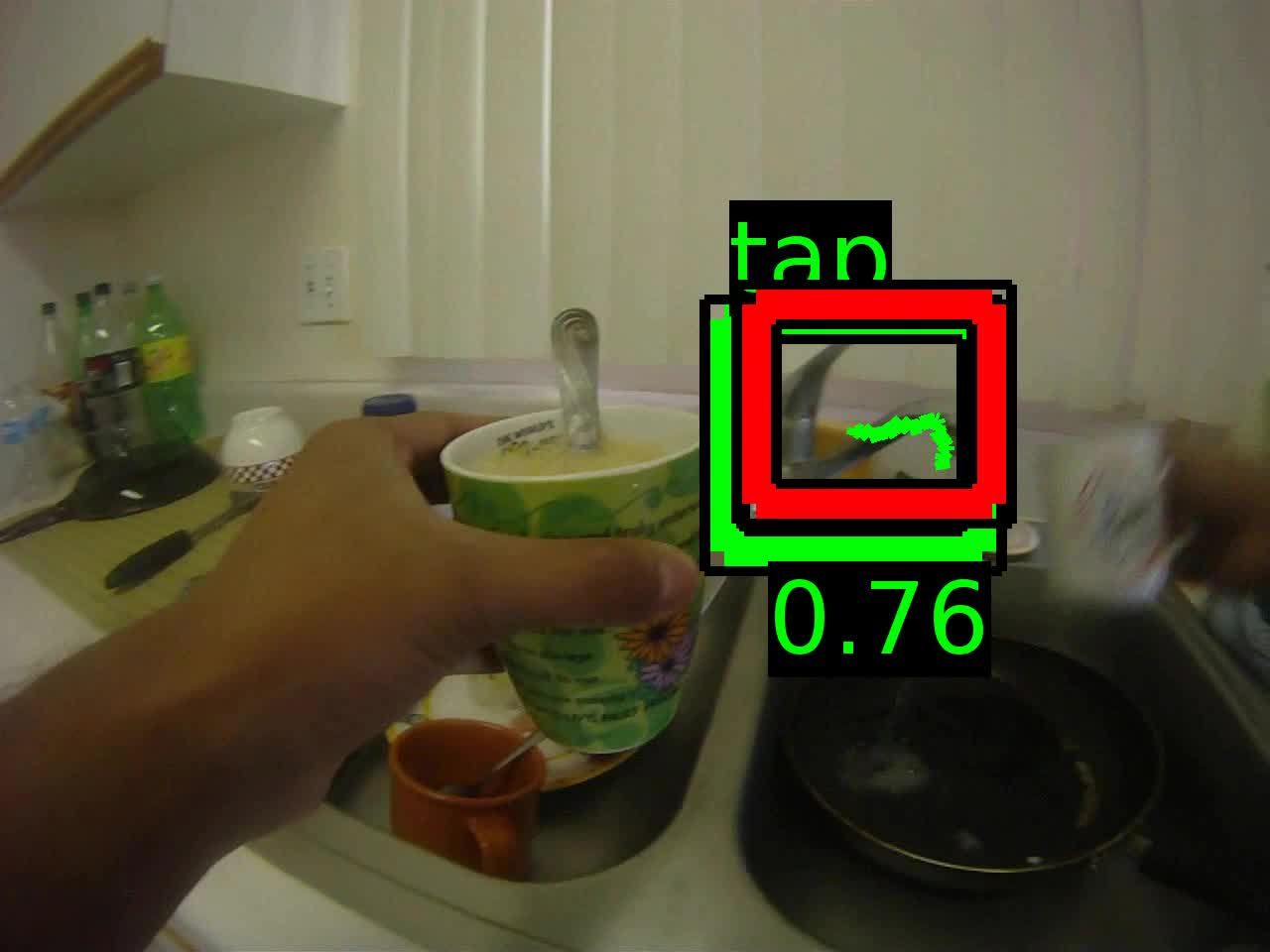} \hfill
	\includegraphics[width=0.15\linewidth]{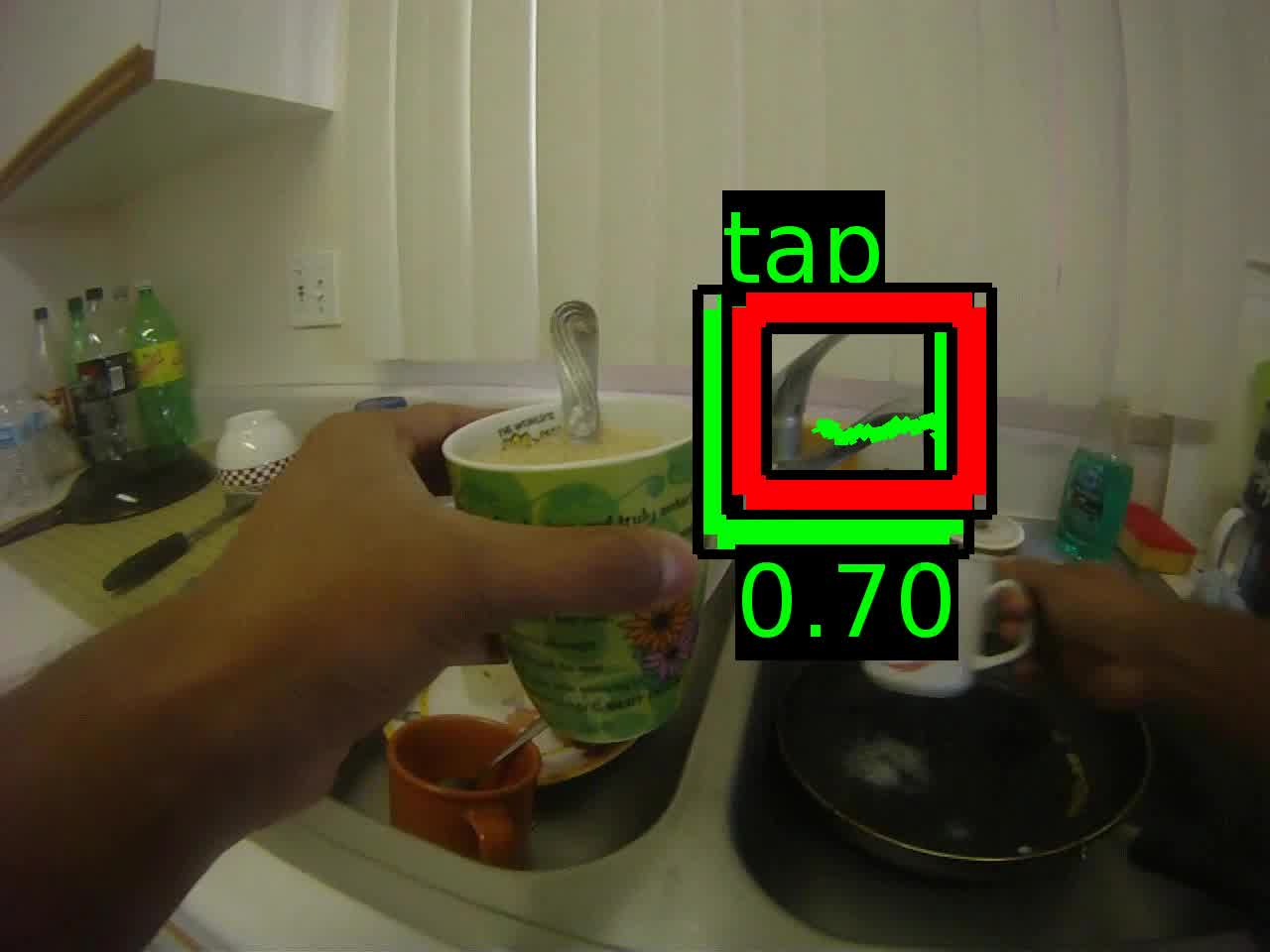}
	
	\vspace{1mm}
	\includegraphics[width=0.15\linewidth]{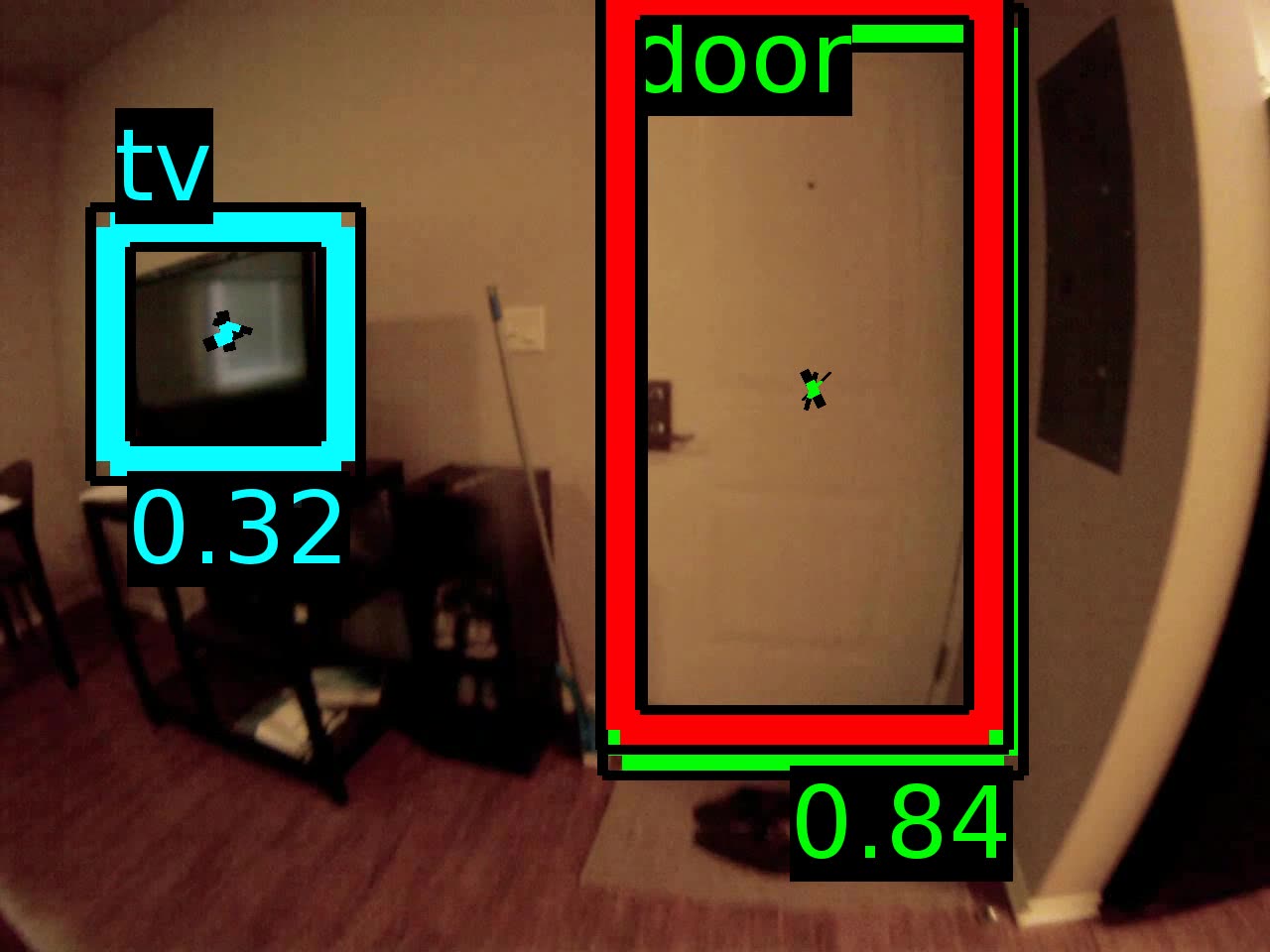} \hfill
	\includegraphics[width=0.15\linewidth]{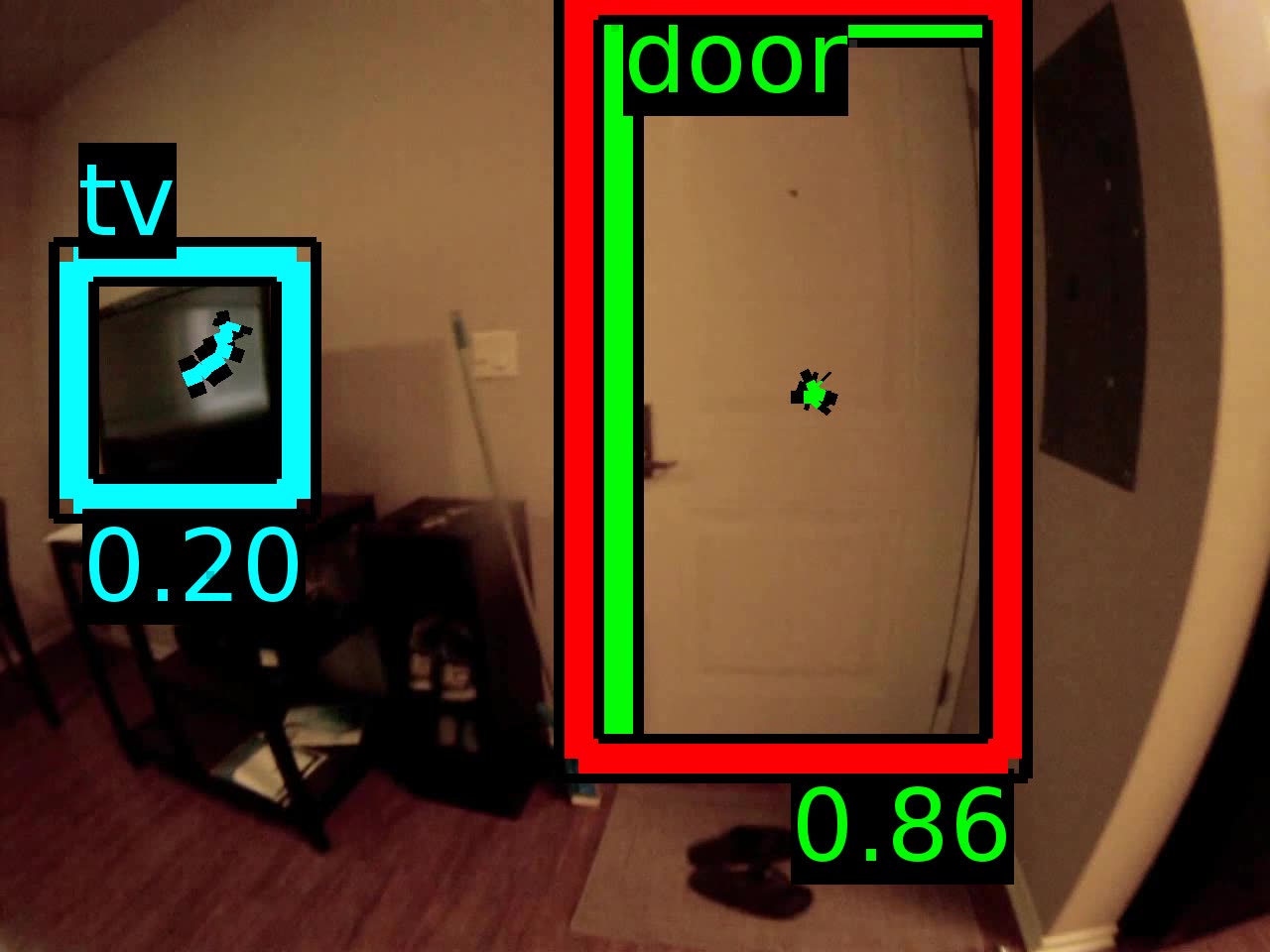} \hfill
	\includegraphics[width=0.15\linewidth]{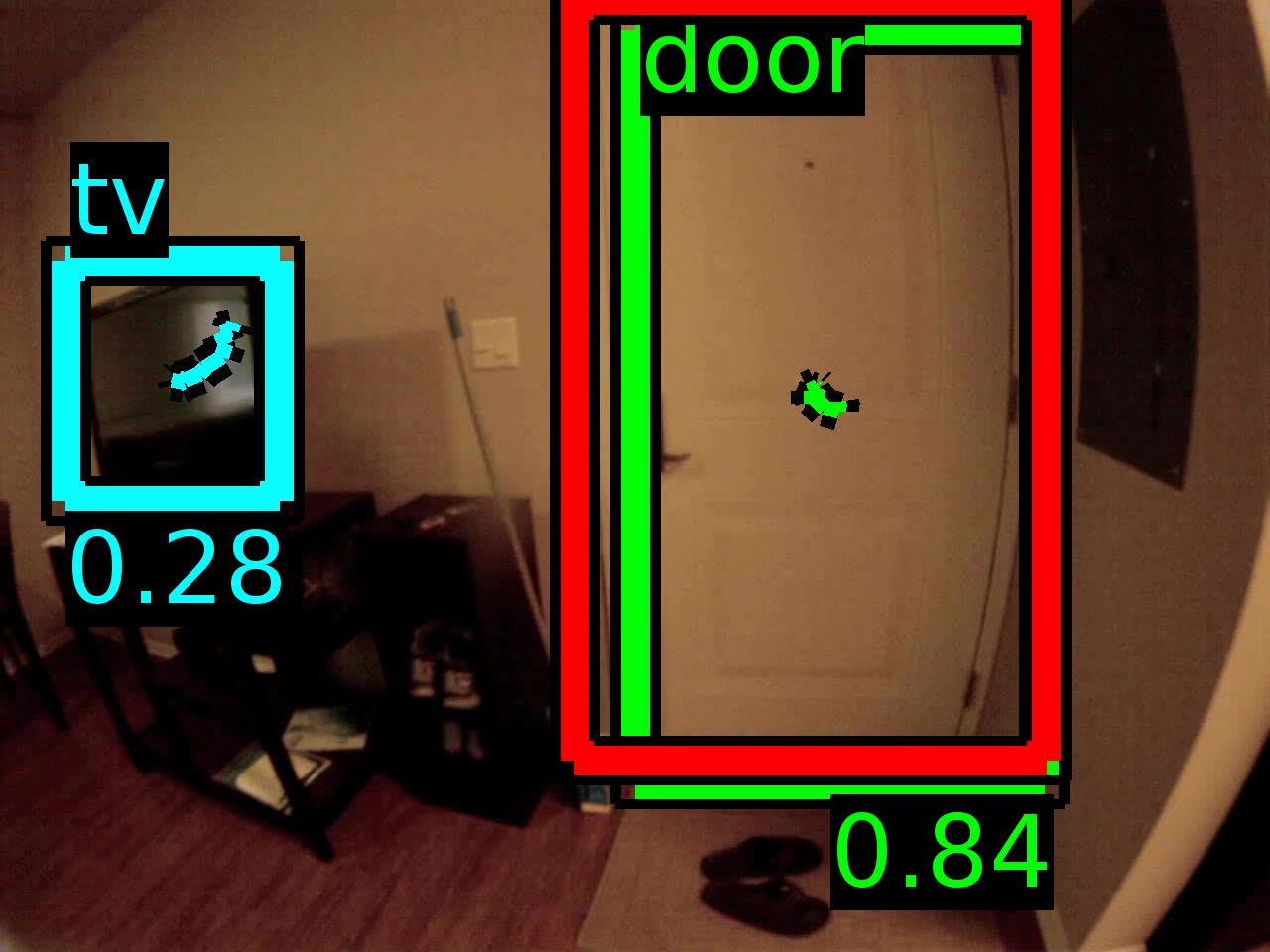} \hfill
	\includegraphics[width=0.15\linewidth]{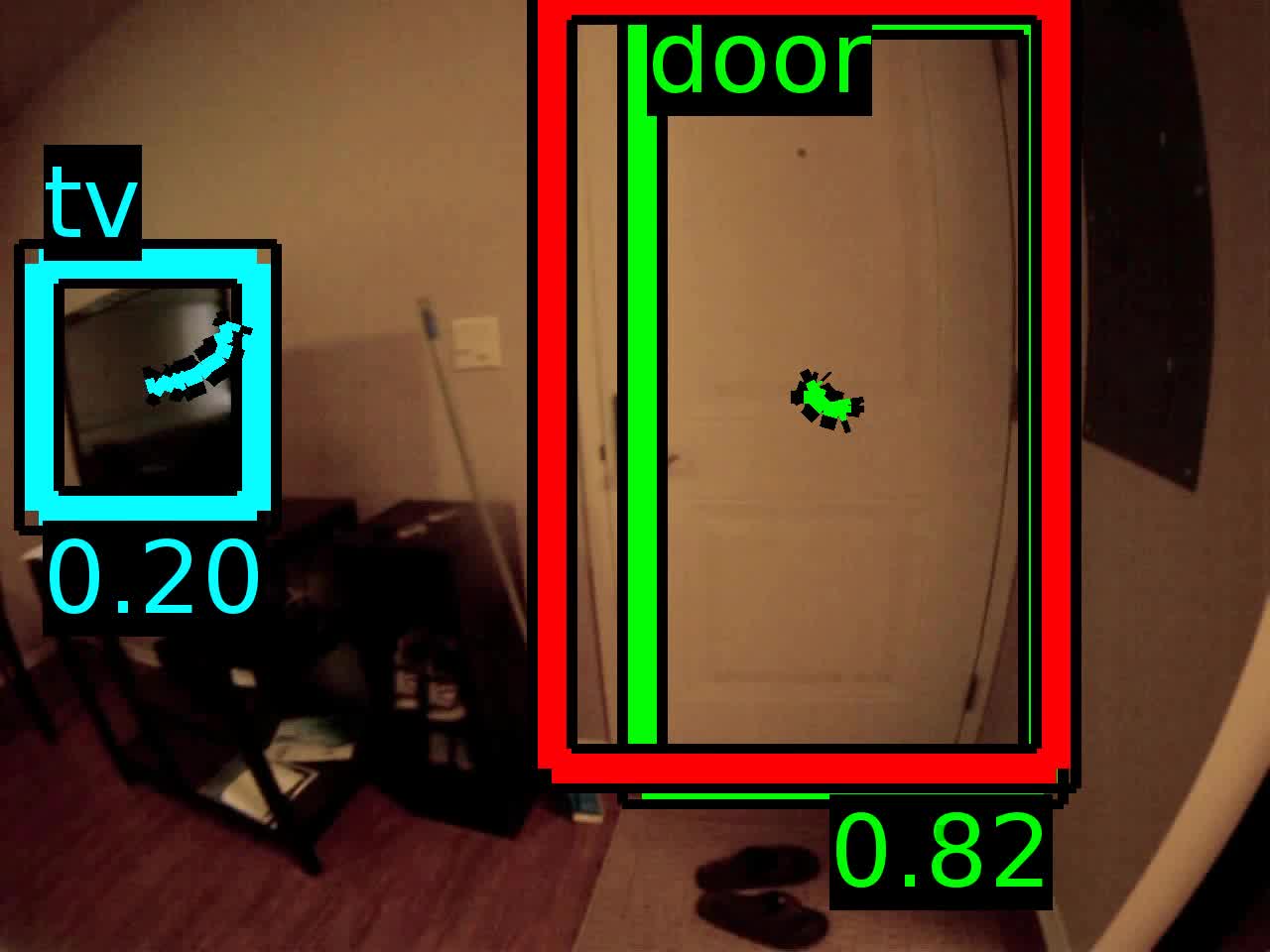} \hfill
	\includegraphics[width=0.15\linewidth]{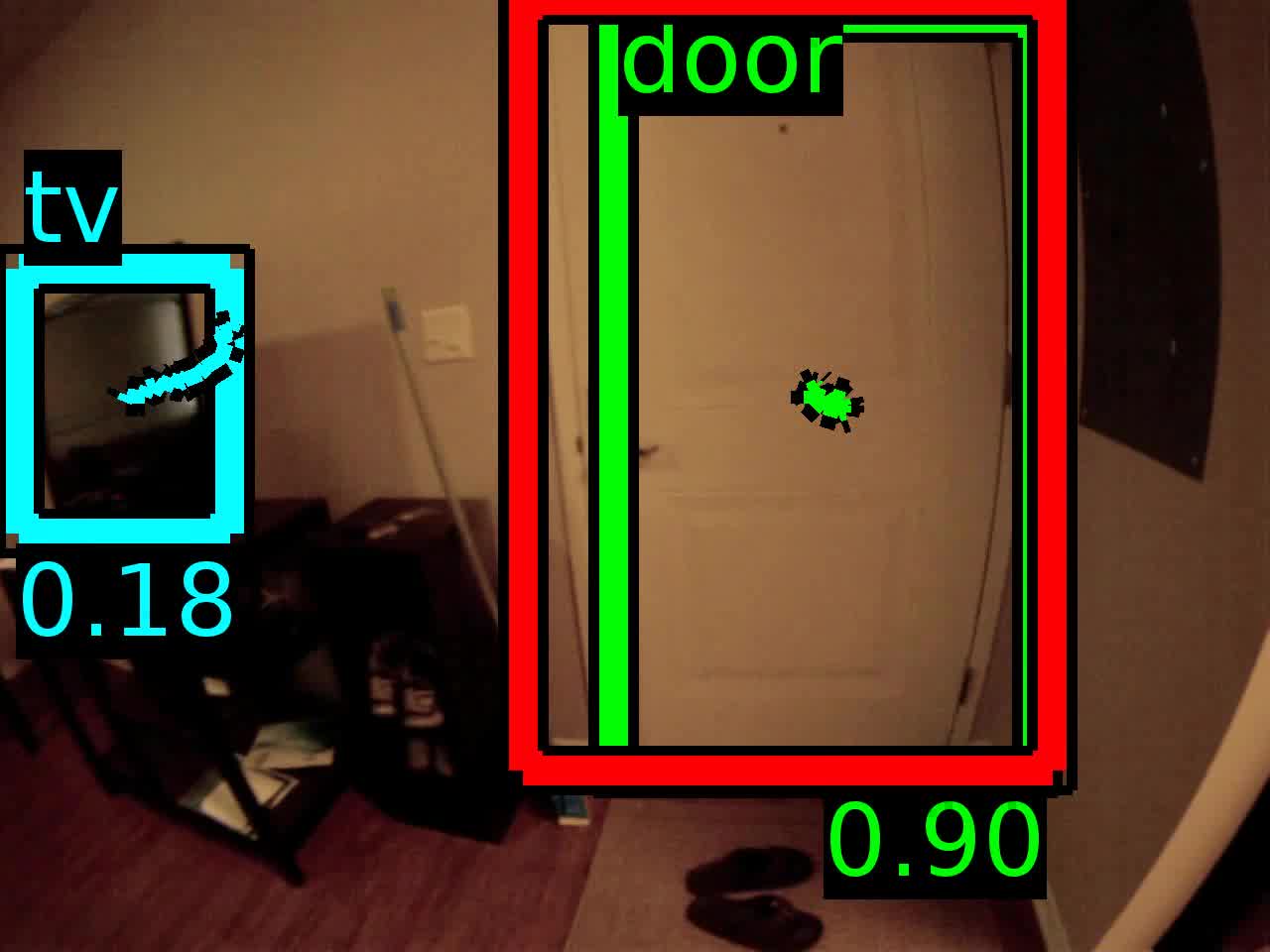} \hfill
	\includegraphics[width=0.15\linewidth]{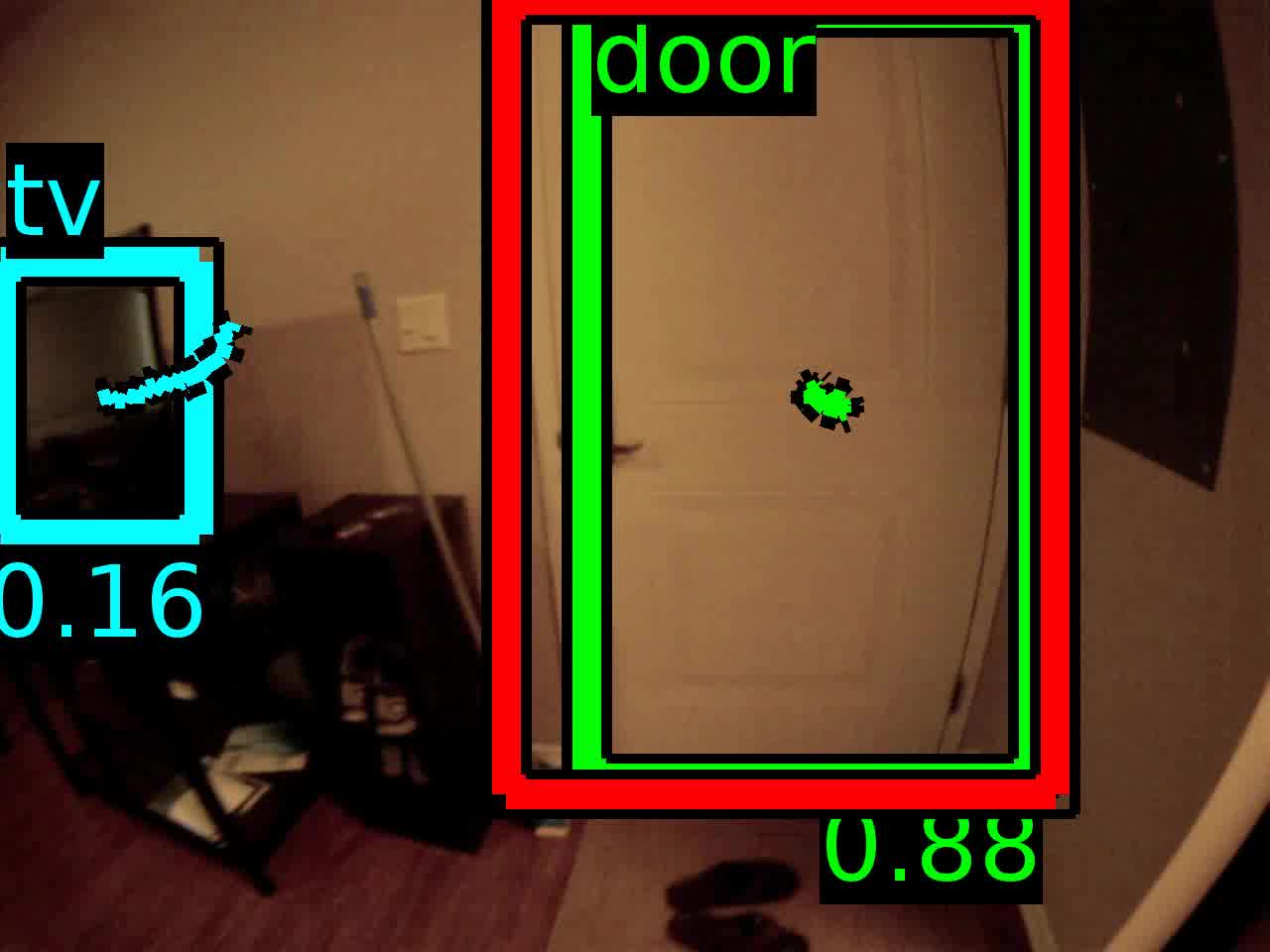}
	
	\vspace{1mm}		
	\includegraphics[width=0.15\linewidth]{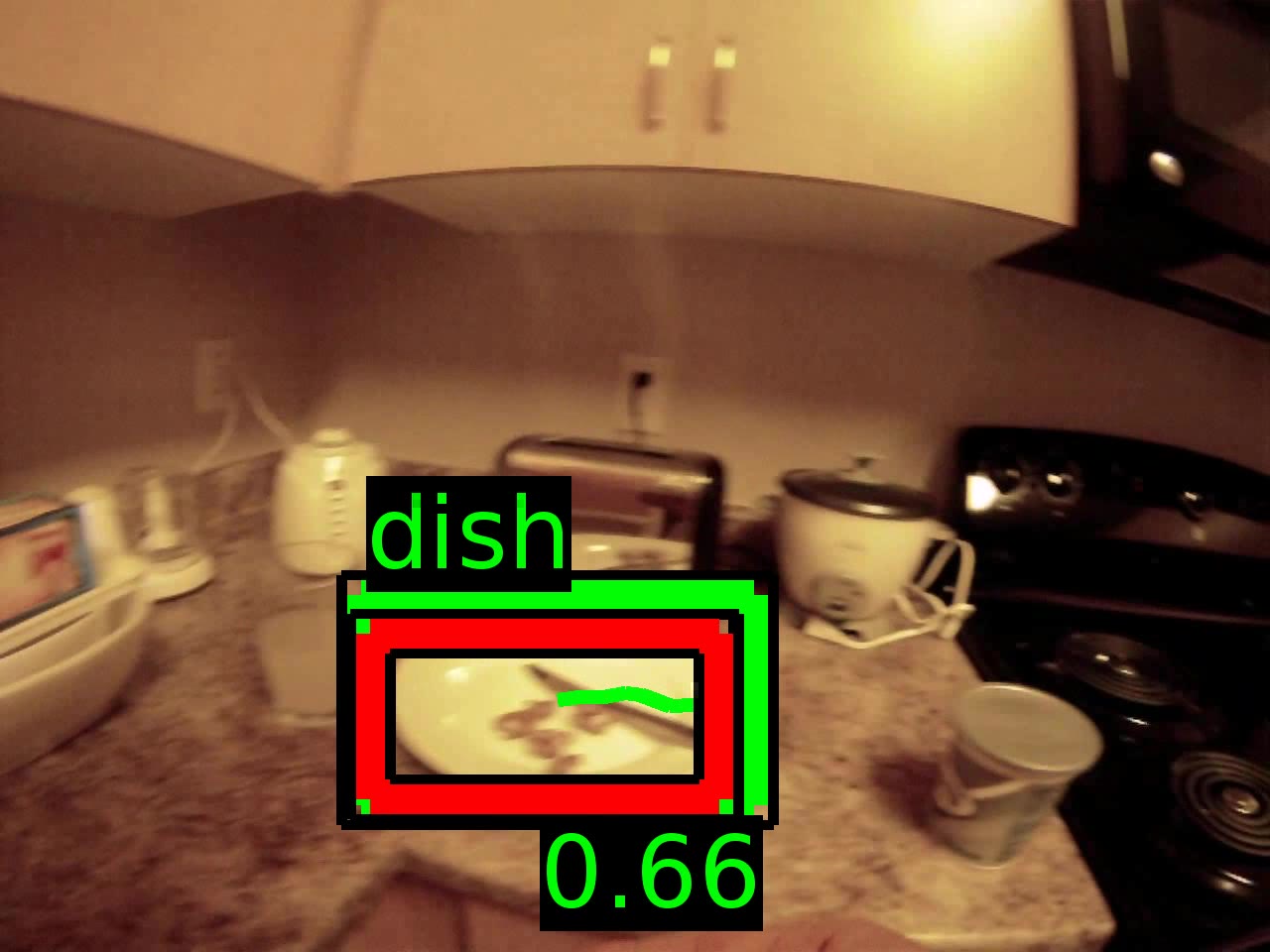} \hfill
	\includegraphics[width=0.15\linewidth]{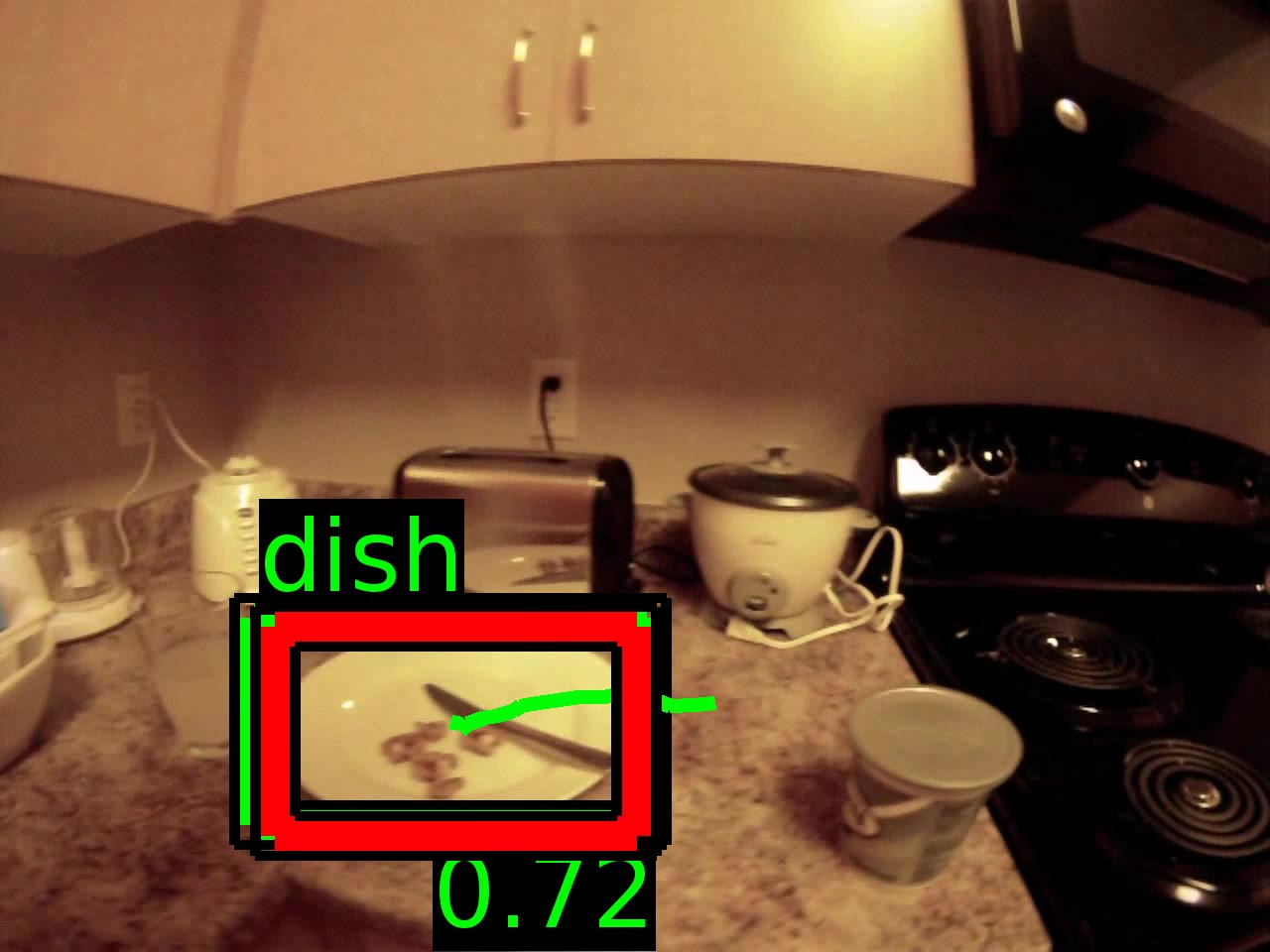} \hfill
	\includegraphics[width=0.15\linewidth]{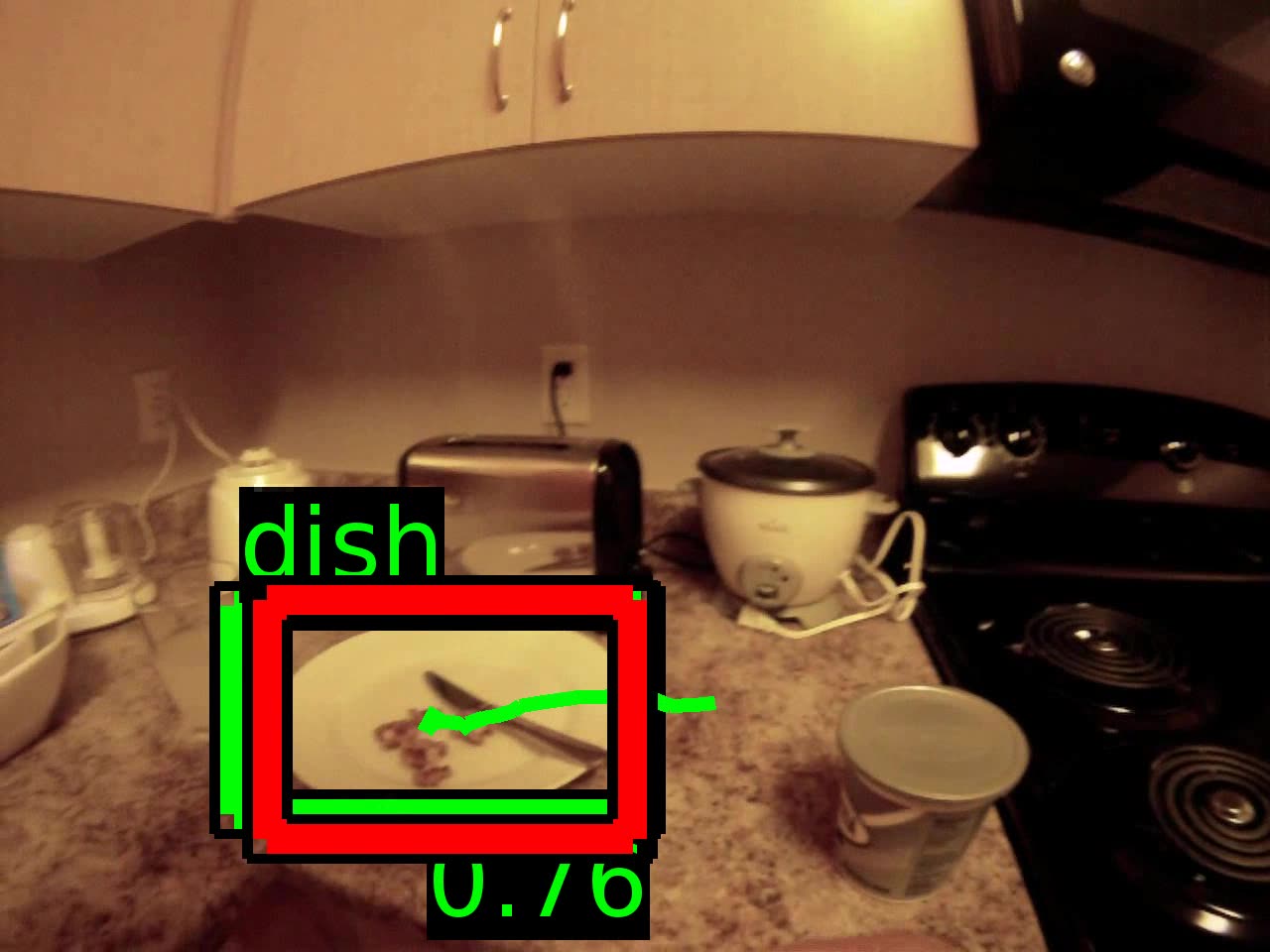} \hfill
	\includegraphics[width=0.15\linewidth]{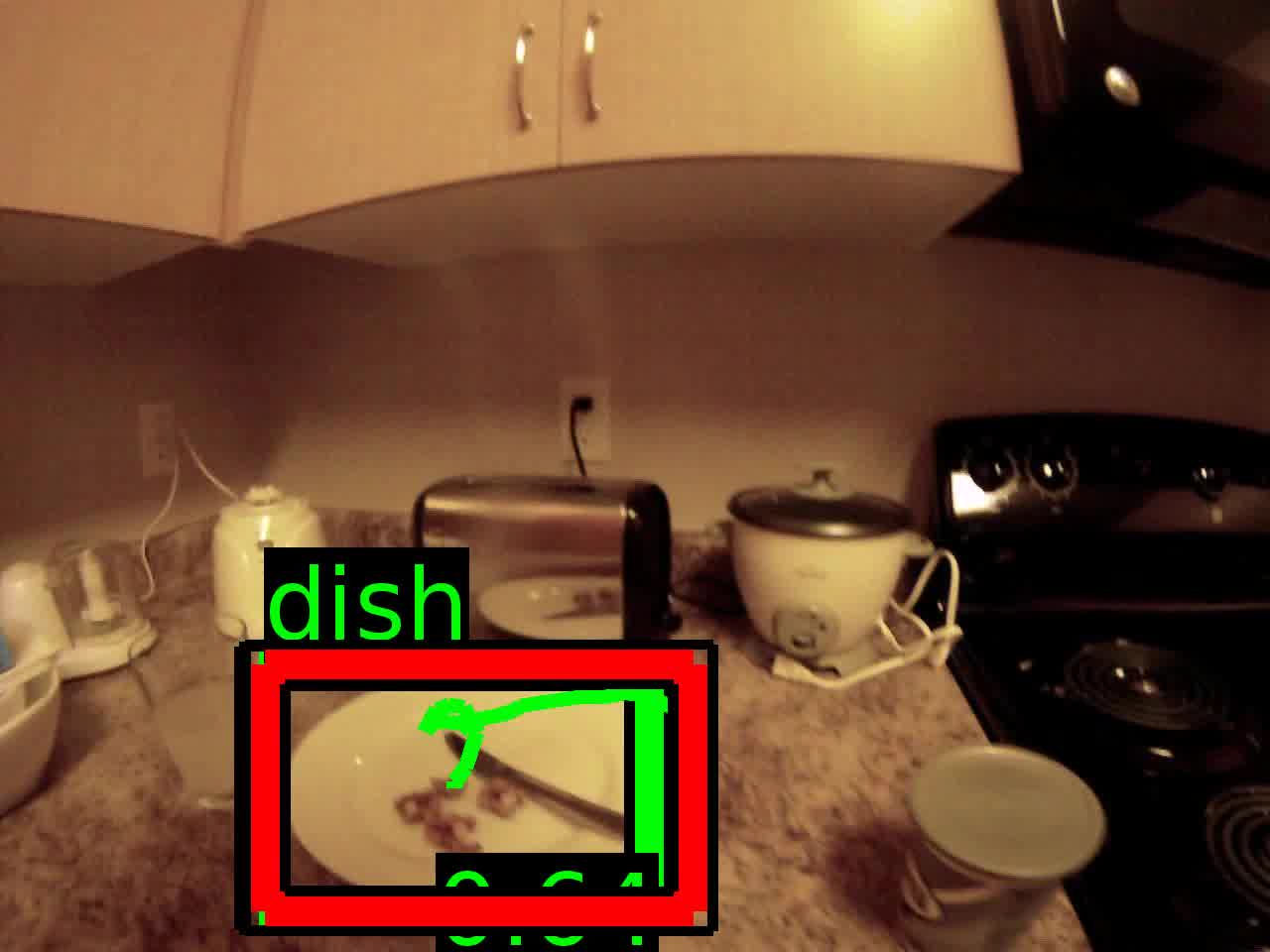} \hfill
	\includegraphics[width=0.15\linewidth]{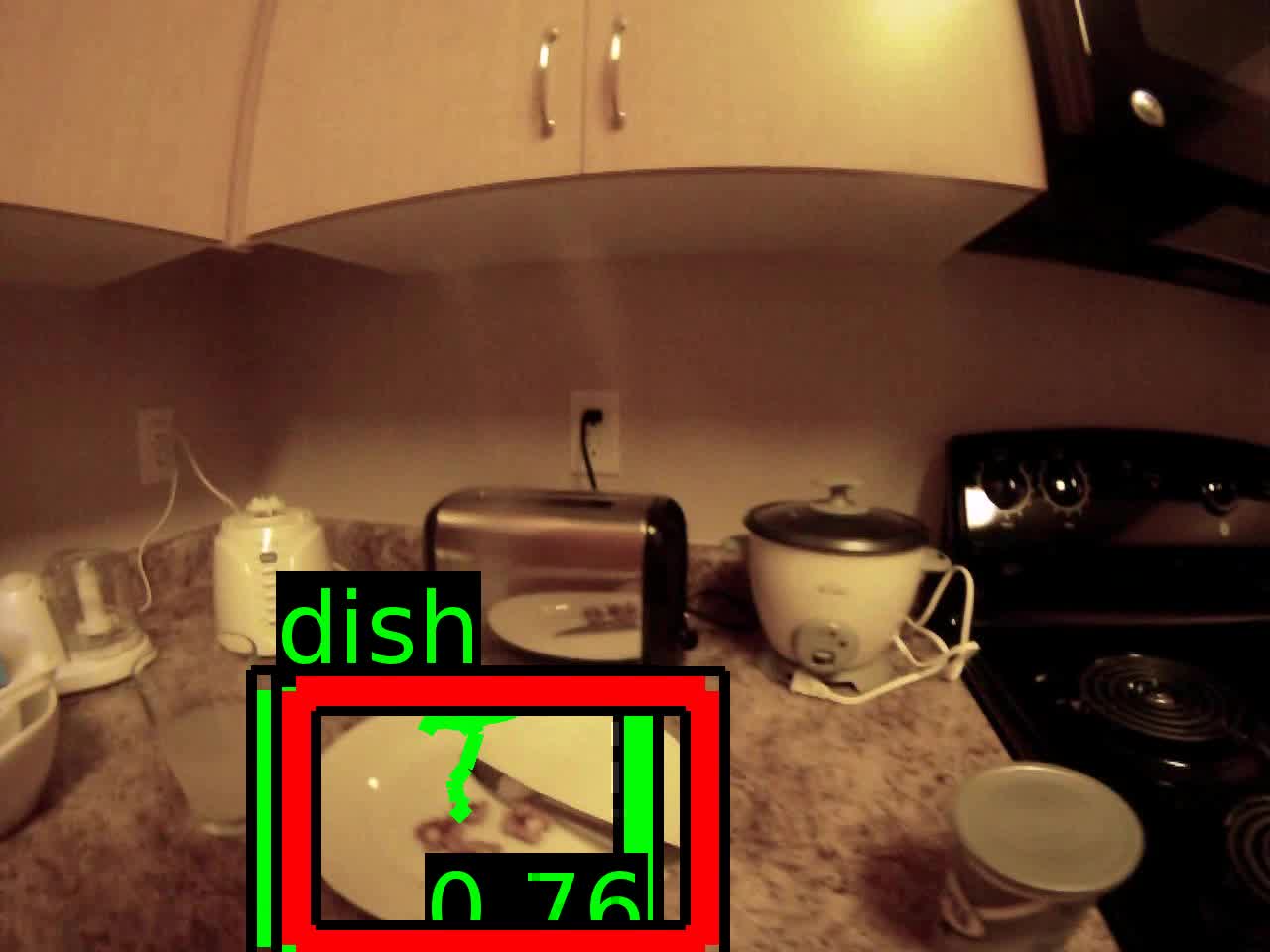} \hfill
	\includegraphics[width=0.15\linewidth]{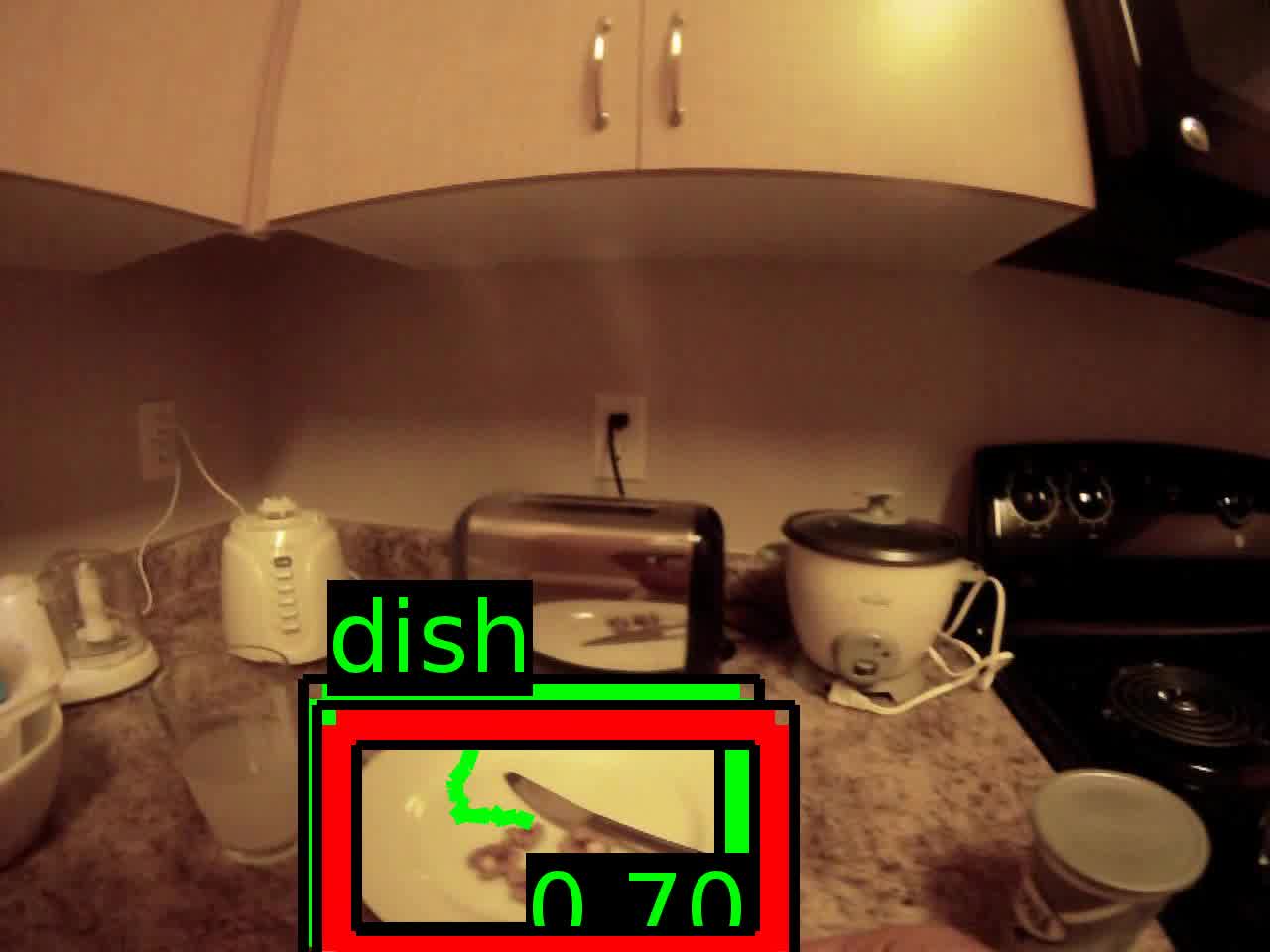}
	
	\vspace{1mm}
	\includegraphics[width=0.15\linewidth]{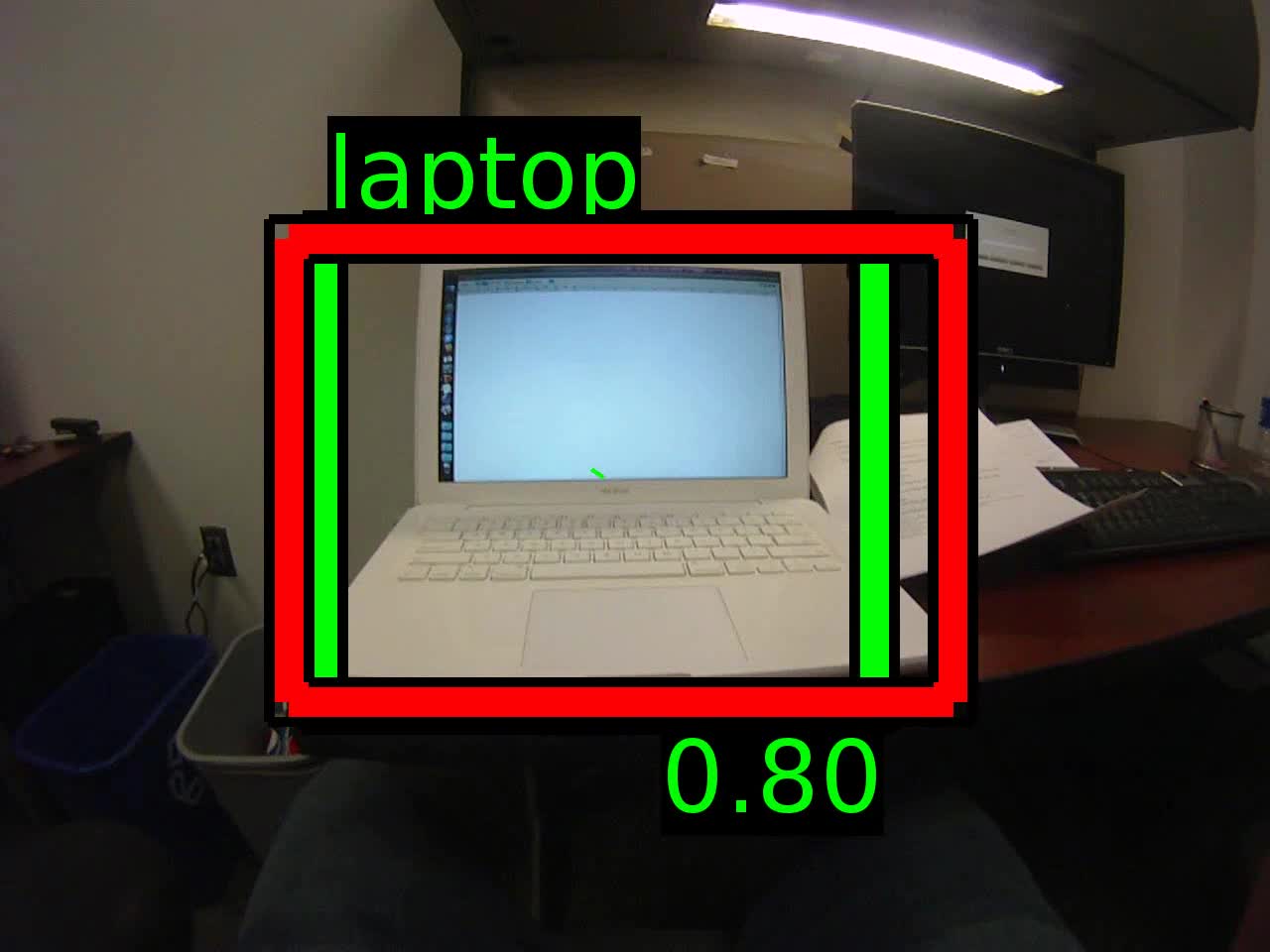} \hfill
	\includegraphics[width=0.15\linewidth]{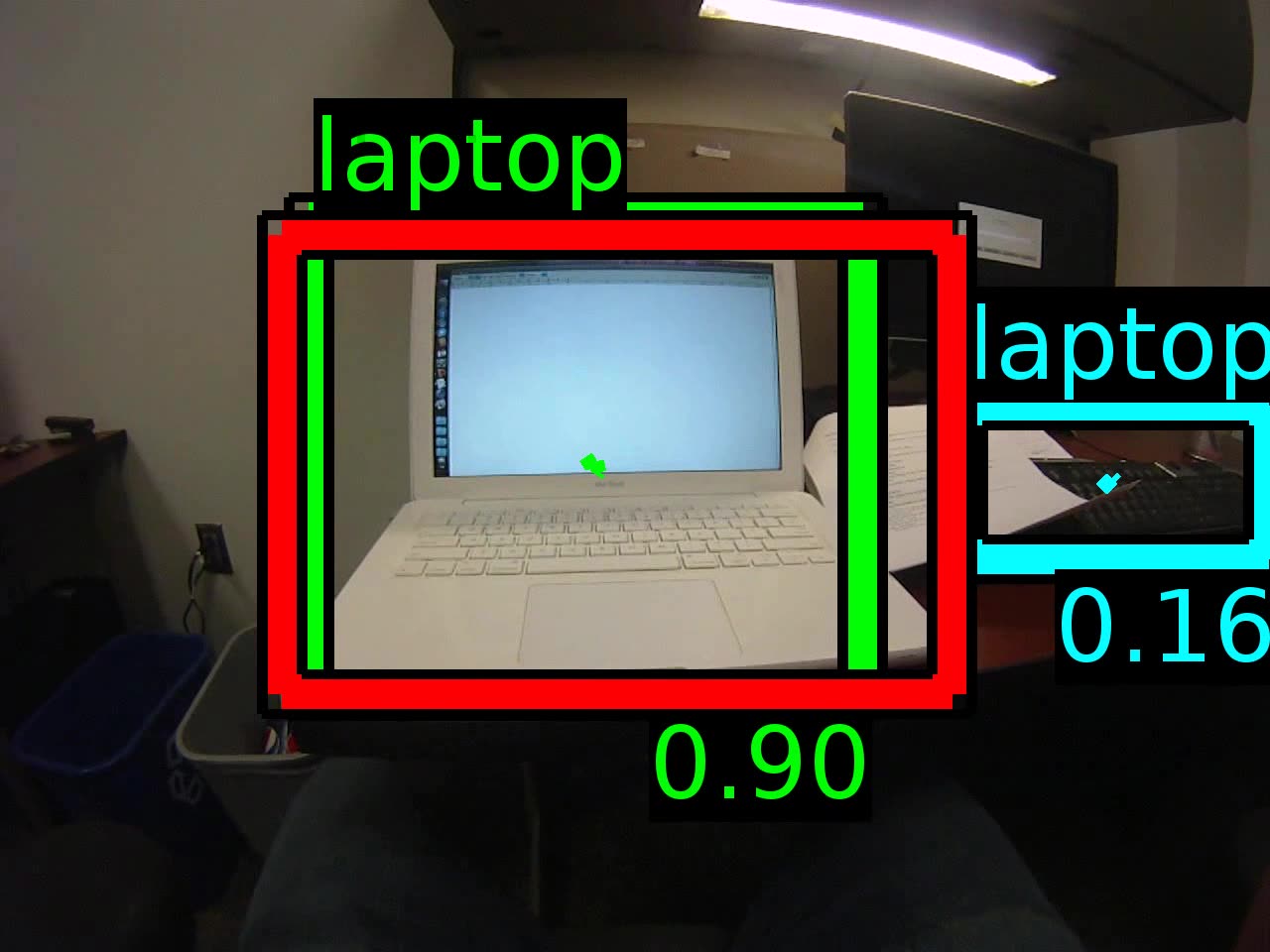} \hfill
	\includegraphics[width=0.15\linewidth]{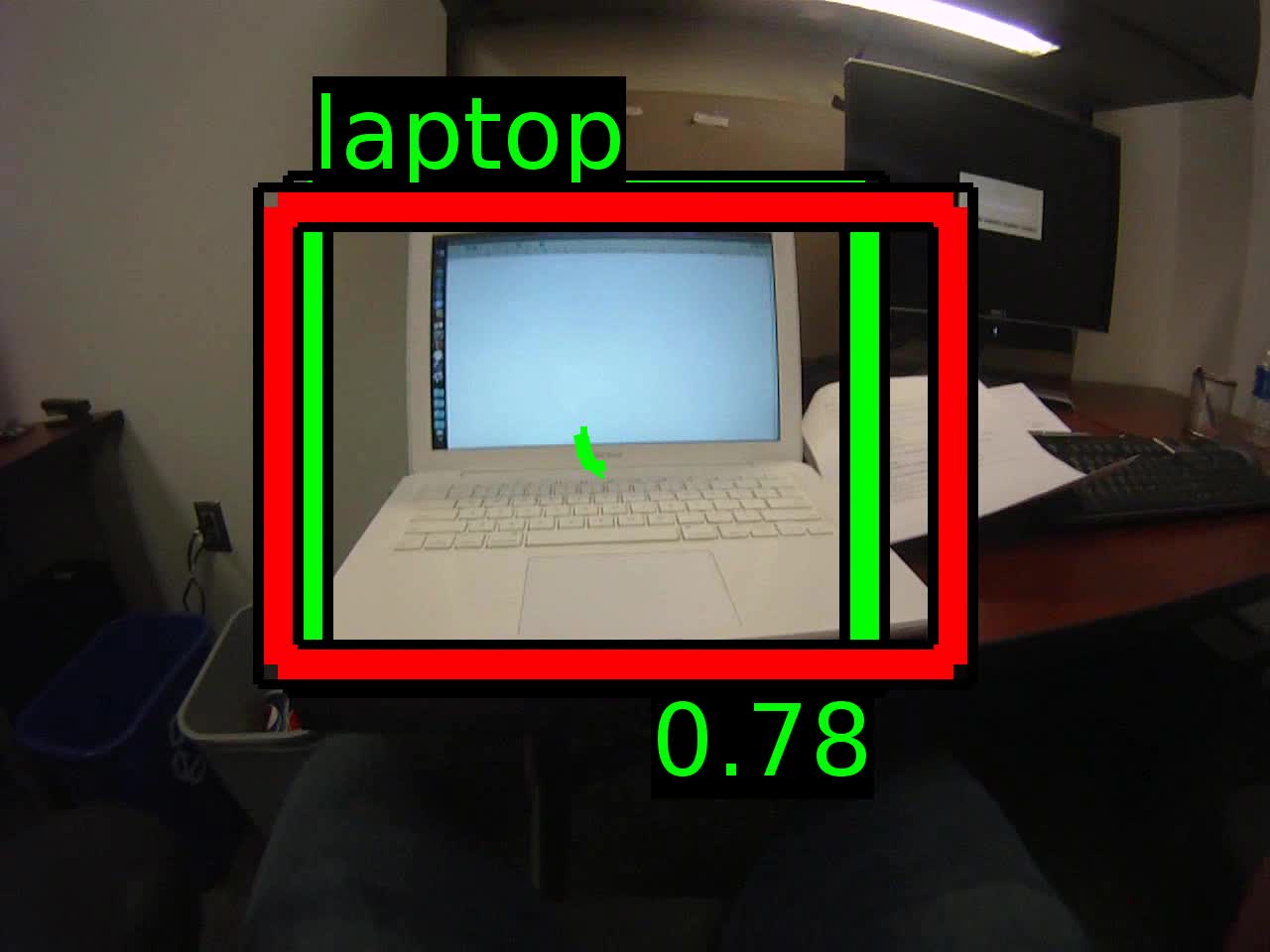} \hfill
	\includegraphics[width=0.15\linewidth]{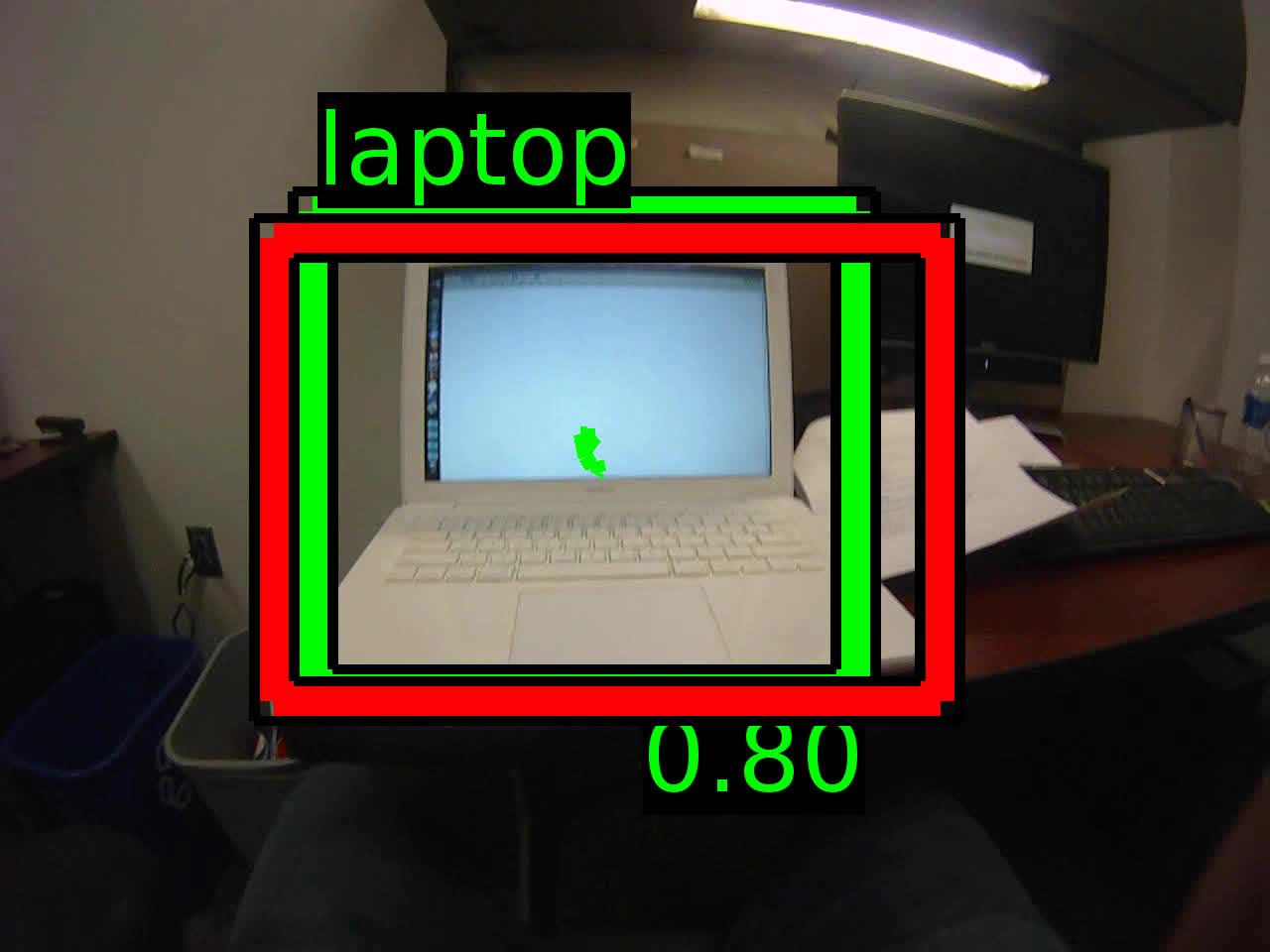} \hfill
	\includegraphics[width=0.15\linewidth]{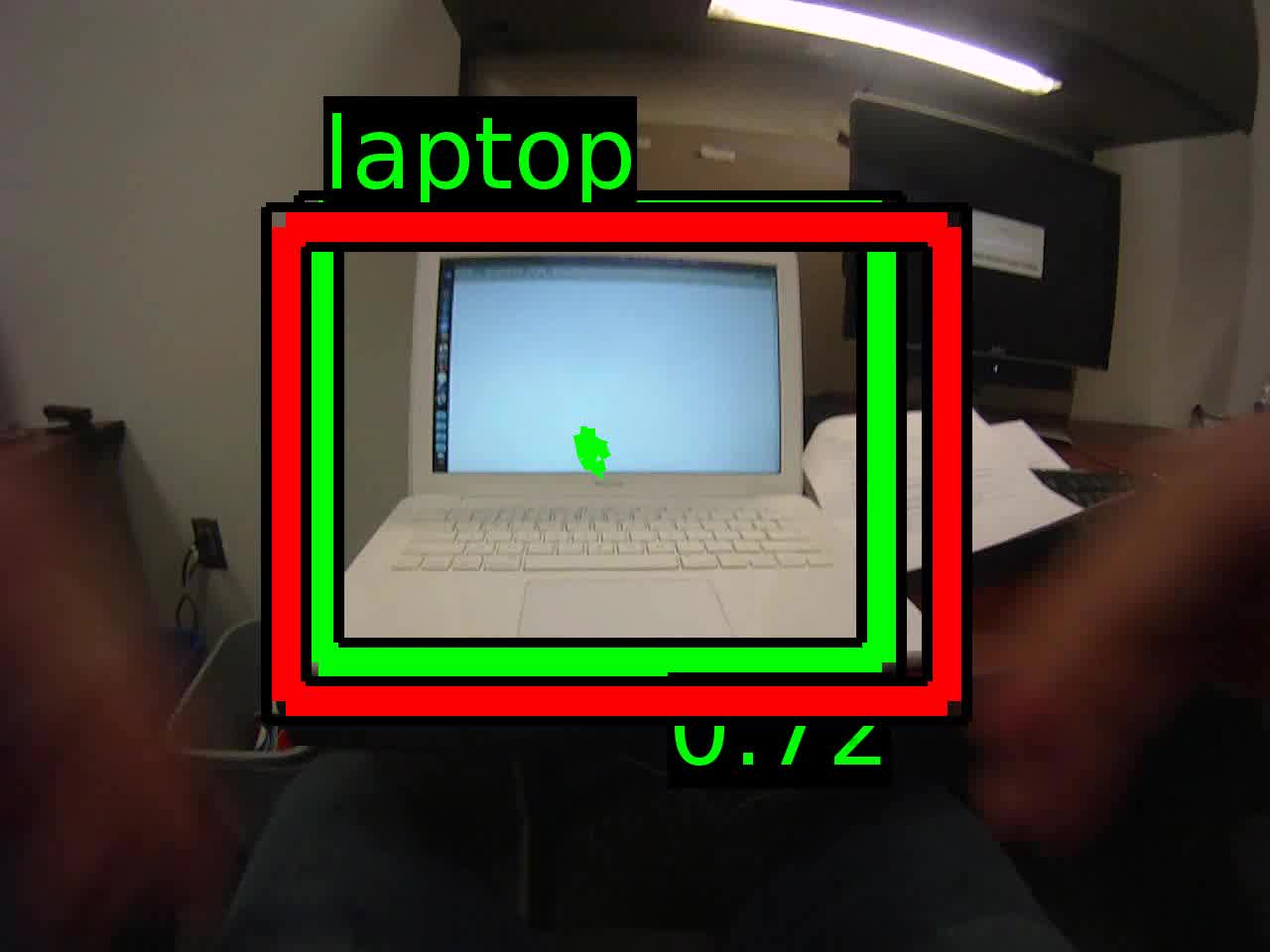} \hfill
	\includegraphics[width=0.15\linewidth]{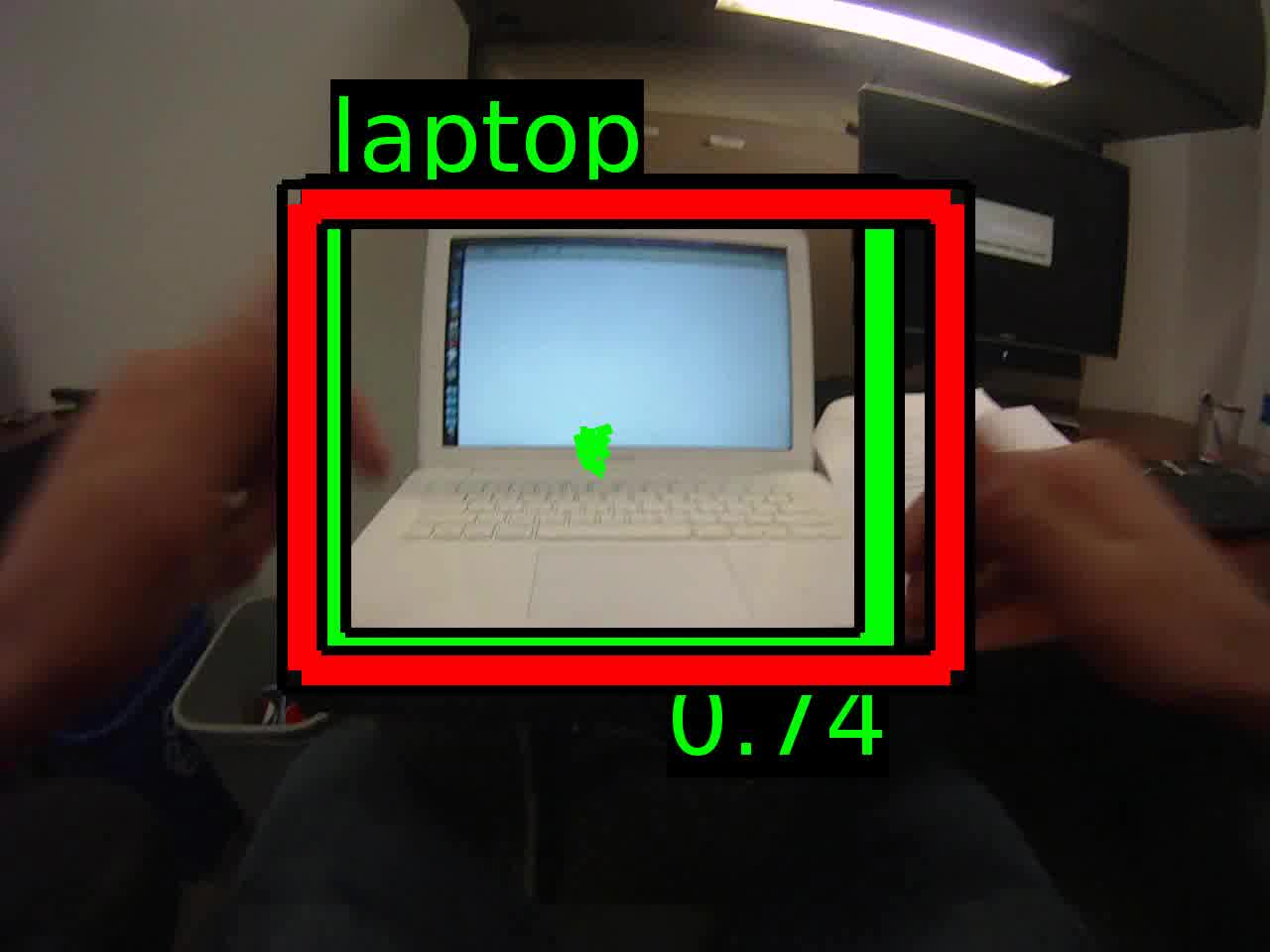}
	
	\centerline{(a) correct predictions}
	
	\vspace{2mm}
	
	\includegraphics[width=0.15\linewidth]{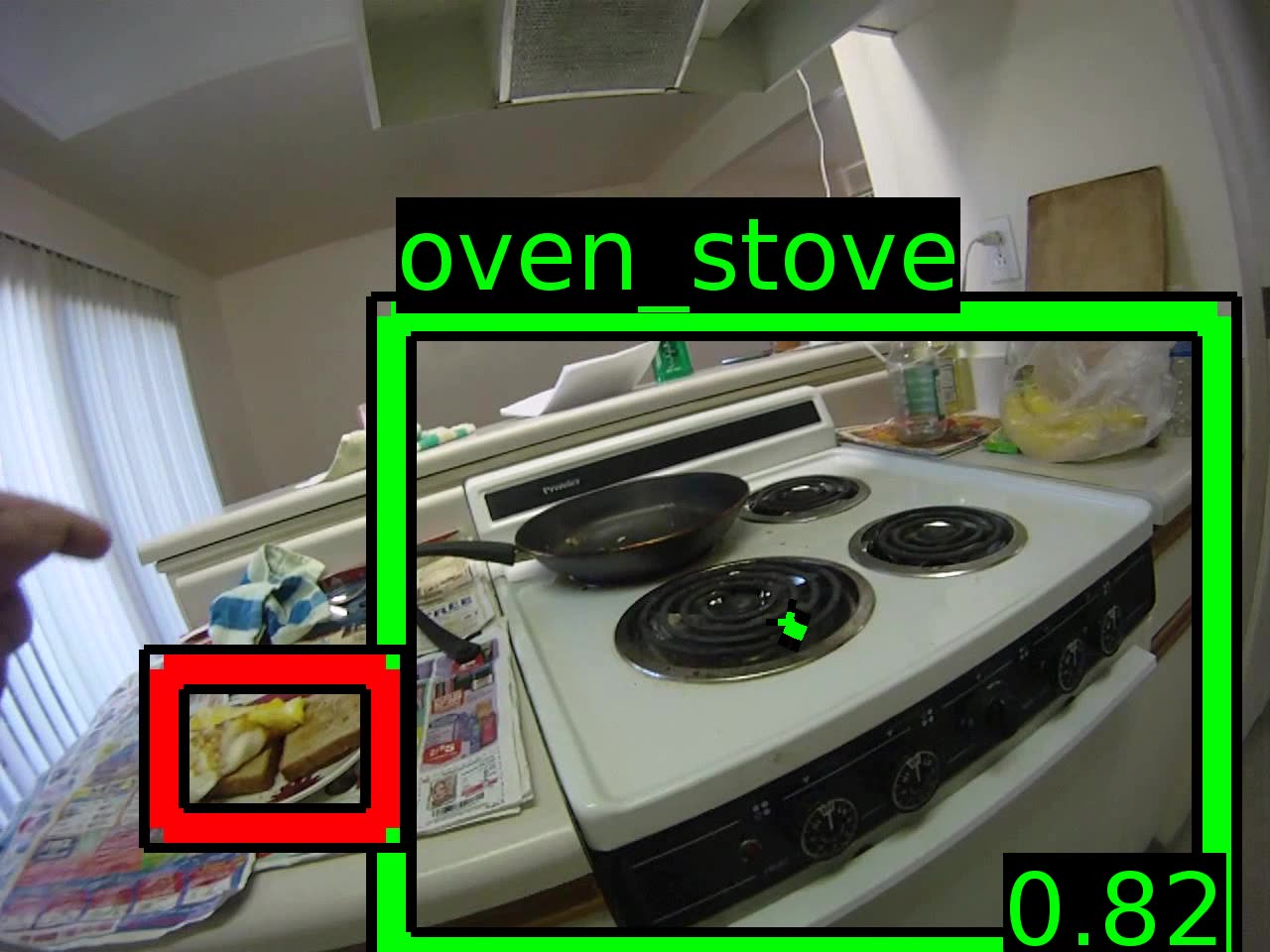} \hfill
	\includegraphics[width=0.15\linewidth]{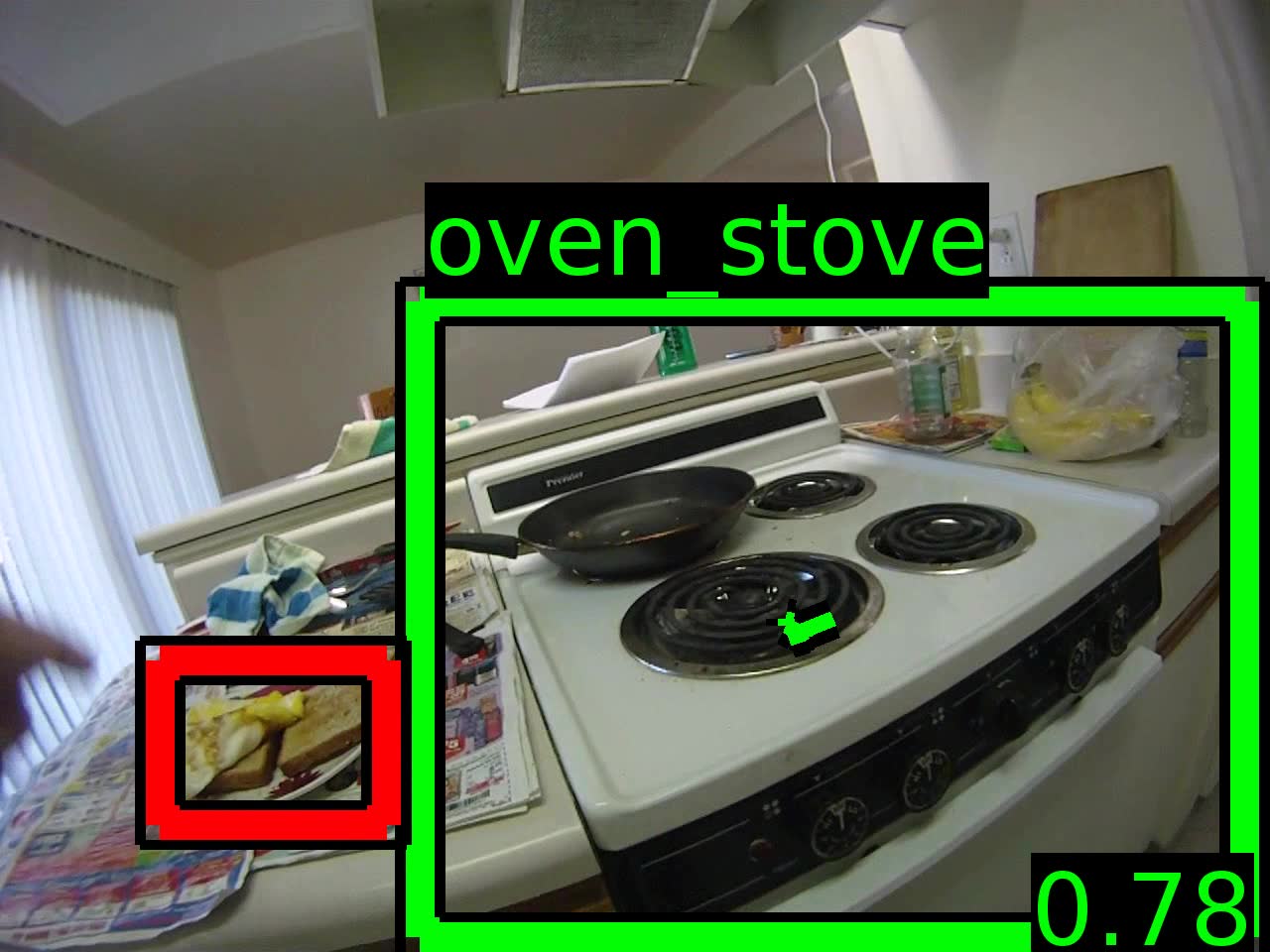} \hfill
	\includegraphics[width=0.15\linewidth]{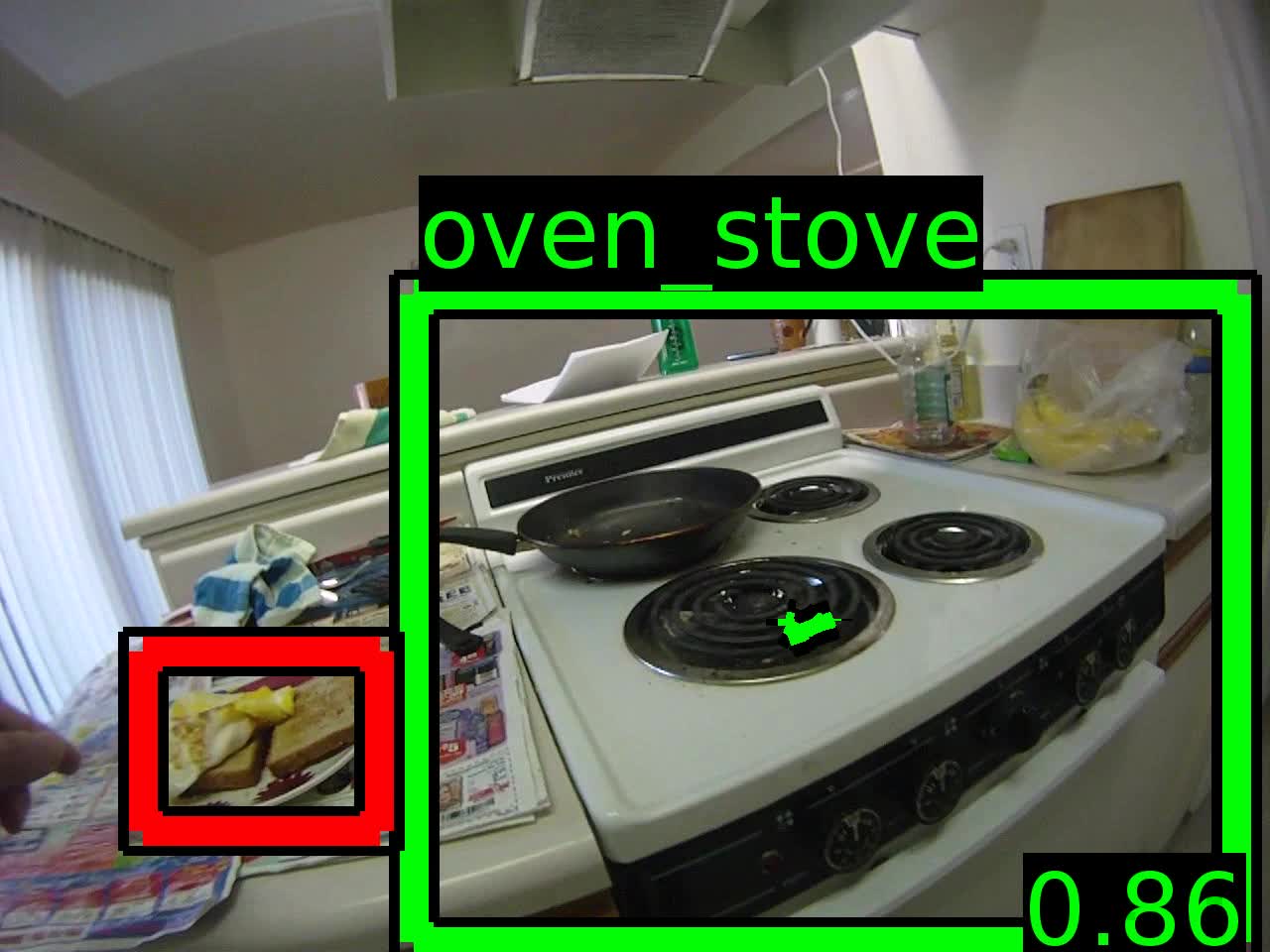} \hfill
	\includegraphics[width=0.15\linewidth]{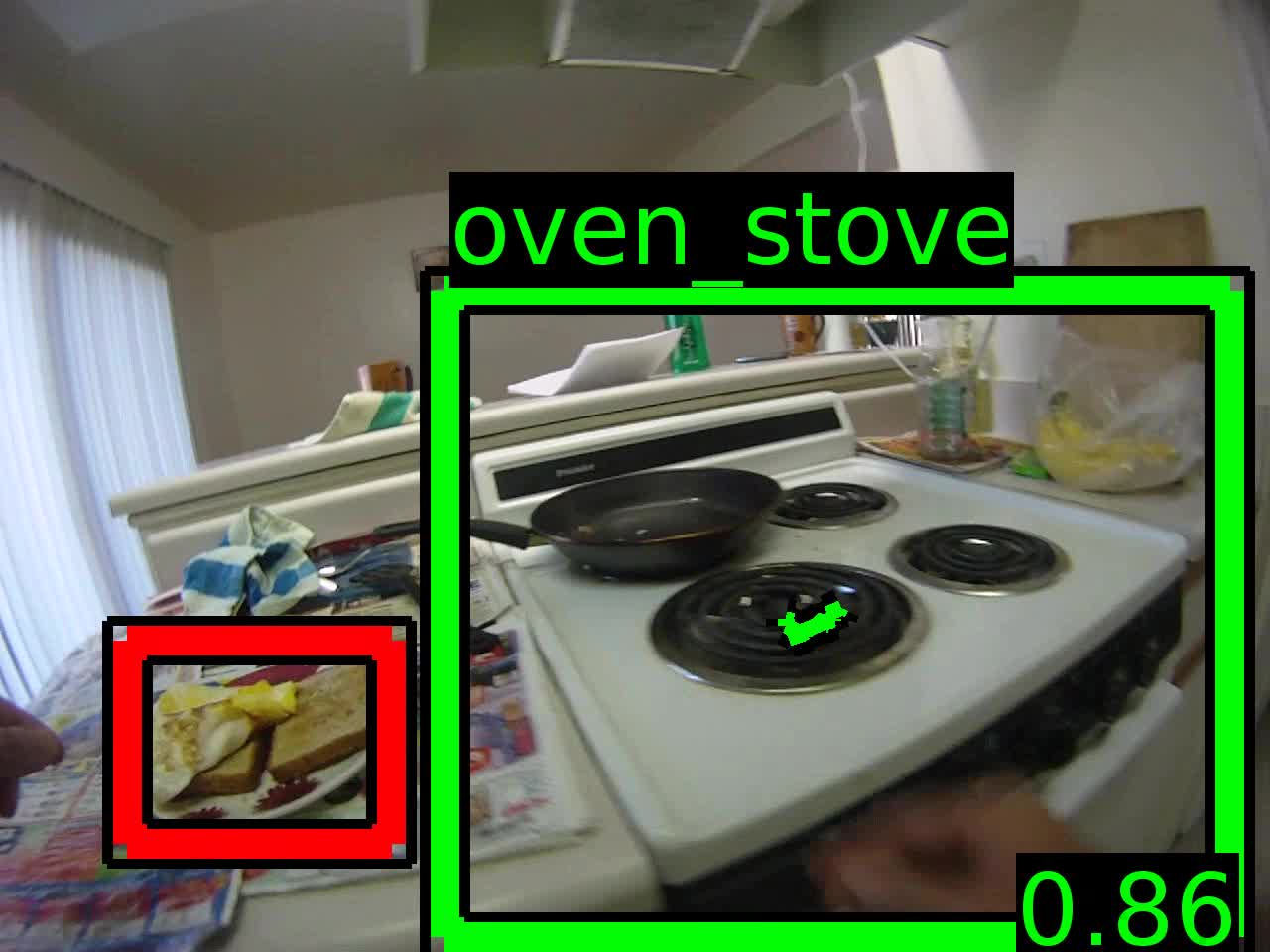} \hfill
	\includegraphics[width=0.15\linewidth]{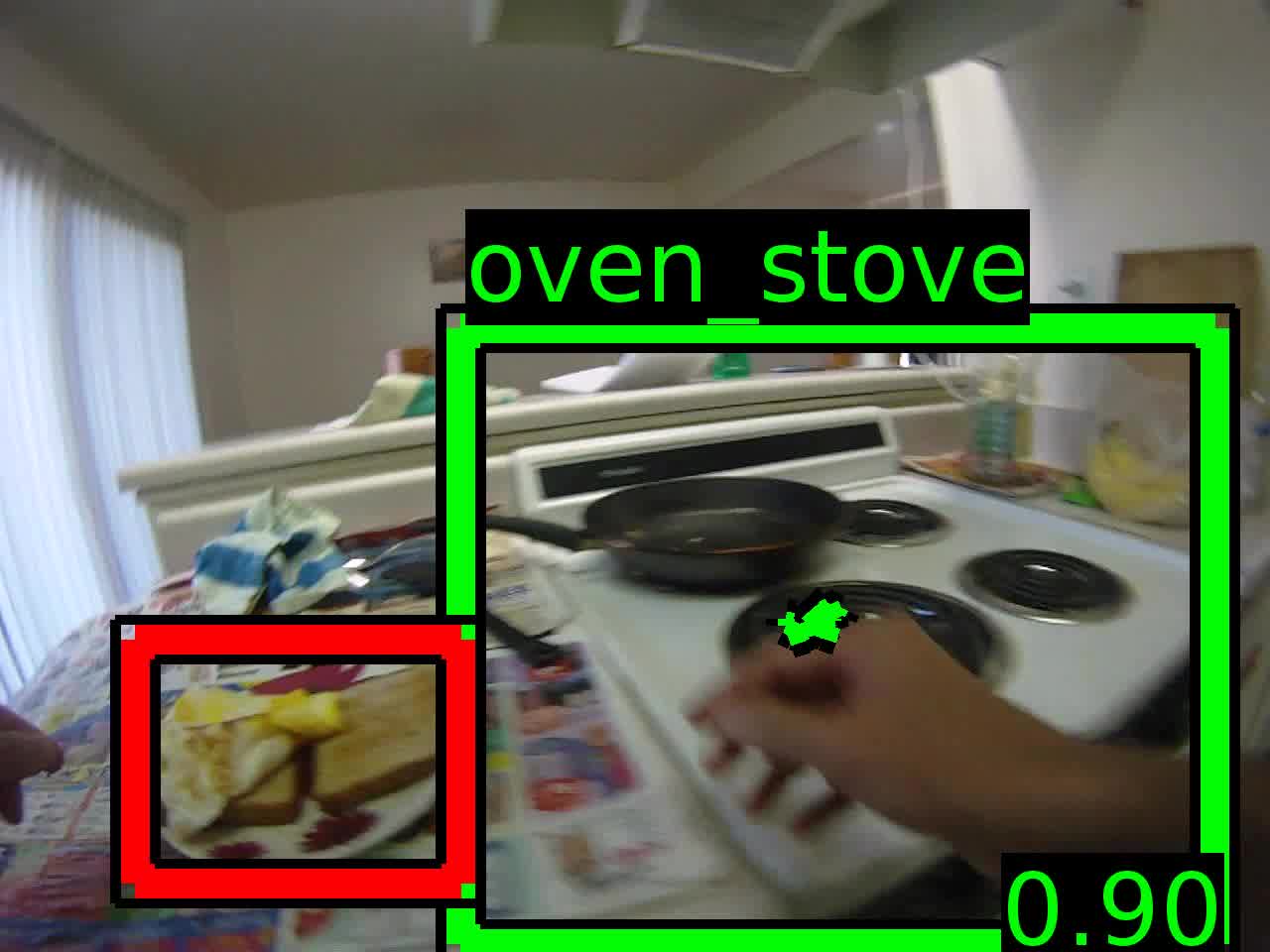} \hfill
	\includegraphics[width=0.15\linewidth]{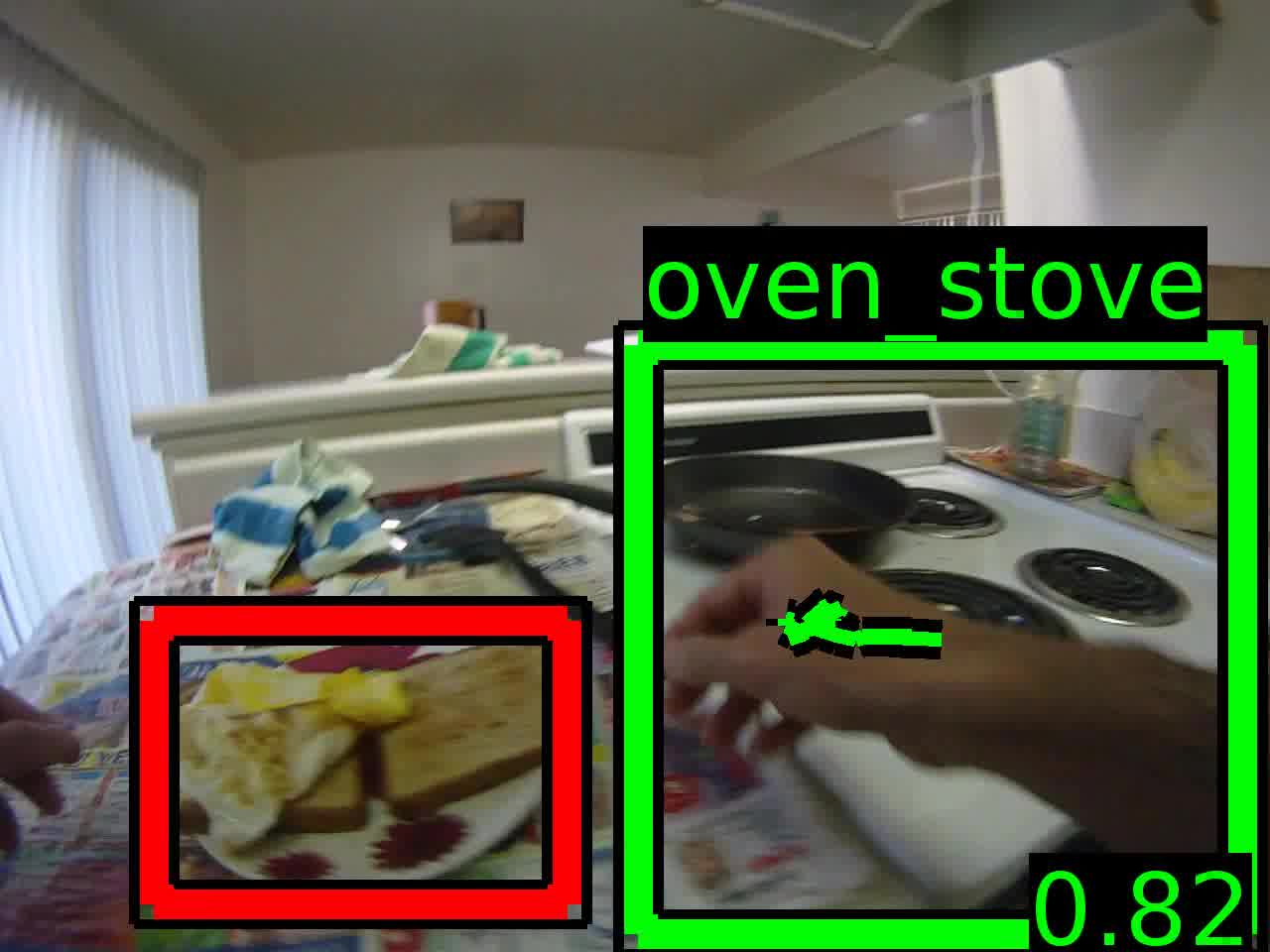}
	
	\vspace{1mm}
	
	\includegraphics[width=0.15\linewidth]{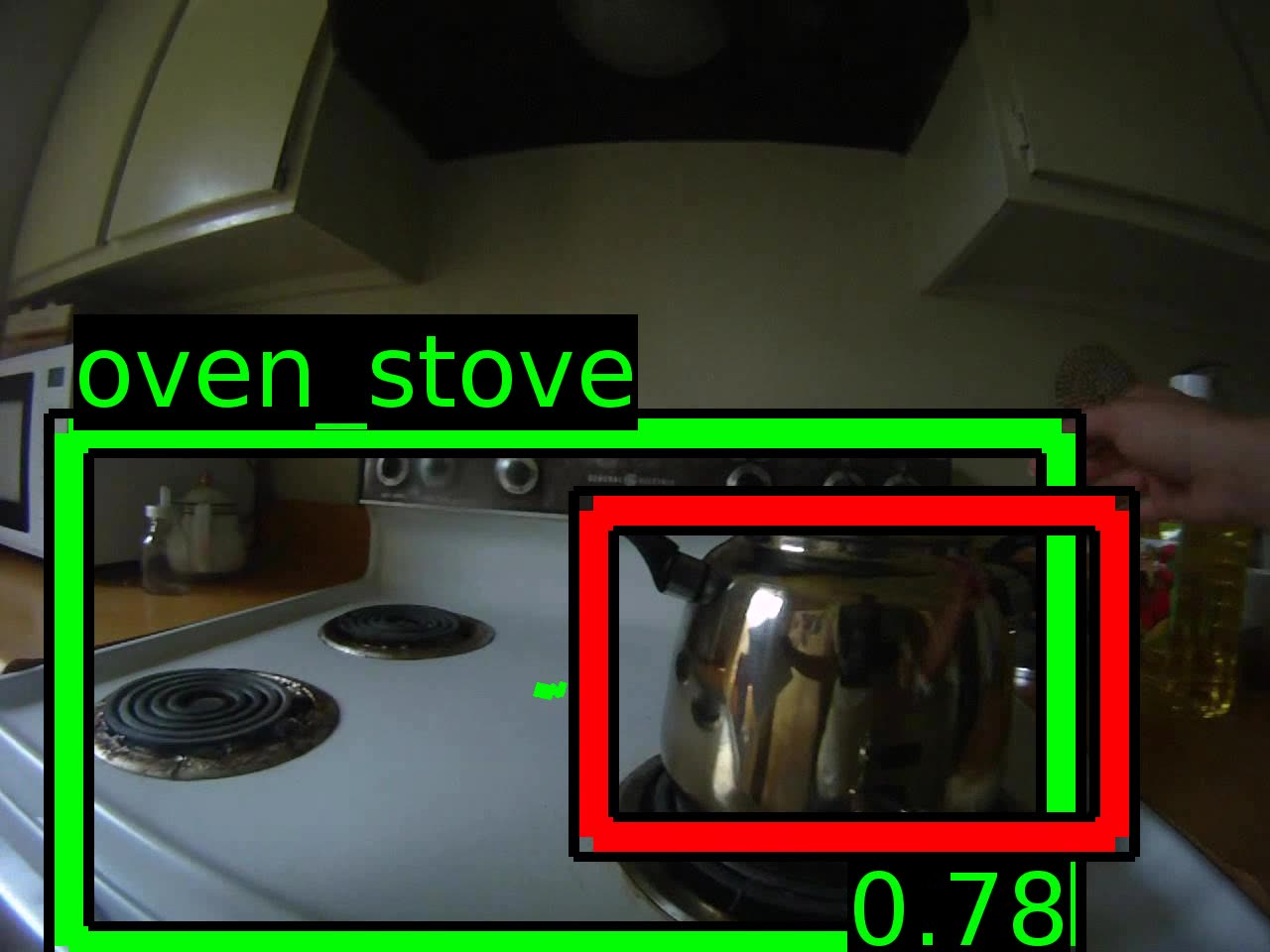} \hfill
	\includegraphics[width=0.15\linewidth]{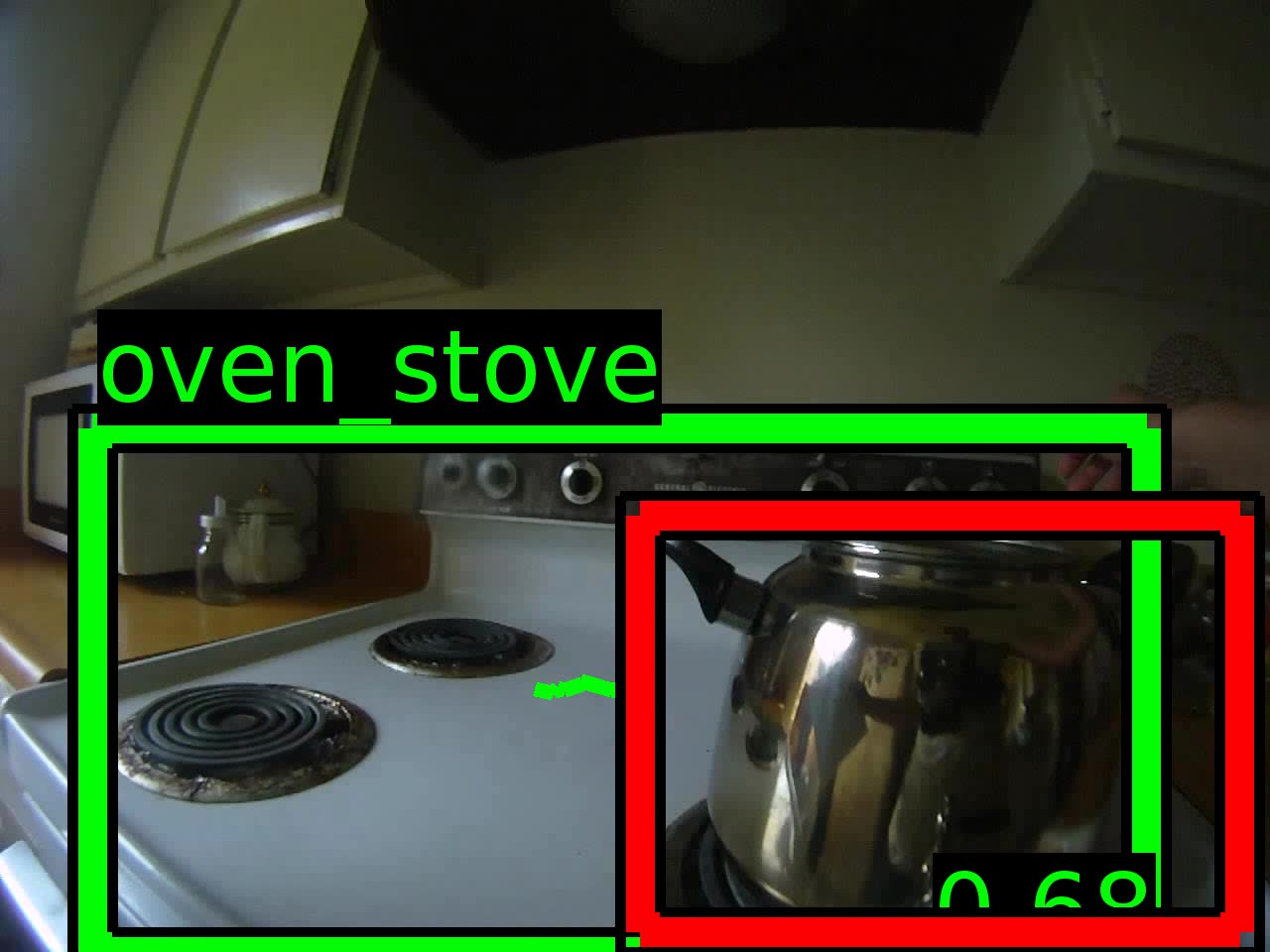} \hfill
	\includegraphics[width=0.15\linewidth]{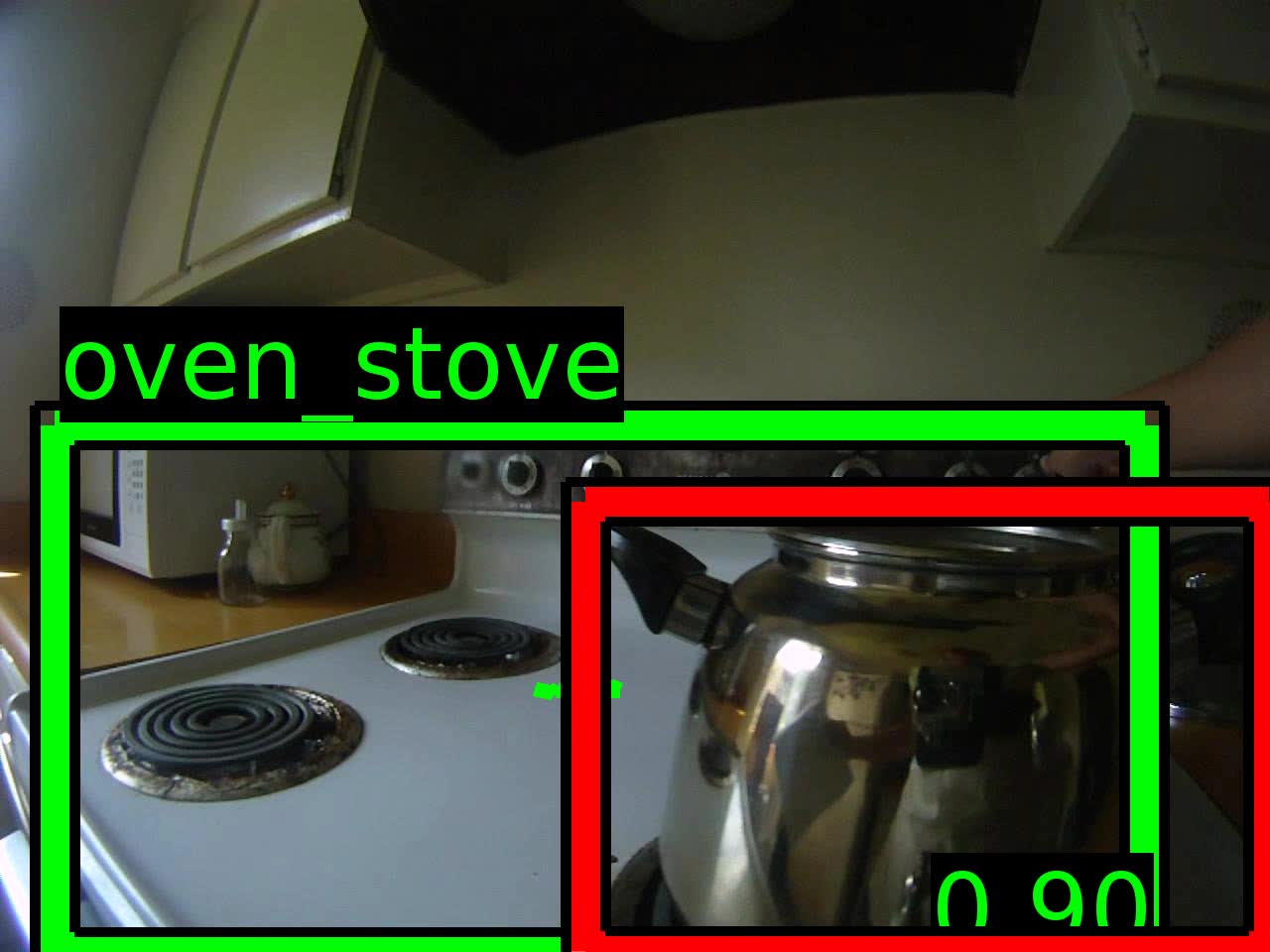} \hfill
	\includegraphics[width=0.15\linewidth]{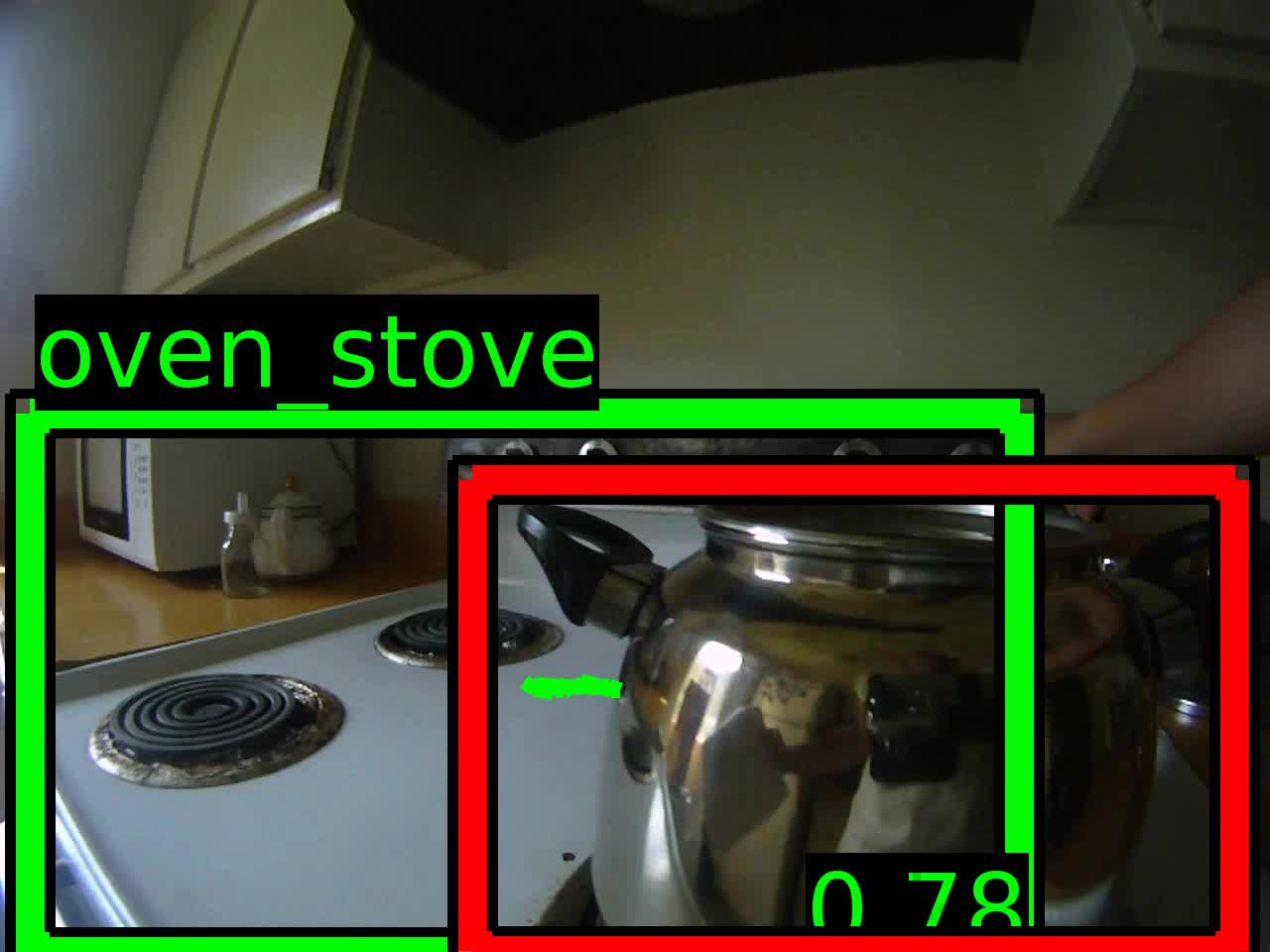} \hfill
	\includegraphics[width=0.15\linewidth]{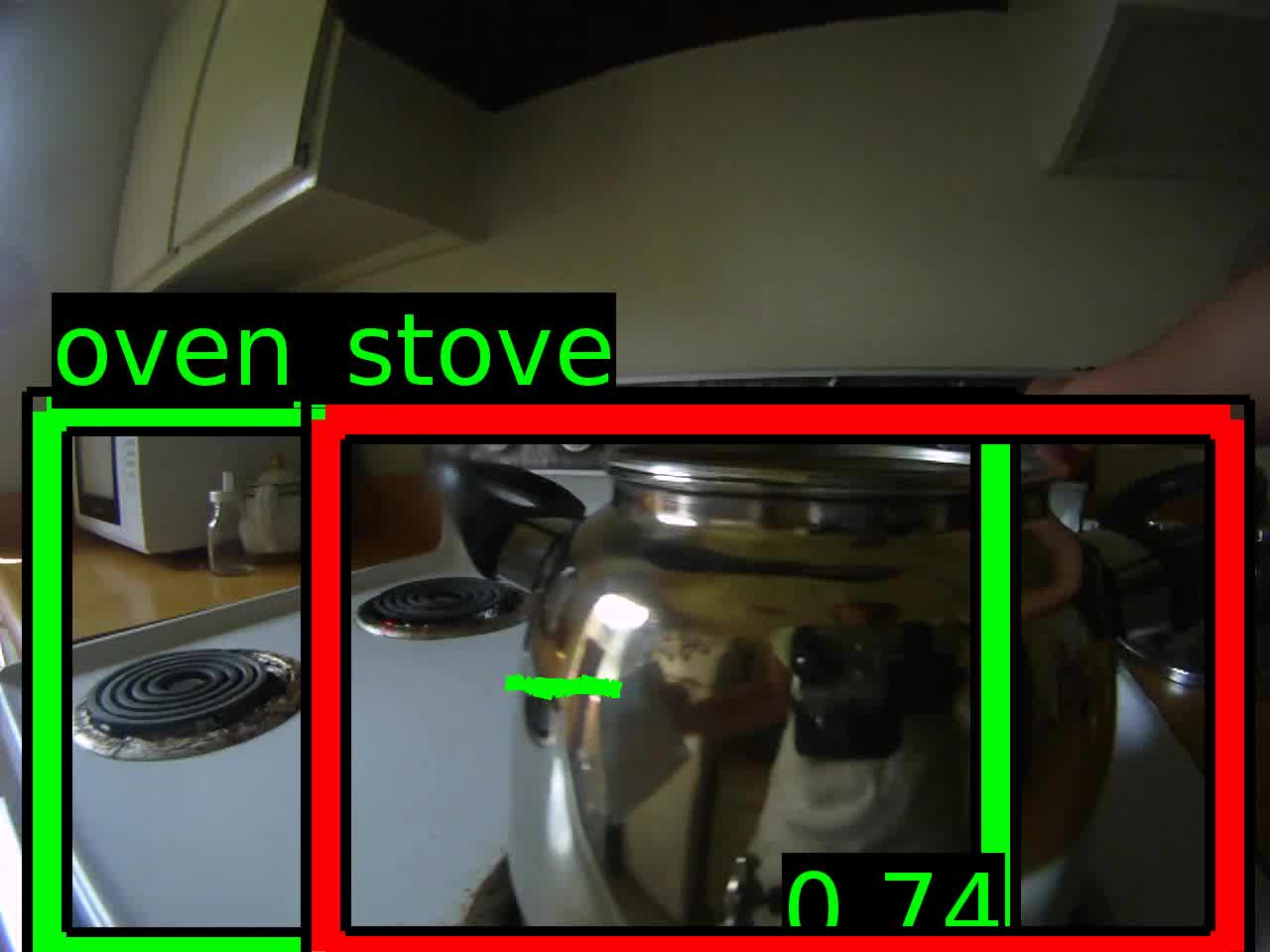} \hfill
	\includegraphics[width=0.15\linewidth]{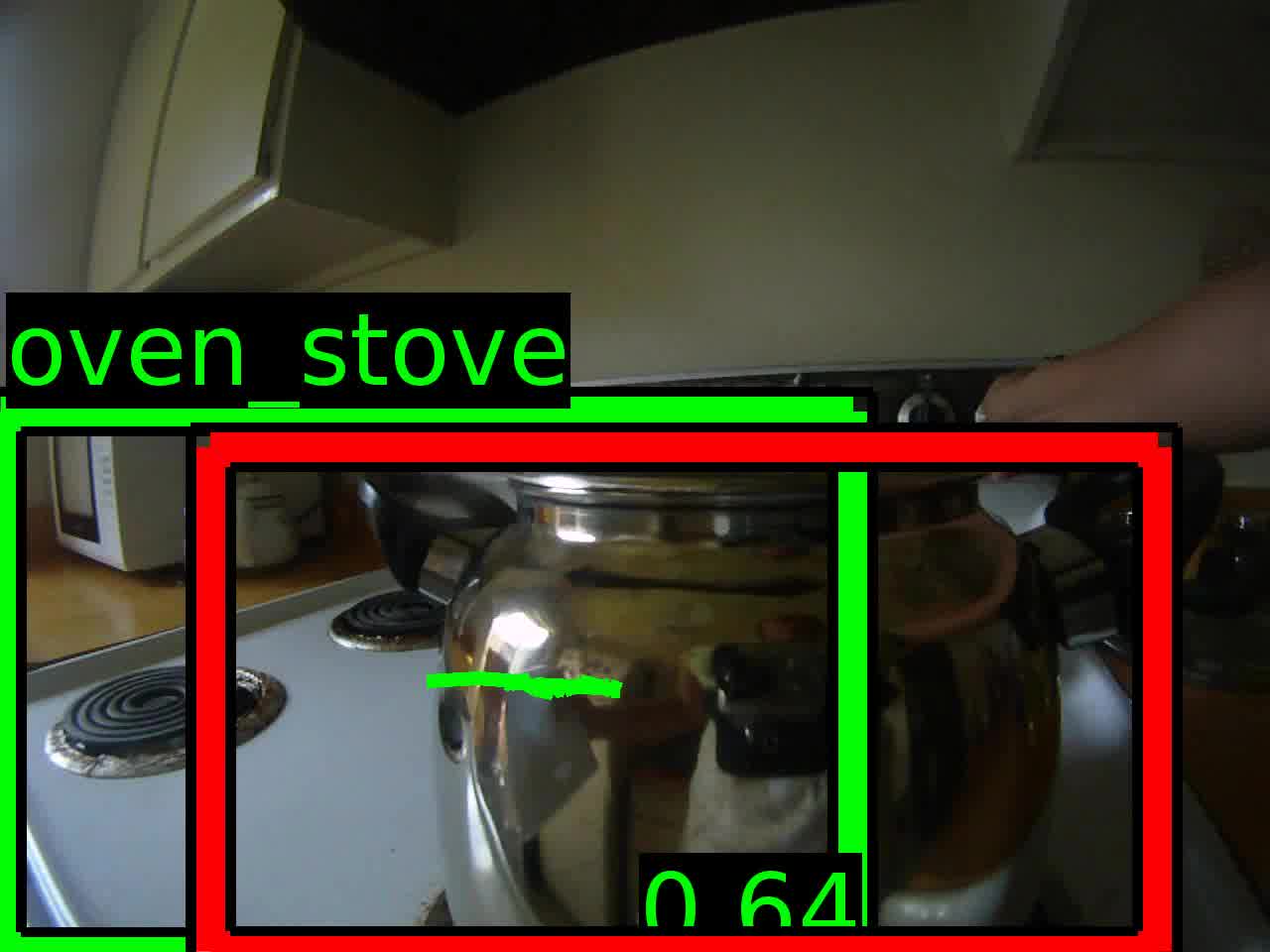}
	
	\vspace{1mm}
	\includegraphics[width=0.15\linewidth]{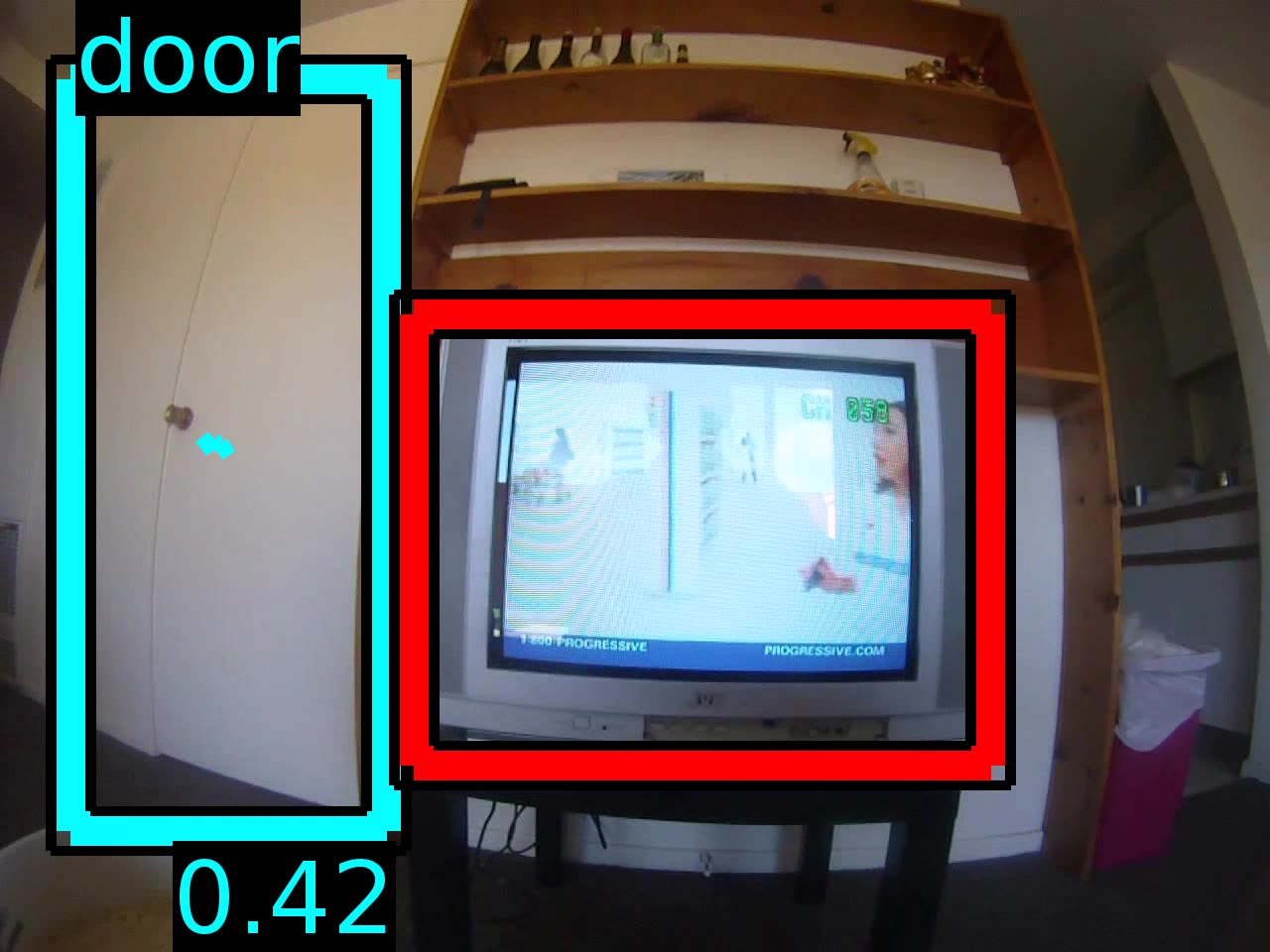} \hfill
	\includegraphics[width=0.15\linewidth]{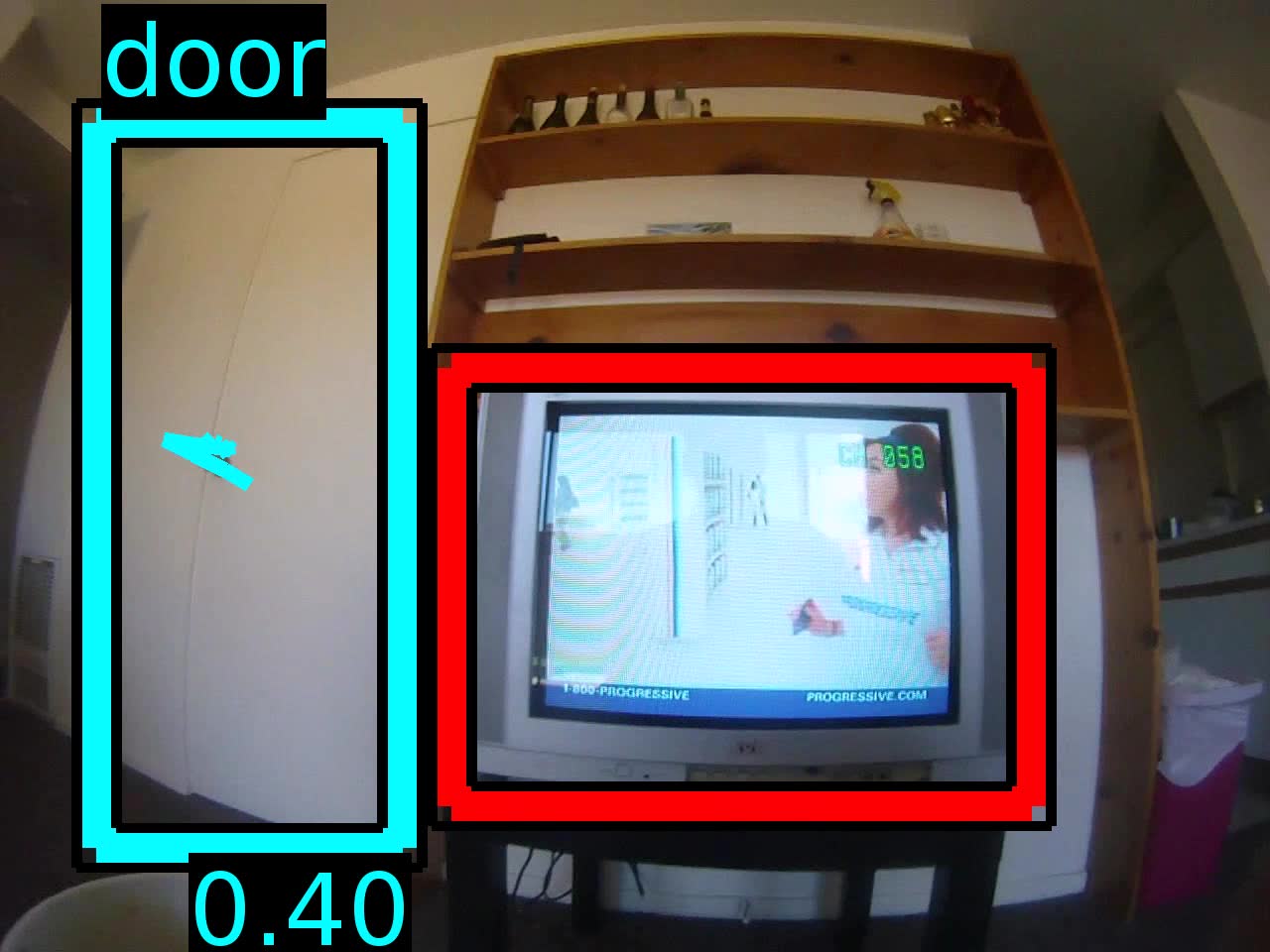} \hfill
	\includegraphics[width=0.15\linewidth]{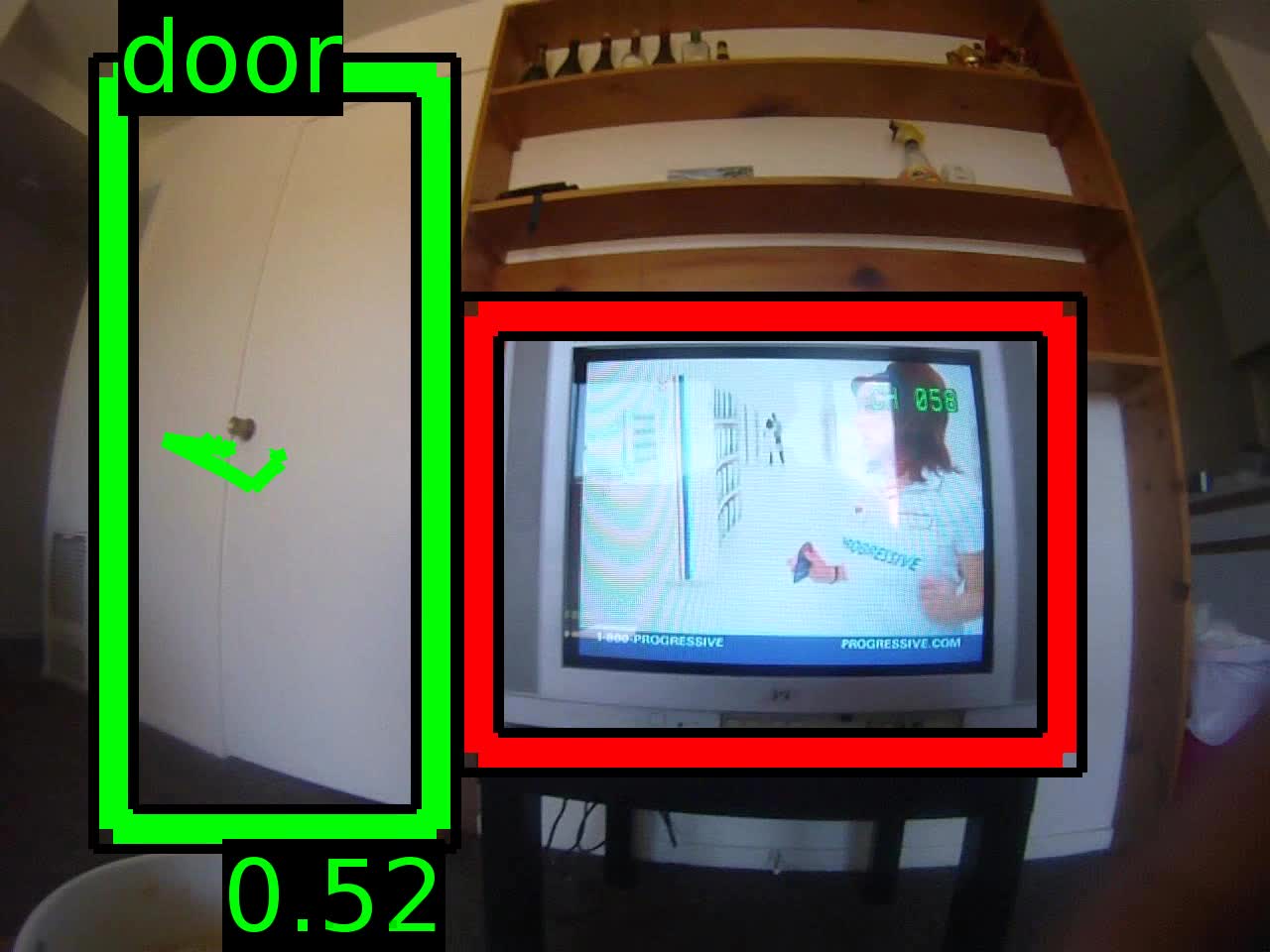} \hfill
	\includegraphics[width=0.15\linewidth]{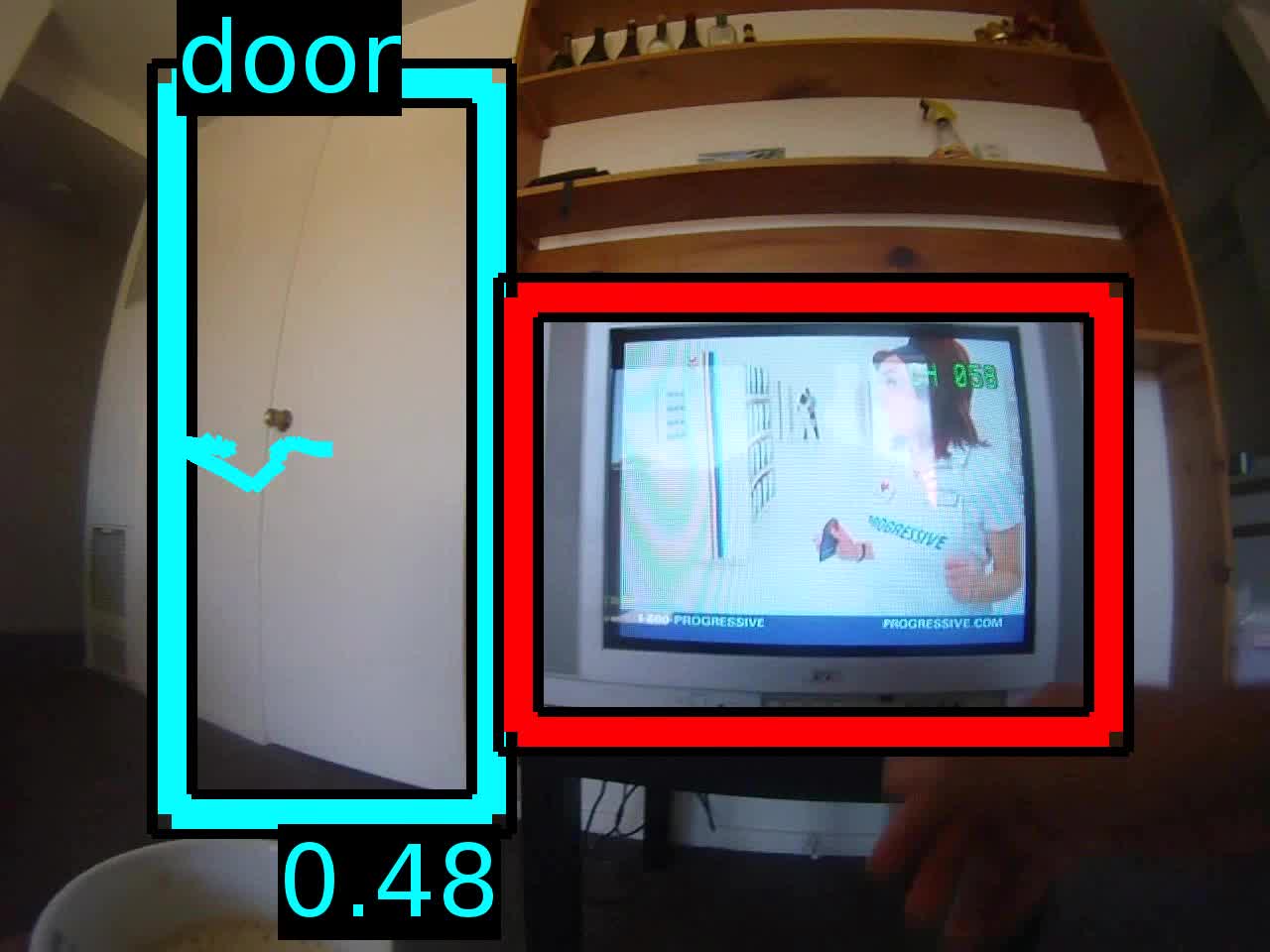} \hfill
	\includegraphics[width=0.15\linewidth]{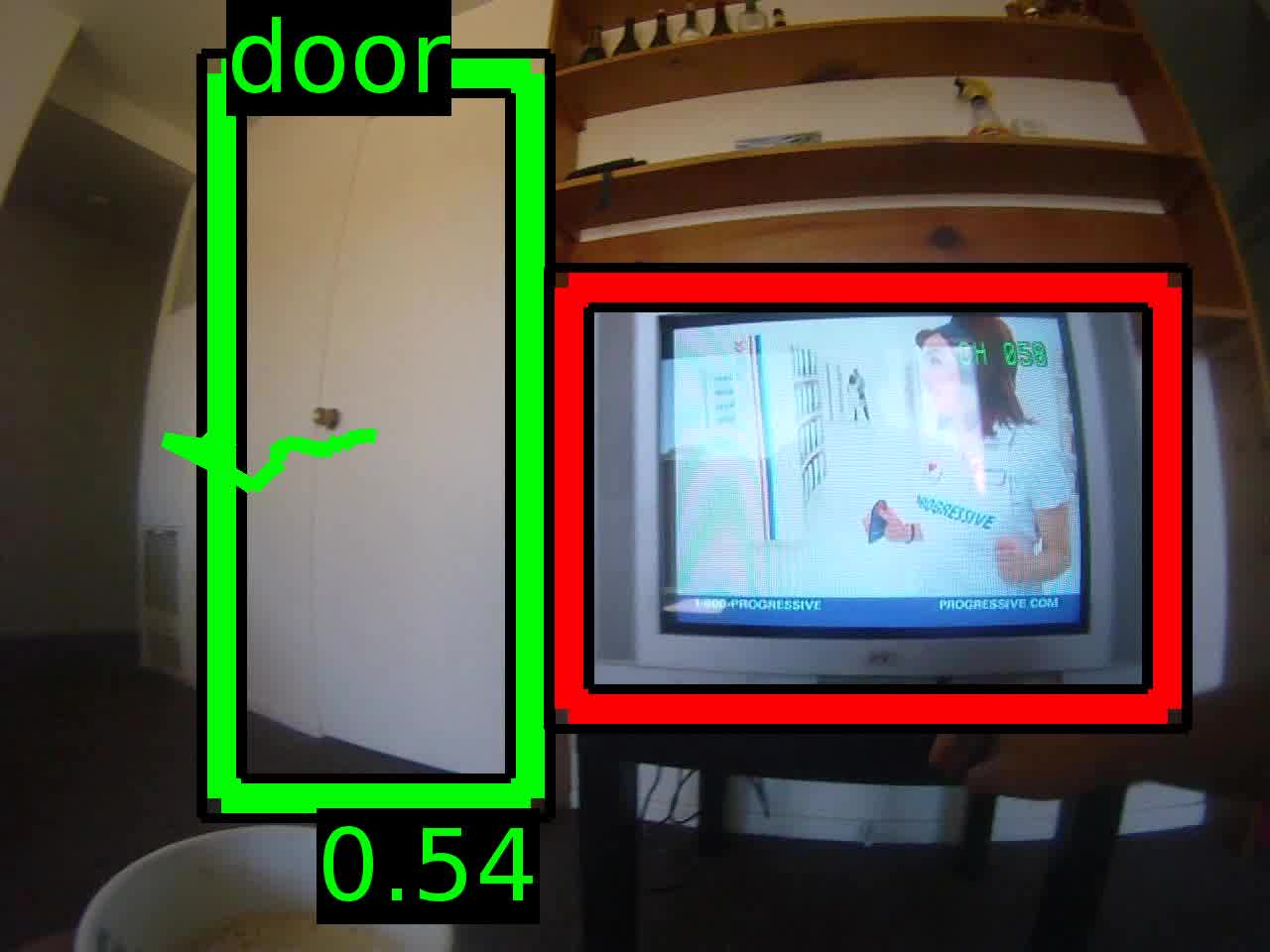} \hfill
	\includegraphics[width=0.15\linewidth]{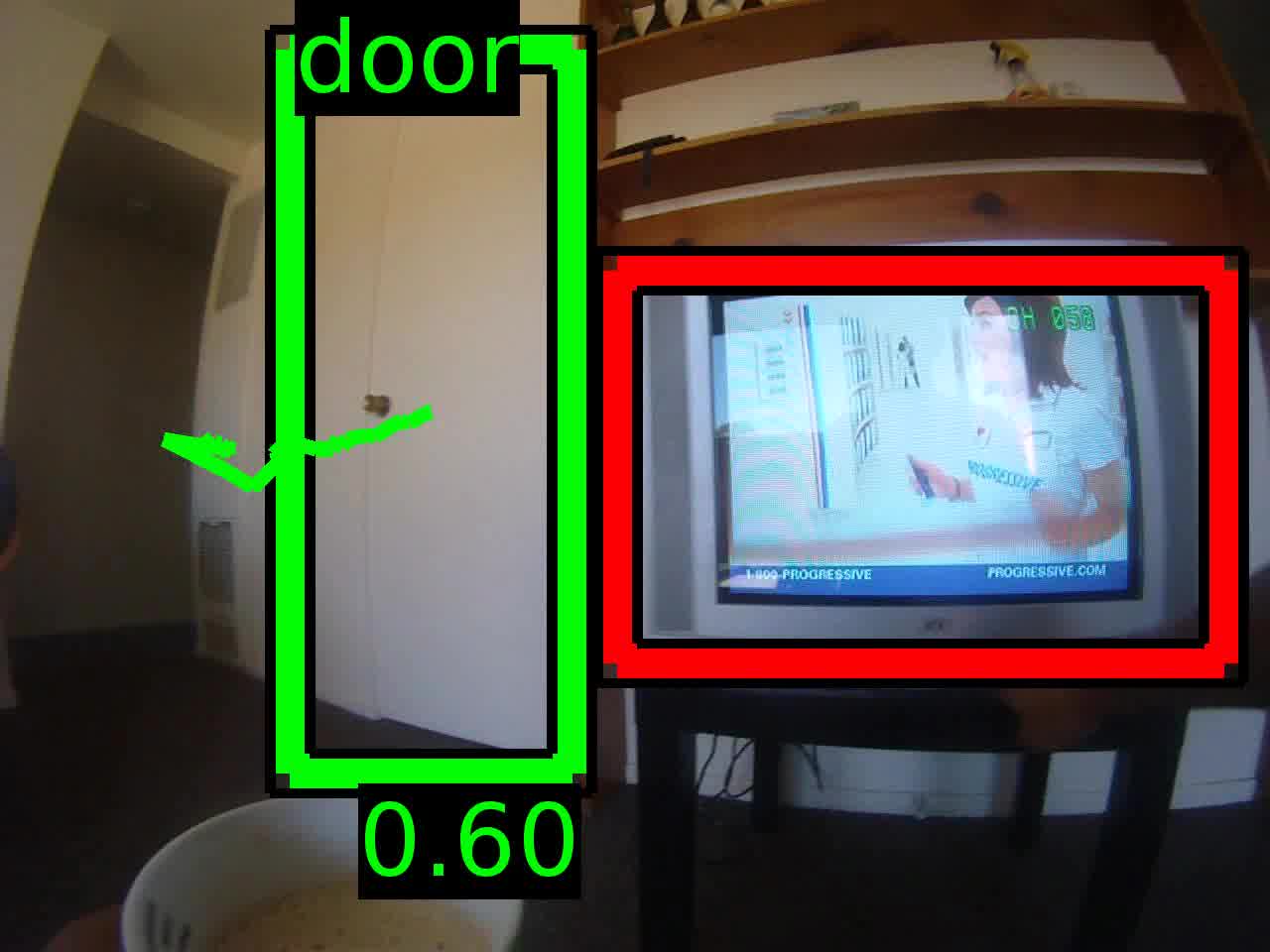}
	
	\vspace{1mm}
	\includegraphics[width=0.15\linewidth]{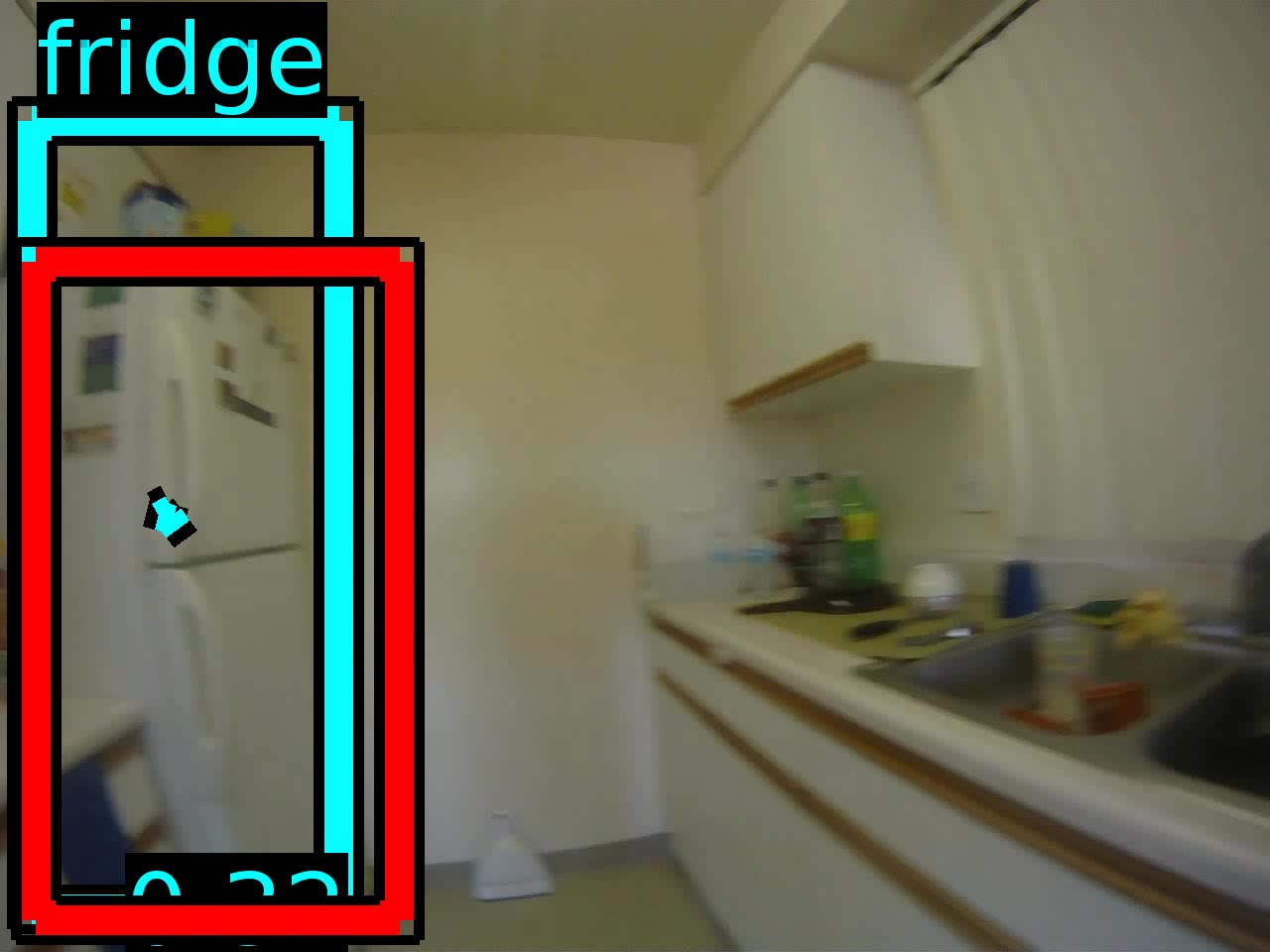} \hfill
	\includegraphics[width=0.15\linewidth]{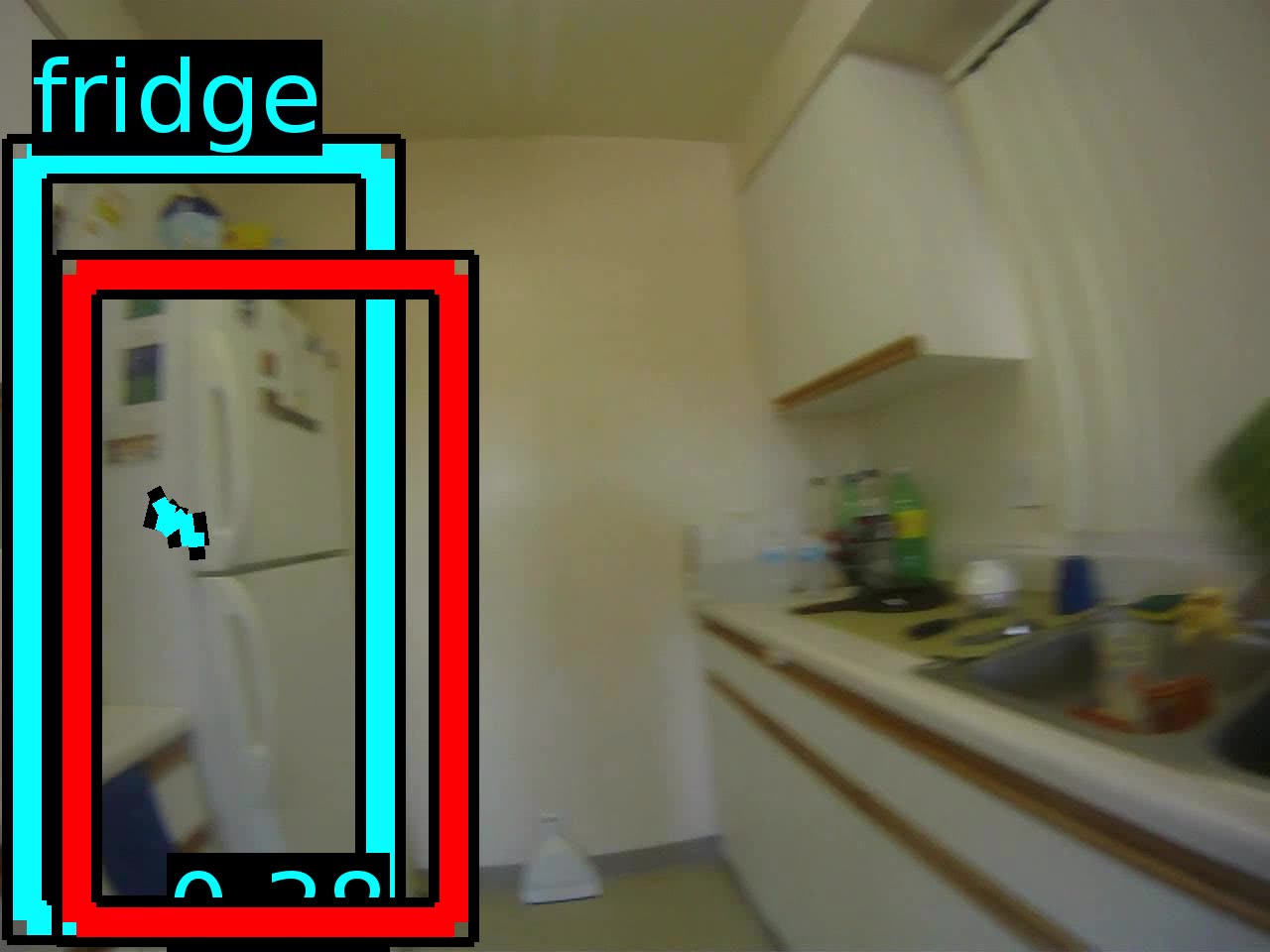} \hfill
	\includegraphics[width=0.15\linewidth]{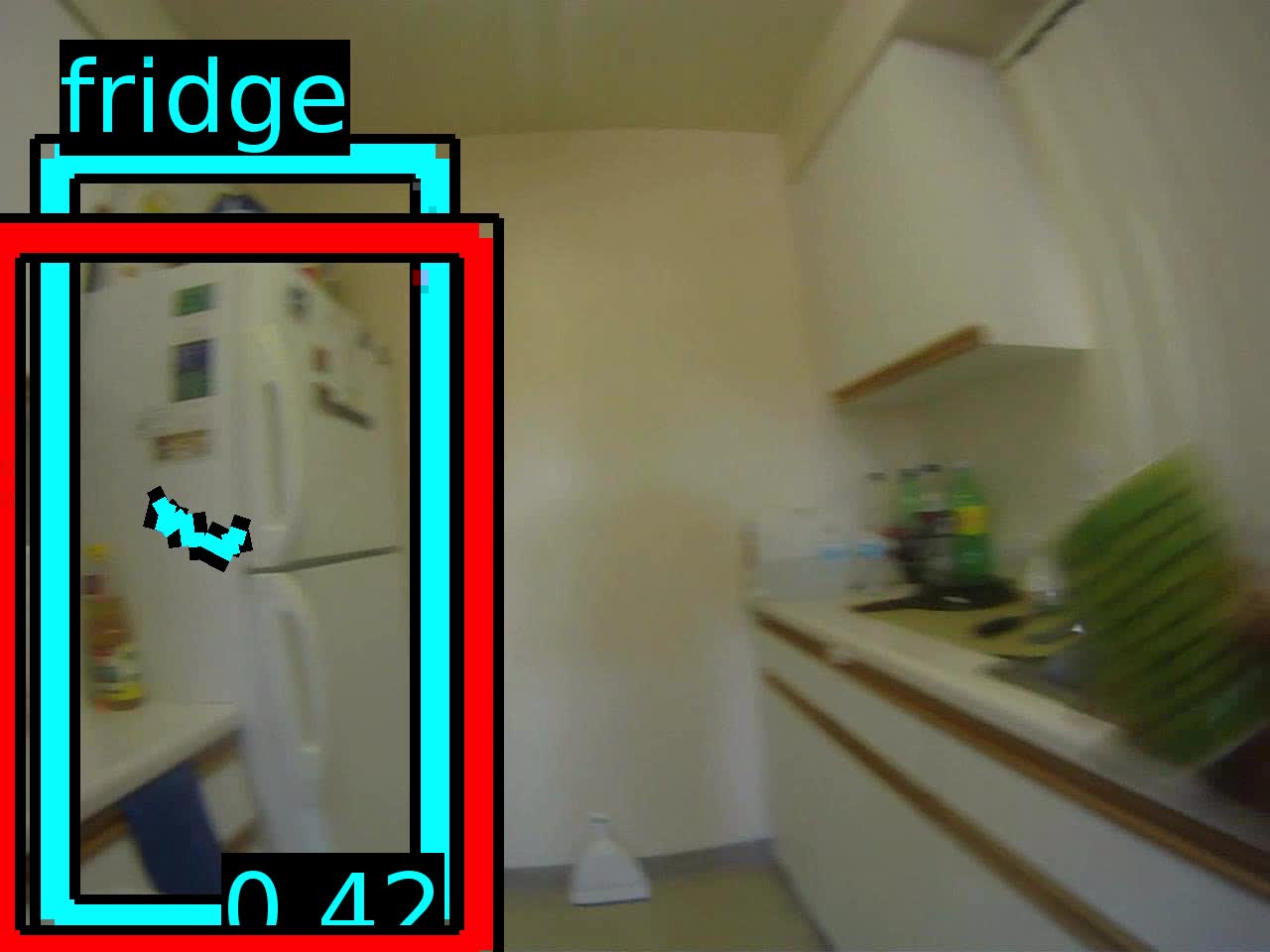} \hfill
	\includegraphics[width=0.15\linewidth]{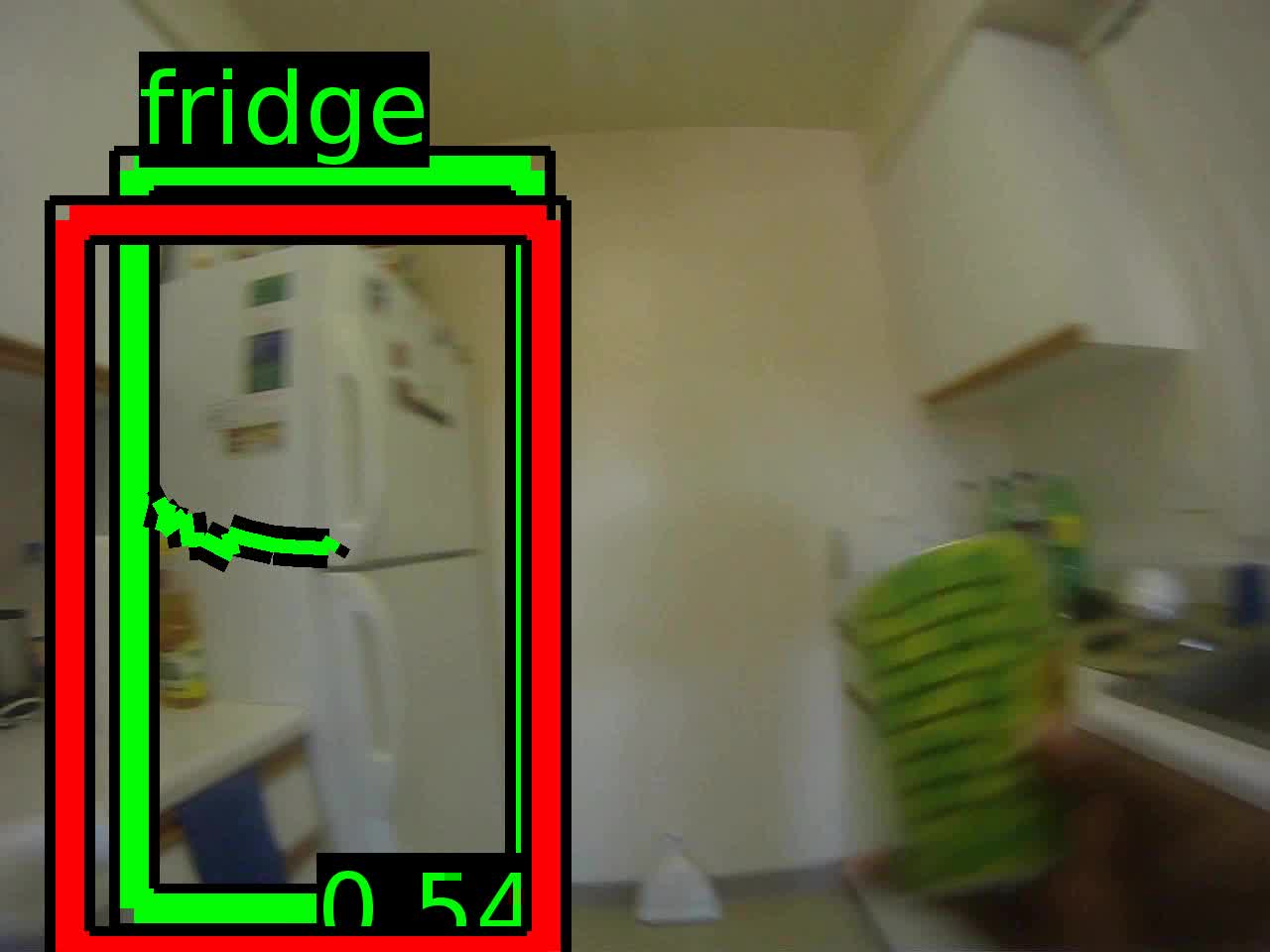} \hfill
	\includegraphics[width=0.15\linewidth]{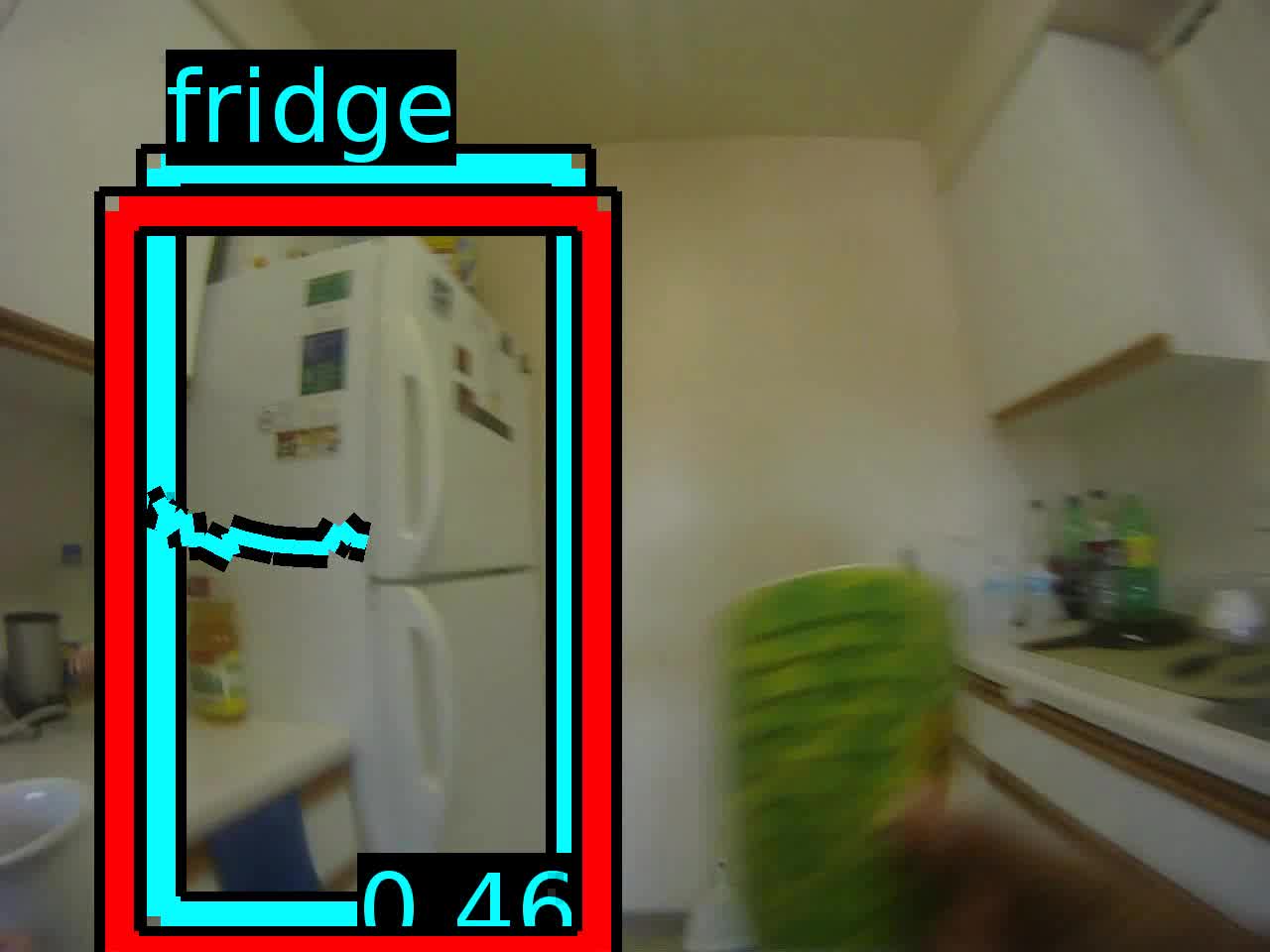} \hfill
	\includegraphics[width=0.15\linewidth]{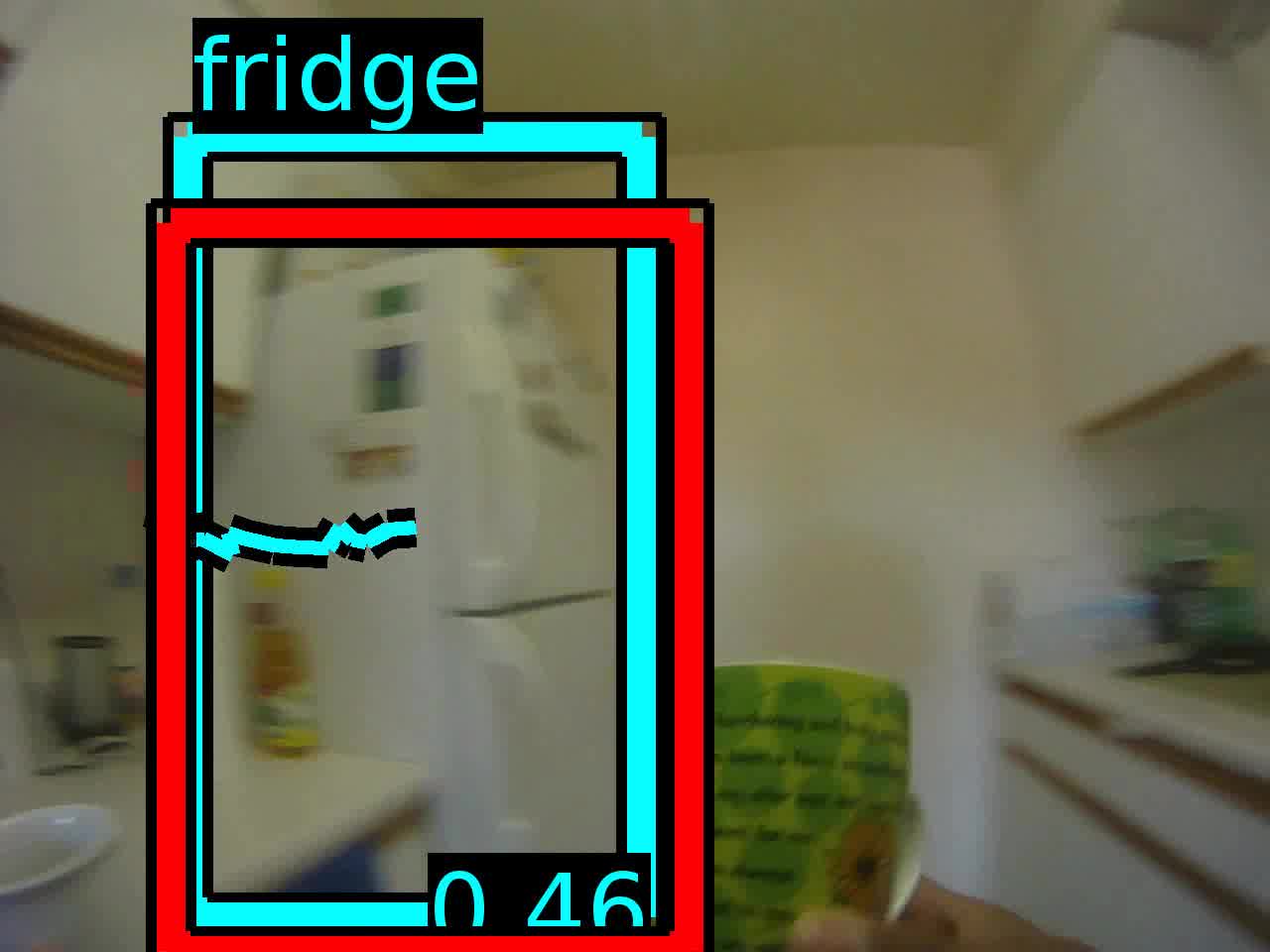}
	
	\vspace{1mm}
	\includegraphics[width=0.15\linewidth]{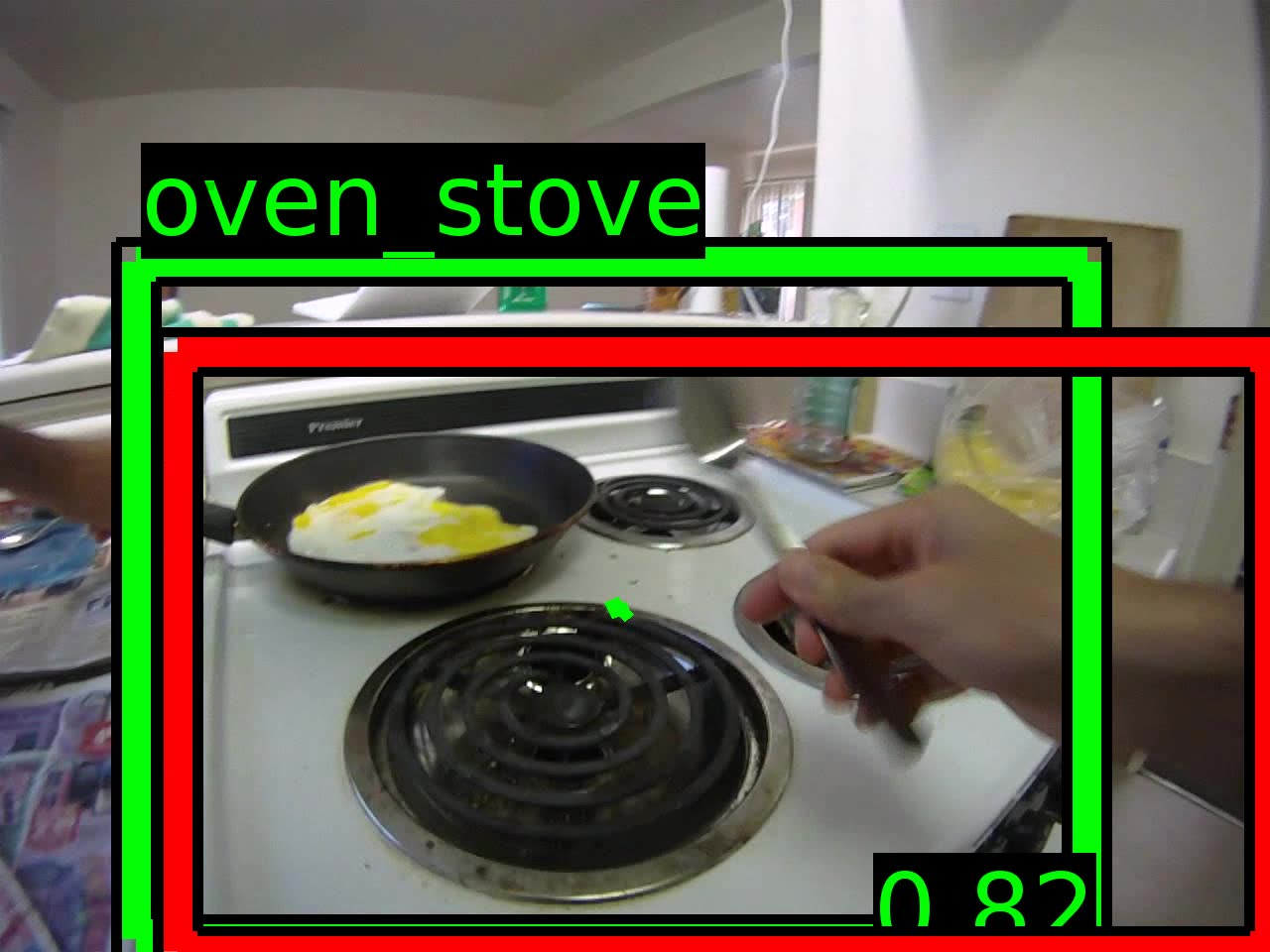} \hfill
	\includegraphics[width=0.15\linewidth]{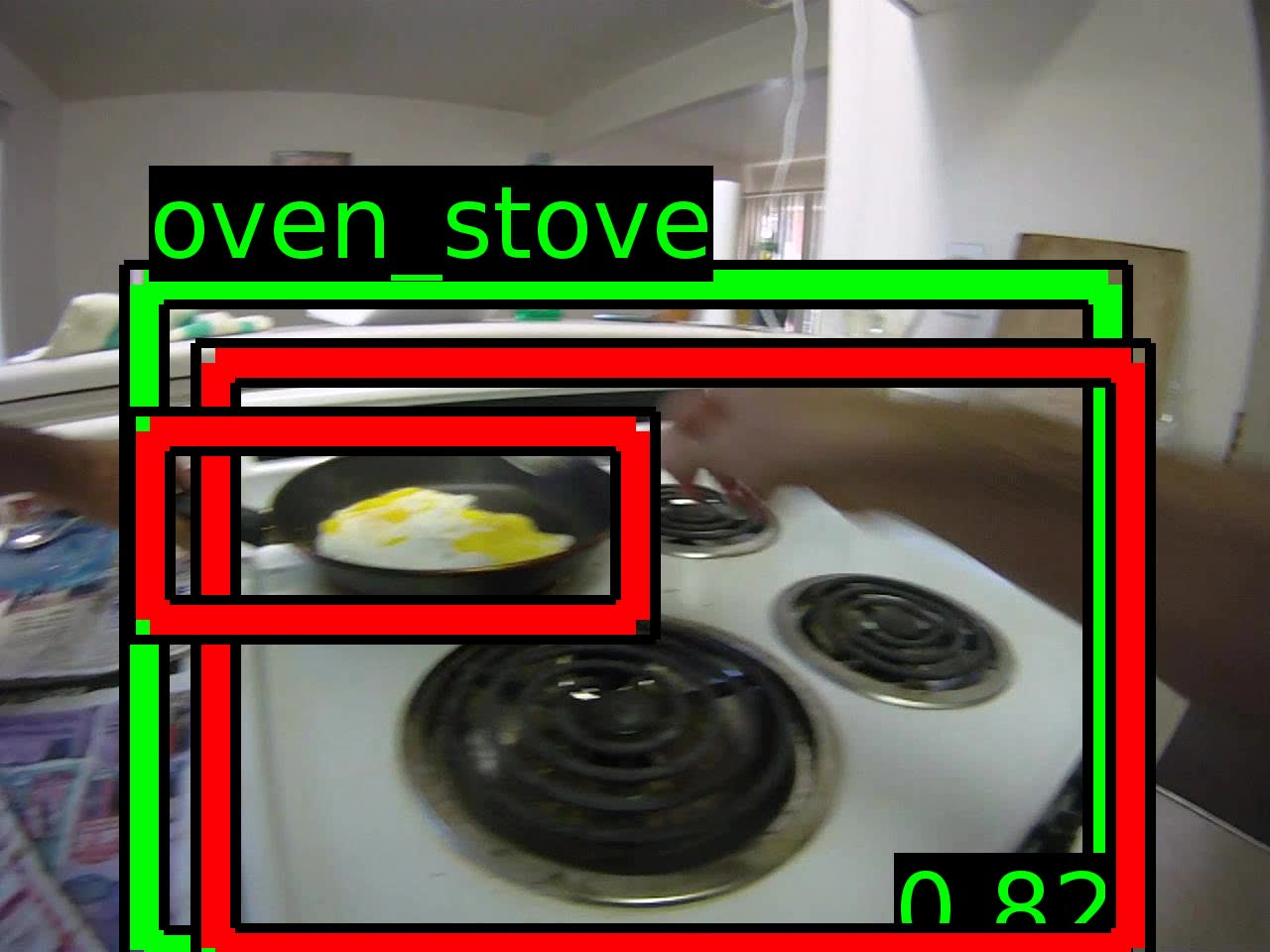} \hfill
	\includegraphics[width=0.15\linewidth]{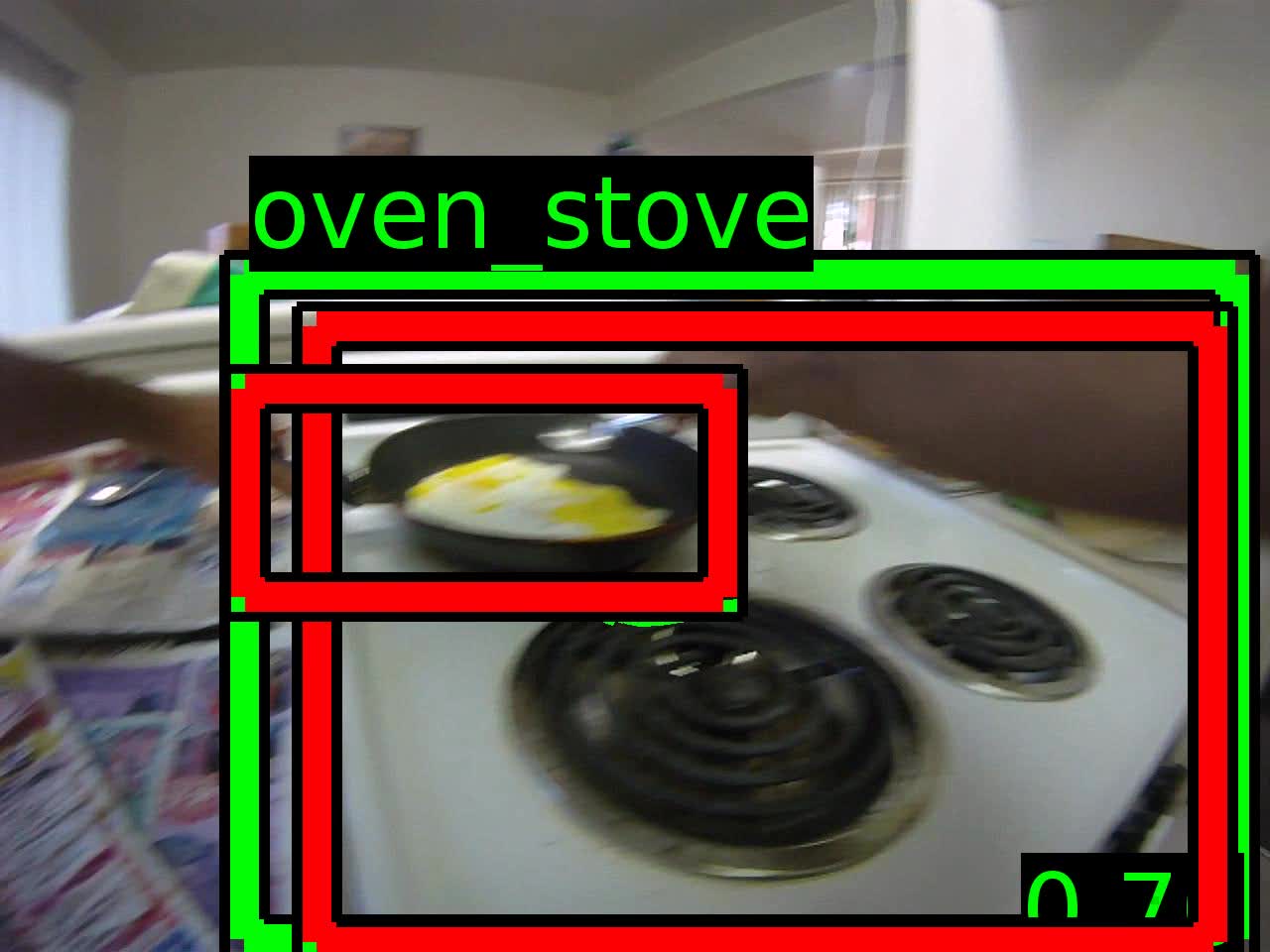} \hfill
	\includegraphics[width=0.15\linewidth]{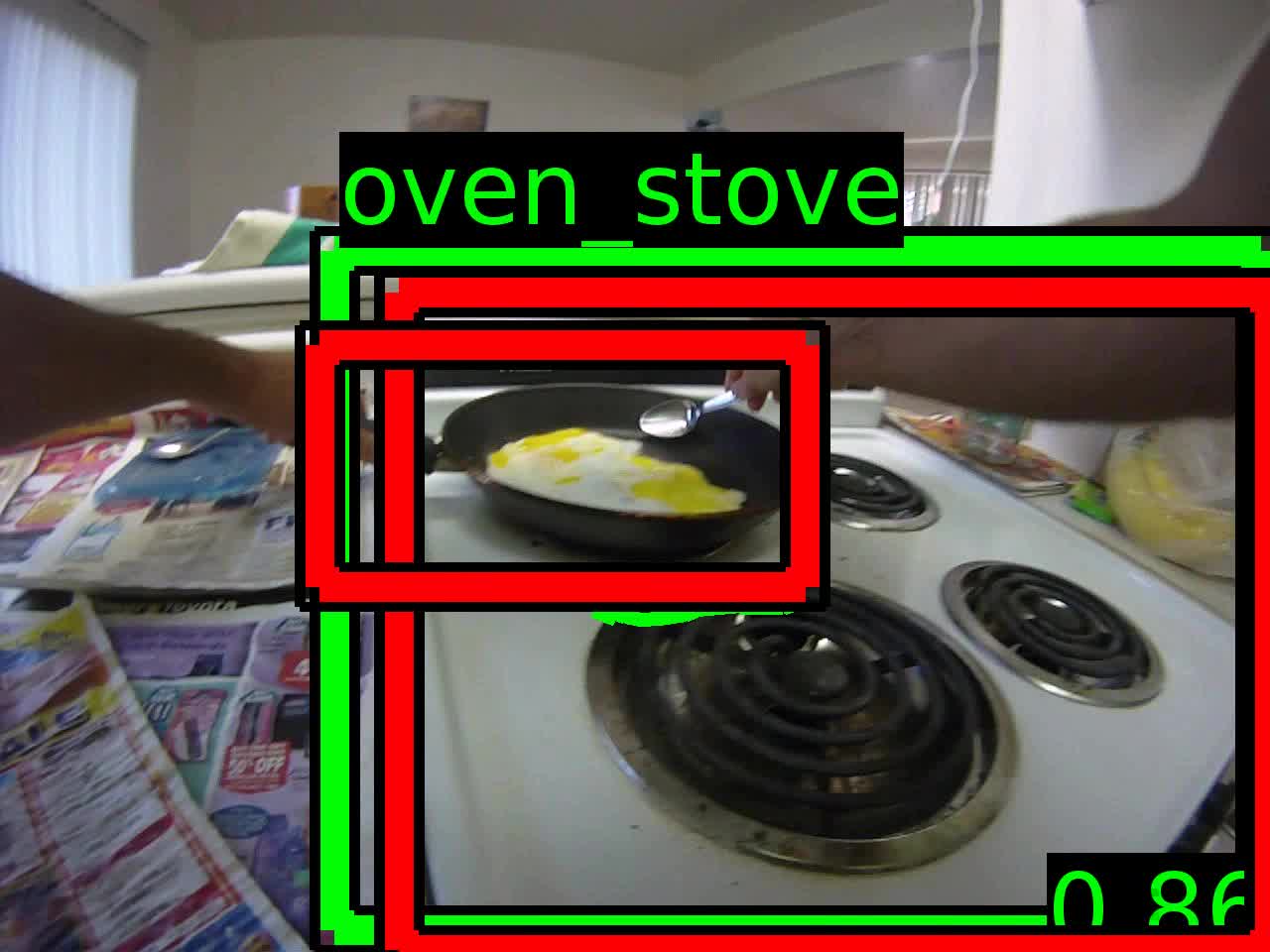} \hfill
	\includegraphics[width=0.15\linewidth]{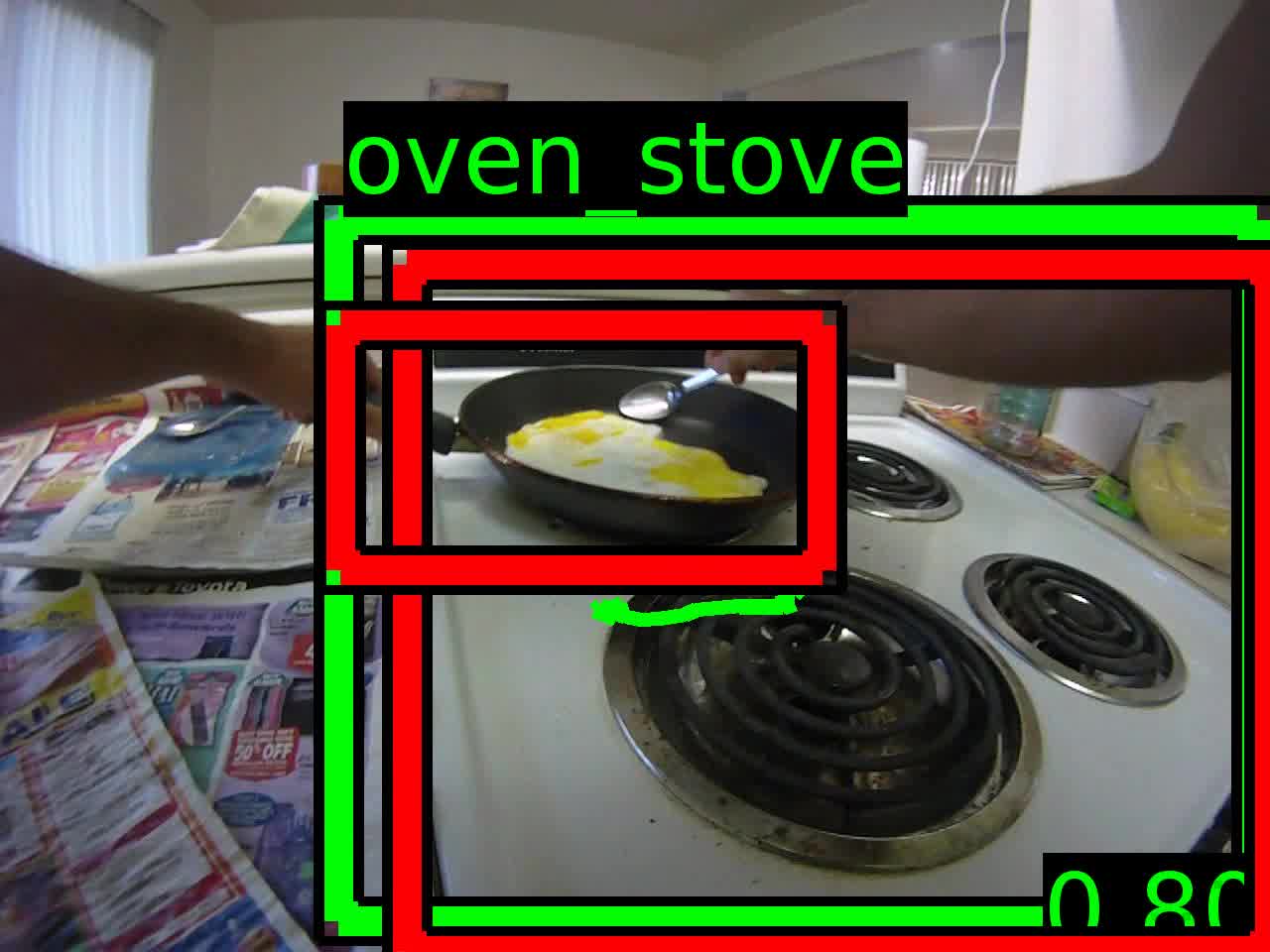} \hfill
	\includegraphics[width=0.15\linewidth]{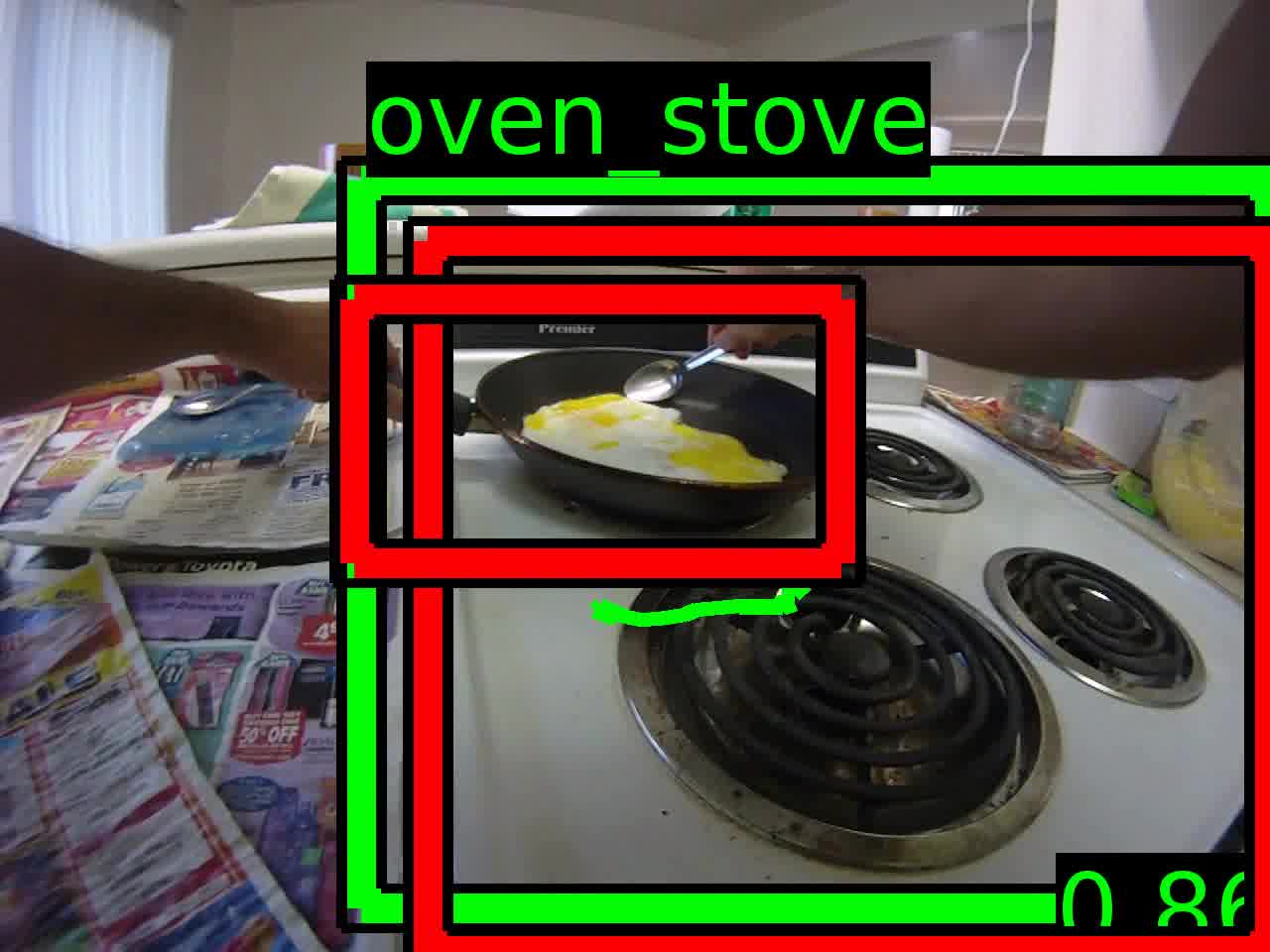}
	
	\centerline{(b) failure cases}
	
	\caption{Some success/failure examples of the proposed method. Red bounding boxes represent ground truth next-active-objects. Positive model predictions are indicated in green, negative ones in cyan. Confidence scores are reported for each prediction.}
	\label{fig:prediction_samples}
	
\end{figure*}

\figurename~\ref{fig:prediction_samples} reports some visual examples of success/failure sequences related to the proposed method. 
In the examples of correct predictions shown in \figurename~\ref{fig:prediction_samples}(a), the model correctly assigns a high score (positive prediction) to next-active-objects {(e.g., the laptop in the first row and the tap in the second row)} and a low score (negative prediction) to passive ones {(e.g., the door in first row and the tv in second row)}. {It should be noted that next-active-objects are not always central objects appearing at a large scale, as it is the case of the tap in the second row of and the dish in the fourth row of} \figurename~\ref{fig:prediction_samples}{(a).}
In the failure examples reported in \figurename~\ref{fig:prediction_samples}(b), the model fails to predict next-active-objects. For instance, in the first row of \figurename~\ref{fig:prediction_samples}(b), the model predicts oven/stove as the next-active-object, while the actual target (dish) is not predicted at all by the object detector/tracker. Similarly, {in the fourth row}, the target next-active-object fridge is correctly detected by the detector/tracker component, but erroneously classified as passive by the next-active-object prediction system. A possible reason for this failure might be the proximity of the object to the border. Videos demonstrating the proposed method are available at our web page: \url{http://iplab.dmi.unict.it/NextActiveObjectprediction/}.

{While implementing a real-time system is out of the scope of this paper, it should be noted that, since the SORT tracker is highly real-time and Random Decision Forests are fast at inference time, the computational performance of the proposed method is dominated by the Faster-RCNN object detection component. Using an NVIDIA Titan X GPU, our method can process video at about 5 frames per second. Such computational performance can be improved with the adoption of real-time object tracking methods such as the one proposed in}~\cite{redmon2016you}.

\section{Conclusion}%
\label{sec:prediction_conclusion}
We introduced and investigated the problem of \emph{next-active-object prediction} from egocentric videos. 
Experiments highlight that 1) active object trajectories can be discriminated from passive ones using absolute positions, scale and differential scale and position information, 2) active trajectory classifiers can be learned independently from object classes, 3) egocentric cues based on object motion outperform baselines based on other cues such as object appearance and the presence of hands.
{In future work, we will extend the analysis also to data acquired using head-mounted cameras.} Moreover, we will investigate 
the integration of other cues such as the way the appearance of objects and scene changes over time. 
{We are also interested in exploring how next-active-objects could benefit  a system for detecting first-person activities in egocentric video.}

\section*{Acknowledgment}
We gratefully acknowledge the support of NVIDIA Corporation with the donation of the Titan X Pascal GPU used for this research. This research is supported in part by ONR PECASE N00014-15-1-2291 (KG).

\bibliography{mybibfile}

\end{document}